\documentclass[letterpaper]{article} 
\usepackage[draft]{aaai2026}  
\usepackage{times}  
\usepackage{helvet}  
\usepackage{courier}  
\usepackage[hyphens]{url}  
\usepackage{graphicx} 
\usepackage{natbib}  
\usepackage{caption} 
\usepackage{times}
\usepackage{helvet}
\usepackage{courier}
\usepackage{algorithm}
\usepackage{algorithmic}

\usepackage{appendix}
\usepackage{caption}
\usepackage{dsfont} 
\usepackage{amsfonts}
\usepackage{bbding}
\usepackage{amssymb}
\usepackage{subcaption}
\usepackage{units}
\usepackage{bbding}
\usepackage{amsmath}
\usepackage{tabularx}
\usepackage{tabulary}
\usepackage{booktabs}
\usepackage{multirow}
\usepackage{makecell}
\usepackage{cancel}
\usepackage[table, dvipsnames]{xcolor}
\newtheorem{theorem}{Theorem}

\newtheorem{proof}{Proof}[section]

\newtheorem{definition}{Definition}
\newtheorem{assumption}{Assumption}

\definecolor{magiccolor}{RGB}{205, 232, 248}
\usepackage{newfloat}
\usepackage{listings}
\DeclareCaptionStyle{ruled}{labelfont=normalfont,labelsep=colon,strut=off} 
\floatstyle{ruled}
\newfloat{listing}{tb}{lst}{}
\floatname{listing}{Listing}
\title{GCHR : Goal-Conditioned Hindsight Regularization for \\ Sample-Efficient Reinforcement Learning}
\author{
    Xing Lei\textsuperscript{\rm 1},
    Wenyan Yang\textsuperscript{\rm 2},
    Kaiqiang Ke\textsuperscript{\rm 3},
    Shentao Yang\textsuperscript{\rm 4},
    Xuetao Zhang\textsuperscript{\rm 1},\\
    Joni Pajarinen\textsuperscript{\rm 2},
    Donglin Wang\textsuperscript{\rm 5}
}
\affiliations{
    \textsuperscript{\rm 1}Xi'an Jiaotong University\\
    \textsuperscript{\rm 2}Aalto University\\
    \textsuperscript{\rm 3}Sun Yat-sen University\\
    \textsuperscript{\rm 4}University of Texas at Austin\\
    \textsuperscript{\rm 5}Westlake University\\
}
\usepackage{bibentry}

\begin{document}

\maketitle

\begin{abstract}
Goal-conditioned reinforcement learning (GCRL) with sparse rewards remains a fundamental challenge in reinforcement learning. While hindsight experience replay (HER) has shown promise by relabeling collected trajectories with achieved goals, we argue that trajectory relabeling alone does not fully exploit the available experiences in off-policy GCRL methods, resulting in limited sample efficiency. In this paper, we propose Hindsight Goal-conditioned Regularization (HGR), a technique that generates action regularization priors based on hindsight goals. When combined with hindsight self-imitation regularization (HSR), our approach enables off-policy RL algorithms to maximize experience utilization. Compared to existing GCRL methods that employ HER and self-imitation techniques, our hindsight regularizations achieve substantially more efficient sample reuse and the best performances, which we empirically demonstrate on a suite of navigation and manipulation tasks.

\end{abstract}


\section{Introduction}
Deep reinforcement learning (RL) has been successful in a variety of tasks, including controlling robots \citep{zheng2024stabilizing,li2025comprehensive}, playing computer games \citep{roayaei2024maximize,rao2025isfors}, and understanding natural language \citep{uc2023survey,shinn2024reflexion}.
Goal-conditioned Reinforcement Learning (GCRL) \citep{liu2022goal} is a subdomain that trains agents to reach desired goal, making it an important step toward real-world robot control and general intelligence \citep{park2024ogbench}.
One of
the most essential factors affecting sample efficiency in GCRL is the sparse reward, in which
case informative learning signals only appear at the end of the trajectory. 
To address this problem, an active line of research focuses on designing novel learning algorithms that efficiently use the data.
One popular approach is Hindsight
Experience Replay (HER) \citet{andrychowicz2017hindsight}, which replaces the desired goal in the data with the one the agent actually reached, effectively increasing the frequency of the learning signals \citep{zheng2024does}. Since HER may introduce some bias during the training, the following works proposed addtional learning or planning techniques to correct the HER bias~\citep{li2020generalized, he2020soft, schramm2023usher, yang2021mher}.


\begin{figure}[t]
   \centering
   \centerline{\includegraphics[width=0.48\textwidth]{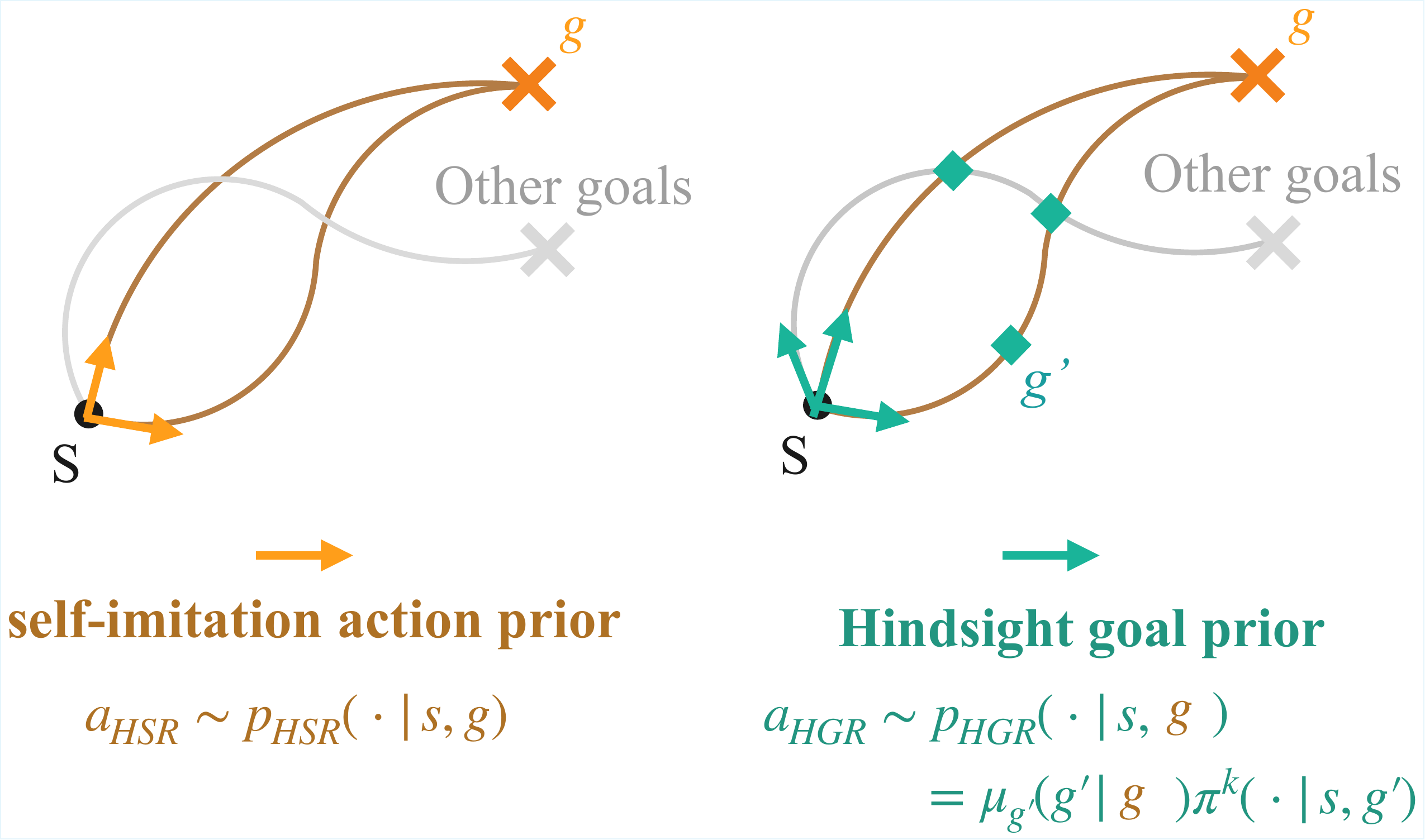}}
   \caption{We propose two policy regularizations to achieve sample-efficient goal-conditioned policy learning: \textbf{H}indsight \textbf{S}elf-imitation \textbf{R}egularization (\textbf{HSR}) and \textbf{H}indsight \textbf{G}oal \textbf{R}egularization (\textbf{HGR}). During training, we aim to regularize the policy $\pi$ to stay close to two action priors through KL regularization: the HSR prior ($\pi_\text{HSR}$) and the HGR prior ($\pi_\text{HGR}$). The HSR prior $\pi_\text{HSR}$ is generated through hindsight experience replay and behavioral cloning, encouraging the policy to reproduce successful past actions. The HGR prior $\pi_\text{HGR}$ is constructed by sampling hindsight goals $g'$ from the hindsight goal distribution $\mu_{g'}(\cdot|g)$ (where $g$ is the desired goal) and aggregates the corresponding actions from a delayed-update target policy $\pi'$. This enables the policy to maximize past experience usage.
   }
   \label{fig:GCHR}
\end{figure}

Based on HER, simple yet efficient self-imitation learning methods have been proposed to learn goal-conditioned policies\citep{ghosh2021learning,yang2022rethinking, ma2022offline, hejna2023distance, eysenbach2022imitating}. These works leverage the hindsight relabeled experience with weighted behavior cloning to learn goal-conditioned policies. Theoretically, they guarantee to optimize over a
lower bound of the objective for goal reaching problem~\citep{yang2022rethinking, hejna2023distance}. In addition to the purely self-imitation works, some GCRL methods also propose to use  self-imitation learning techniques as policy regularizations (such as Imagined Subgoals (RIS) \citep{chane2021goal} and GRSIL \citep{li2023self}.)


However, these methods typically require additional networks or planning, which makes them prone to training instability and high computational \citep{li2020generalized, he2020soft, yang2021mher, chane2021goal,schramm2023usher}. In addition, we find that HER has limited action coverage of the collected experiences, which limit the ability of maximizing sample-efficiency (see Figure~\ref{fig:GCHR}, where the hindsight action set $a_\text{HSR}$ covers limited goal-conditioned actions).

In this paper, we propose simple yet effective  Goal-Conditioned Hindsight Regularization (GCHR), which \textit{maximizes the off-policy GCRL's experience utilization} for sample-efficient goal-conditioned policy learning. 
GCHR incorporates two HGR techniques: Hindsight Self-Imitation Regularization (HSR) and Hindsight Goal Regularization (HGR). These components are designed to complement and enhance policy learning, as illustrated in Figure \ref{fig:GCHR}. First, we integrate the well established hindsight self-imitation regularization (HSR), which is HER with self-imitation regularization to optimize the relabeled goal policy; In addition to HSR, we use HGR to generate wider action prior, which covers the HSR's action set and is able to continuously optimize its HGR prior throughout training, adapting to the evolving policy distribution.  We theoretically show that our method can learn a better policy than both the self-imitation policy and the goal-conditioned RL methods. Our contributions can be summarized as followings:
\begin{itemize}
    \item Our proposed goal-conditioned hindsight regularizations (GCHR), combining HGR and HSR, significantly improve sample efficiency and performance in off-policy GCRL.
    \item Our analysis shows that HGR achieves broader action coverage using only collected experiences, maximizing experience utilization compared to HER with self-imitation techniques.
    \item GCHR is simple to implement: requiring only five lines of code without additional training modules or planning algorithms.
\end{itemize}


\section{Related Work} \label{sec:prior-work}
Our work concentrates on sample efficiency in goal-conditioned reinforcement learning (GCRL) with sparse rewards.
HER \citep{andrychowicz2017hindsight} addresses the issue of spare rewards in GCRL by regarding the achieved goal as the desired goal to train the action value. Specifically, the visited trajectories are used to sample the suitable
subgoals, where some of the non-sparse rewards are designed to enable
effective training. Based on HER, a few studies have been investigated to find more critical goal for desired goal. CHER \citep{fang2019curriculum} selects goals in a heuristic way to balance the diversity of selected goals and the proximity to original goals.
Nevertheless, the curriculum is hand-designed and needs to be adjusted
through hyperparameters.
MHER \citep{yang2021mher} constructs a dynamics model using historical trajectories and combines current
policy and desired goal to generate virtual relabeled goals. Although MHER additionally samples the desired goal and enhances exploration efficiency, they face the issue of inaccurate learning in the dynamics model.

In addition to improving the goal sampling methods,
recent advances in self-imitation have further improved learning efficiency, as exemplified by methods such as Goal-conditioned Supervised Learning (GCSL) \citep{ghosh2021learning}, Weighted Goal-conditioned Supervised Learning (WGCSL) \citep{yang2022rethinking}, $f$-weighted Regression (GoFar) \citep{ma2022offline} and Distance Weighted Supervised Learning (DWSL) \citep{hejna2023distance}.
However, they always biased toward past experience and can lead to suboptimal \citep{eysenbach2022imitating}; while our HGR's prior is monotonically improve during policy training.

Self-Imitation learning (SIL) can be applied to conventional goal-conditioned RL methods, where samples from past trajectories
address issues such as sparse reward or facilitate efficient learning \citep{tang2020self}.
SIL \citep{oh2018self} aims to reproduce the agent's past good decisions by learning
from visited trajectories.
RIS \citep{chane2021goal} optimizes an objective incorporating the divergence between the target policy and a prior policy.
GRSIL \citep{li2023self} aggregate the self-imitated policy and the goal-conditioned
policy by greedily taking the most promising one that leads to a larger
estimated future return.
We share a similar idea with RIS and GRSIL; however, unlike these methods, we eliminate the training instability introduced by learning additional value functions or policies. Furthermore, we show that our regularizations can cover a broader action space than previous self-imitation learning methods, thereby improving exploration efficiency.

\section{Preliminaries}\label{sc:prelimi}
\subsection{Goal-Conditioned Reinforcement Learning}
We consider a goal-conditioned Markov Decision Process (MDP) defined by the tuple $(\mathcal{S}, \mathcal{A}, \mathcal{G}, P, r, \gamma)$, where $\mathcal{S}$ is the state space, $\mathcal{A}$ is the action space, $\mathcal{G}$ is the goal space, $P: \mathcal{S} \times \mathcal{A} \rightarrow \Delta(\mathcal{S})$ is the transition dynamics, $r: \mathcal{S} \times \mathcal{G} \rightarrow \{0, 1\}$ is a sparse binary reward function, and $\gamma \in [0, 1)$ is the discount factor. Following prior work \citep{andrychowicz2017hindsight,2018Multi,yang2021mher}, we assume a deterministic state-to-goal mapping $\phi: \mathcal{S} \rightarrow \mathcal{G}$ that extracts goal-relevant features from states.
The reward function takes the form:
\begin{equation}
r(s, g) = \mathbf{I}\{\phi(s) = g\}
\end{equation}
This sparse indicator reward is appealing because it avoids requiring domain-specific distance metrics or hand-crafted shaping functions. A goal-conditioned policy $\pi(a | s, g)$ maps state-goal pairs to action distributions, with the objective of reaching states that satisfy the desired goal.
\subsubsection{Absorbing-Goal Formulation}
To connect value functions with reachability measures, we adopt an absorbing-goal formulation. Define $\mathcal{S}_g = \{s \in \mathcal{S} : \phi(s) = g\}$ as the set of states satisfying goal $g$. In this formulation:
\begin{itemize}
\item States in $\mathcal{S}_g$ are absorbing: once the agent reaches any state $s \in \mathcal{S}_g$, it remains there regardless of actions taken, i.e., $P(s' | s, a) = \mathbf{I}\{s' = s\}$ for all $a \in \mathcal{A}$
\item The reward remains $r(s, g) = \mathbf{I}\{\phi(s) = g\}$ at every timestep
\end{itemize}
This assumption simplifies analysis by making goal achievement permanent and allows us to interpret value functions through occupancy measures. The action-value function becomes:
\begin{align}
Q^\pi(s, a, g) = \mathbb{E}\left[\sum_{t=0}^{\infty} \gamma^t r(s_t, g) \,\Big|\, s_0 = s, a_0 = a, \pi, g\right] \nonumber\\= \sum_{\Delta=1}^{\infty} \gamma^\Delta \Pr_\pi(\phi(s_\Delta) = g \,|\, s_0 = s, a_0 = a)
\end{align}

\subsubsection{From Value Functions to Occupancy Measures}

The key insight is that value functions directly quantify how likely (and when) the policy will reach the goal. We formalize this through two occupancy measures \citep{eysenbach2020c}:

\begin{definition}[Future-state occupancy ]
\label{def:future_state_occu}
The $\gamma$-discounted probability of visiting state $s'$ after taking action $a$ in state $s$ under policy $\pi$ is:
\begin{equation}
\label{eq:future-occupancy}
\begin{split}
d^\pi(s' \mid s, a, g)
  &= (1 - \gamma)\sum_{\Delta=1}^{\infty}\gamma^\Delta\\
  &\quad\Pr_\pi\bigl(s_\Delta = s' \mid s_0 = s,\;a_0 = a,\;g\bigr)\,.
\end{split}
\end{equation}
\end{definition}
Correspondingly, we can define the marginal future-state occupancy
$
d^\pi\!\bigl(s' \mid s, g\bigr)
=
\mathbb{E}_{a_0 \sim \pi(\,\cdot \mid s, g)}
\!\bigl[\,d^\pi\!\bigl(s' \mid s, a_0, g\bigr)\bigr]
$.
The factor $(1-\gamma)$ ensures $\sum_{s'} d^\pi(s' | s, a, g) = 1$, making it a proper probability distribution.

\begin{definition}[First‐Hitting State Occupancy]
\label{def:hit_occu}
Based Definition~\ref{def:future_state_occu}, we define a  normalized "first‐hitting" distribution over the goal‐set $S_{g}=\{s:\phi(s)=g\}$ is
\begin{equation}
\tilde{d}^\pi(s' \mid s, g)
\;=\;
\frac{
  d^\pi(\phi(s')=g \mid s, g)
}{
  \displaystyle\sum_{\tilde{s}\in S_{g'}}d^\pi(\tilde{s}\mid s, g')
}
\end{equation}

\end{definition}
\emph{First‐hitting state occupancy} $\tilde d^\pi(s'\!\mid s, g')$ is the normalized, $\gamma$‐discounted probability that, under policy~$\pi$ targeting subgoal~$g'$ from initial state~$s$, the agent's \emph{first} arrival in the set of goal‐satisfying states $S_{g'}$ occurs at $s'$.

\begin{definition}[Future-goal density]
By marginalizing over all goal-satisfying states, we obtain the future-goal density:
\begin{equation}
\begin{aligned}
& p^\pi\bigl(g\mid s,a,g\bigr)
  = \sum_{s'\in\mathcal{S}_g}
    d^\pi\bigl(s'\mid s,a,g\bigr) 
  &&\\
&\quad\quad\quad= (1-\gamma)\sum_{\Delta=1}^{\infty}\gamma^\Delta
    \Pr_\pi\!\bigl(\phi(s_\Delta)=g
    \mid s_0=s,a_0=a\bigr)
  && \\
\end{aligned}
\end{equation}
\end{definition}
Correspondingly,  we define the marginal future-goal density:
$    p^\pi(g|s,g) = \mathbb{E}_{a \sim \pi(\cdot | s, g)}\bigl[p^\pi\bigl(g\mid s,a,g\bigr) \bigr]
$.
This density allows us to express:
\begin{equation}
Q^\pi(s, a, g) = \frac{1}{1-\gamma} p^\pi(g | s, a, g)
\end{equation}
\begin{equation}
V^\pi(s, g) = \mathbb{E}_{a \sim \pi(\cdot | s, g)}[Q^\pi(s, a, g)] = \frac{1}{1-\gamma} p^\pi(g | s, g)
\end{equation}
where $p^\pi(g | s, g)$ denotes the marginal future-goal density after sampling actions from the policy.
Intuitively, this connection shows that maximizing the value function is equivalent to maximizing the discounted probability of reaching the goal. We will use these definition to analyze our regularization technique in later of this work.


\subsection{Hindsight Goal Relabeling}
Sparse rewards pose a fundamental challenge in goal-conditioned RL: without reaching the desired goal $g$, the agent receives no learning signal. Hindsight Experience Replay (HER) \citep{andrychowicz2017hindsight} addresses this by relabeling failed trajectories with achieved goals, transforming them into successful experiences.

Given a trajectory $\tau = (s_0, a_0, \ldots, s_T)$ that aimed for goal $g$, we can relabel transitions with goals that were actually achieved. For each timestep $t$, the set of achievable future goals is:
\begin{equation}
\mathcal{G}_t^{\text{future}}(\tau) = \{\phi(s_{t'}) : t \leq t' \leq T\}
\end{equation}
A relabeling strategy $\mathcal{R}$ selects hindsight goals $\hat{g}_t$ from $\mathcal{G}_t^{\text{future}}(\tau)$ (e.g., final state, random future state). This produces an augmented dataset:
\begin{equation}
\mathcal{D}_{\text{HER}} = \{(s_t, a_t, s_{t+1}, \hat{g}_t) : (s_t, a_t, s_{t+1}) \in \tau, \hat{g}_t = \mathcal{R}(\tau, t)\}
\end{equation}
By construction, these relabeled transitions have positive rewards since $\phi(s_{t'}) = \hat{g}_t$ for some $t' > t$, providing dense learning signal even from initially unsuccessful trajectories.

\section{Method}\label{sec:method}
In this section, we present Goal-Conditioned Hindsight Regularization for sample-efficient policy learning. Our key insight is that trajectories collected during exploration contain valuable information about reachability between different goals, which can be leveraged to accelerate learning. We propose two complementary regularization techniques that exploit this information: (1) Hindsight Action Regularization (HSR), which encourages reproducing successful goal-reaching behaviors, and (2) Hindsight Goal Regularization (HGR), which constructs informative action priors from visited goals.

\subsection{Assumption:  Uniform Reachability}   
In this work, we assume that any state that already satisfies $g$ can reach (up to an arbitrarily small neighbourhood) any other state that also satisfies $g$, and that this property induces uniform reachability to other goals. More specifically:
\begin{assumption}[Uniform Reachability]\label{asmp:internal}
Given a goal $g \in \mathcal{G}$ and $S_g = \{\,s \in \mathcal{S} \mid \phi(s) = g\,\}$ denotes the set of states satisfy goal~$g$:
\begin{enumerate}
\item  For any state $s \in S_g$, there exists a finite sequence of actions that drives the agent from $s$ to a any arbitrary state $s_g \in S_g$.
\item For a state $s\in S_g$ , if there exist any arbitrary goal $g',g'\neq g$, and a policy $\pi$ satisfies  $V^\pi(s, g') > 0$. Then for any arbitrary state $s_g \in S_g$, there is a small $\delta > 0$.
\begin{equation}
\bigl|V^\pi(s, g') - V^\pi(s_g, g')\bigr| < \delta  
\end{equation}
\end{enumerate}
\end{assumption}
$S_g$ is the equivalence class of all states that satisfy goal $g$, forming a manifold in state space where different states map to the same goal achievement. 
This assumption captures the intuition that states in the  same goal-specific state set can reach each other through appropriate actions. consequently, all states within the same goal set have similar reachability properties to other goals.

This assumption also naturally satisfies many domains such as 1) navigation tasks, where states at the same location $(x,y)$ but with different orientations/velocities can transition between each other and have similar distances to other locations; 2) Manipulation tasks, where different joint configurations achieving the same end-effector pose are connected through the configuration space and have similar reachability to other end-effector poses.

\subsection{Hindsight Self-imitation Regularization}\label{sec:HSR}
We begin with a well-established technique that forms the foundation of our approach. HER enables learning from failed trajectories by relabeling them with achieved goals. We formalize the self-imitation (behavioral cloning) component often used with HER as a regularization term.

Consider a trajectory $\tau = (s_0, a_0, \ldots, s_T)$ originally aimed at goal $g$ but reaching state $s_T$. Through hindsight relabeling, we obtain a dataset $\mathcal{D}_{\text{HER}}$ containing tuples $(s_t, a_t, g'_t)$ where $g'_t = \phi(s_{t'})$ for some $t' > t$ along the trajectory. For each hindsight-labeled transition, the action $a_t$ empirically led to achieving goal $\hat{g}_t$. We can view this as a self-imitation prior: given state $s_t$ and goal $\hat{g}_t$, the "good" action is $a_t$. Formally, this prior is a Dirac distribution $\delta_{a_t}$.
The hindsight self-imitation regularizer encourages the policy to reproduce these successful actions:
\begin{equation}
\mathcal{L}_{\text{HSR}} = \mathbb{E}_{(s_t, a_t, \hat{g}_t) \sim \mathcal{D}_{\text{HER}}} \left[ \log \pi_\theta(a_t \mid s_t, g'_t) \right]
\label{eq:HSR}
\end{equation}
This regularization has been shown to stabilize learning and improve sample efficiency in goal-conditioned settings. However, it only leverages information about achieved goals, not the relationships between different goals along trajectories.  We argue that solely use HSR does not fully leverage the collected experience and we propose hindsight goal regularization to for better sample-efficiency.

\subsection{Hindsight Goal Regularization}\label{sec:hgr}
While HSR exploits successful goal-reaching behaviors, it does not leverage the rich structure of goal-to-goal reachability implicit in collected trajectories. We introduce Hindsight Goal Regularization (HGR) to address this limitation. 
\subsubsection{Motivation}
Consider a trajectory that visits states $s_0, s_1, \ldots, s_T$, corresponding to goals $g'_0 = \phi(s_0), g'_1 = \phi(s_1), \ldots, g'_T = \phi(s_T)$. From state $s_i$, the agent has demonstrated the ability to reach any subsequent goal $g'_j$ for $j > i$. Based on our Assumption~\ref{asmp:internal}, if any subsequent goal $g'_j$ (where $ i\leq j\leq T$ can reach the task goal $g$, then for all subsequent state $s_i$ (where $0\leq i\leq j$) can also leading toward the desired goal $g$. 

In other words, if there exist an action $a$ that lead state $s$ to a goal $g'$, and there exist a reachability between $g'$ and $g$, then $a$ is possible to lead $s$ to $g$. Based on this motivation, Given a trajectory $\tau = (s_0, a_0, \ldots, s_T)$ desired toward goal $g$, we define the hindsight goal set:
\begin{equation}
\mathcal{G}_{\text{H}}(\tau) = \{g_t' : 0 \leq t \leq T\}
\end{equation}
For a state $s$ in the trajectory and the desired goal $g$, we construct an action prior by aggregating policies toward visited goals. Specifically, we sample $K$ hindsight goals $\{g'_k\}_{k=1}^K$ from $\mathcal{G}_{\text{H}}(\tau)$ according to a distribution $\mu_{g'}(\cdot|g)$ (e.g., uniform sampling).
Using a delayed target policy $\pi'$ (as in off-policy RLs such as  SAC \citep{haarnoja2018soft}), we define the hindsight action prior:
\begin{equation}
\pi_{\text{HG}_\text{prior}}(a \mid s, g) = \mathbb{E}_{g' \sim \mu_{g}(\cdot|g)} \left[ \pi'(a \mid s, g') \right] 
\label{eq:hindsight_prior}
\end{equation}
A visual example is given in Fig.~\ref{fig:GCHR} ($a_\text{HGR}$). This prior represents a mixture of actions that would be taken if targeting the visited goals. The intuition is that these actions encode useful exploratory behaviors that have been proven to reach various goals.
Based on this prior $\pi_{\text{HG}_{\text{prior}}}$,  we regularize the policy toward this hindsight prior by minimizing KL divergence:
\begin{equation}
\mathcal{L}_{\text{HGR}} = -\mathbb{E}_{(s, g) \sim \mathcal{D}} \left[ D_\text{KL}\left( \pi_{\text{HG}_{\text{prior}}}(\cdot \mid s, g) \,\|\,  \pi_\theta(\cdot \mid s, g)\,\right) \right]
\label{eq:hgr}
\end{equation}
This KL divergence regularization encourages the policy to cover the support of the hindsight prior, promoting exploration toward diverse goals while maintaining focus on the desired goal through the Q-function optimization.

\subsection{Overall Objective}\label{sc:total objective}

We introduce a unified learning objective that maximizes the goal-conditioned Q-function while penalizing deviations in the action distribution via two KL-based regularizations. Concretely, we seek a policy $\pi$ that both improves task performance and stays close to (i) hindsight-relabeled behavior and (ii) a given prior (see Algorithm~\ref{alg:GCHR} in Appendix \ref{ap:GCHR algorithm}):

\begin{equation}
\begin{aligned}
\max_{\pi}\quad
&\underbrace{\mathbb{E}_{(s,g)\sim\mathcal{B}\cup\mathcal{B}_r,\;a\sim\pi(\cdot\mid s,g)}
\bigl[Q^{\pi}(s,a,g)\bigr]}_{\text{task performance}}\\
&\;-\;\alpha\;\underbrace{\mathbb{E}_{(s,g,g')\sim\mathcal{B}_r}\bigl[-\log\pi(a\mid s,g')\bigr]}_{\text{HSR}}
\;\\&-\;\beta\;\underbrace{\mathbb{E}_{(s,g)\sim\mathcal{B}}\bigl[D_{\mathrm{KL}}\bigl( \pi_{\text{HG}_\text{prior}}(\cdot\mid s,g) \,\|\, \pi(\cdot\mid s,g)\bigr)\bigr]}_{\text{HGR}}.
\end{aligned}
\label{eq:gchr}
\end{equation}


\section{Theoretical Analysis }\label{sec:theory}

In this section, we analyze why HGR as a regularization contributes to GCRL learning. We examine two key perspectives: 1) HGR's expanded action coverage compared to HSR, and 2) how HGR's prior improves through compositional value functions.

\subsection{HGR vs HSR: Action Coverage Analysis}

We now analyze how HGR expands the action space coverage compared to HSR. While HSR can only replay actions that historically achieved the desired goal, HGR leverages actions that reached \textit{any} goal visited along a trajectory. This seemingly simple difference has profound implications for exploration and compositional learning.
Consider a state $s$, desired goal $g$, and trajectories in the replay buffer $\mathcal{D}$. For any trajectory $\tau \in \mathcal{D}$, let $\mathcal{G}_H(\tau) = \{g_0', g_1', \ldots, g_T'\}$ denote the set of achieved goals along $\tau$.

\begin{definition}[Action Support]
For a state $s$ and desired goal $g$, define the action supports:
\begin{align}
\mathcal{A}_{\text{HSR}}(s,g) &= \{a \in \mathcal{A} : \exists \tau \in \mathcal{D},\, (s,a) \in \tau \text{ led to } g\}\\
\mathcal{A}_{\text{HGR}}(s,g) &= \bigcup_{g' \in \bigcup_{\tau \in \mathcal{D}} \mathcal{G}_H(\tau)} \{a : \pi'(a|s,g') > 0\}
\end{align}
where $\mathcal{G}_H(\tau)$ is the set of goals visited along trajectory $\tau$.
\end{definition}
Our main result establishes that HGR provides broader action coverage:

\begin{theorem}[Coverage Expansion]\label{thm:coverage}
For any state-goal pair $(s,g)$:
\begin{equation}
 \mathcal{A}_{\text{HSR}}(s,g) \subseteq \mathcal{A}_{\text{HGR}}(s,g)   
\end{equation}
In particular, when state $s$ has never reached goal $g$ directly, HSR has no actions to suggest ($\mathcal{A}_{\text{HSR}}(s,g) = \emptyset$), while HGR can still propose actions that worked for other goals. 
\end{theorem}
 Proof is in Appendix \ref{ap:proof1}.The expanded action coverage enables HGR to explore through compositional strategies.  A simple example is also shown in  Figure~\ref{fig:GCHR}, HSR simply imitate the past experience, while HGR can cover additional actions via hindsight goal action generation. 

 By generating direct action toward hindsight goals, HGR can leverage the compositional structure of goal achievement---using previously visited goals as stepping stones toward unachieved goals. This expanded coverage, combined with the continuous optimization of the behavioral prior (Section \ref{sc:total objective}), provides HGR with an implicit curriculum that guides exploration toward the frontier of achievable goals. 

While HGR does not incorporate explicit exploration mechanisms, it achieves broader action coverage through its relabeling strategy. This improved coverage emerges naturally from the method's ability to leverage diverse past experiences, providing an implicit exploration benefit despite the absence of explicit exploration design (Section \ref{ap:exploration visual}).
\subsection{Monotonic HGR prior Improvement}
We first formalize a compositional strategies to achieve a task goal. Consider a  policy performs a two stage execution: from state $s$, the policy is given a subgoal $g'$ and the policy first tries to reach $g'$; then from the reached state $s'$ ($s'\in S_{g'}$) to reach the final goal $g'$. Correspondingly, the value function can be defined as follows:
\begin{definition}[Via-Goal Value Function]
The value of reaching goal $g$ via intermediate goal $g'$ is:
\begin{equation}
V_{\text{via}}^\pi(s,g;g') = p^{\pi}(g'|s) \sum_{s' \in S_{g'}} \tilde{d^\pi}(s'|s,g') \cdot V^{\pi}(s',g)
\end{equation}
\end{definition}
where:
 $p^{\pi}(g'|s) =
\mathbb{E}_{a\sim\pi(\cdot\mid s,g')}
\bigl[p^\pi(g'\!\mid s,a,g')\bigr]$ is the probability of achieving the absorbing goal~$g'$ from state~$s$ based on the policy $\pi$.
$\tilde{d^\pi}(s'|s,g')$ is the normalized first-hitting occupancy distribution (see Definition~\ref{def:hit_occu}).

This value function captures the expected return of a compositional strategy: \textit{first reach intermediate goal $g'$ with probability $p^{\pi}(g'|s)$, then from the resulting state distribution over $S_{g'}$, reach the final goal $g$}. This is exactly how HGR leverages the compositional structure of goal achievement ($V_\text{via}^\pi$ in on-policy evaluation of $\pi_{\text{HG}_\text{prior}}$ ). 
Based on this definition, we establish that the HGR prior quality improves as the base policy improves.
\begin{figure*}[t]
    \centering
    \includegraphics[width=0.63\textwidth]{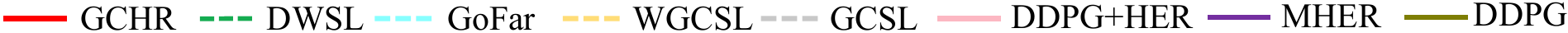}
    
    \begin{minipage}{0.22\linewidth}
        \centering
        \includegraphics[width=\textwidth]{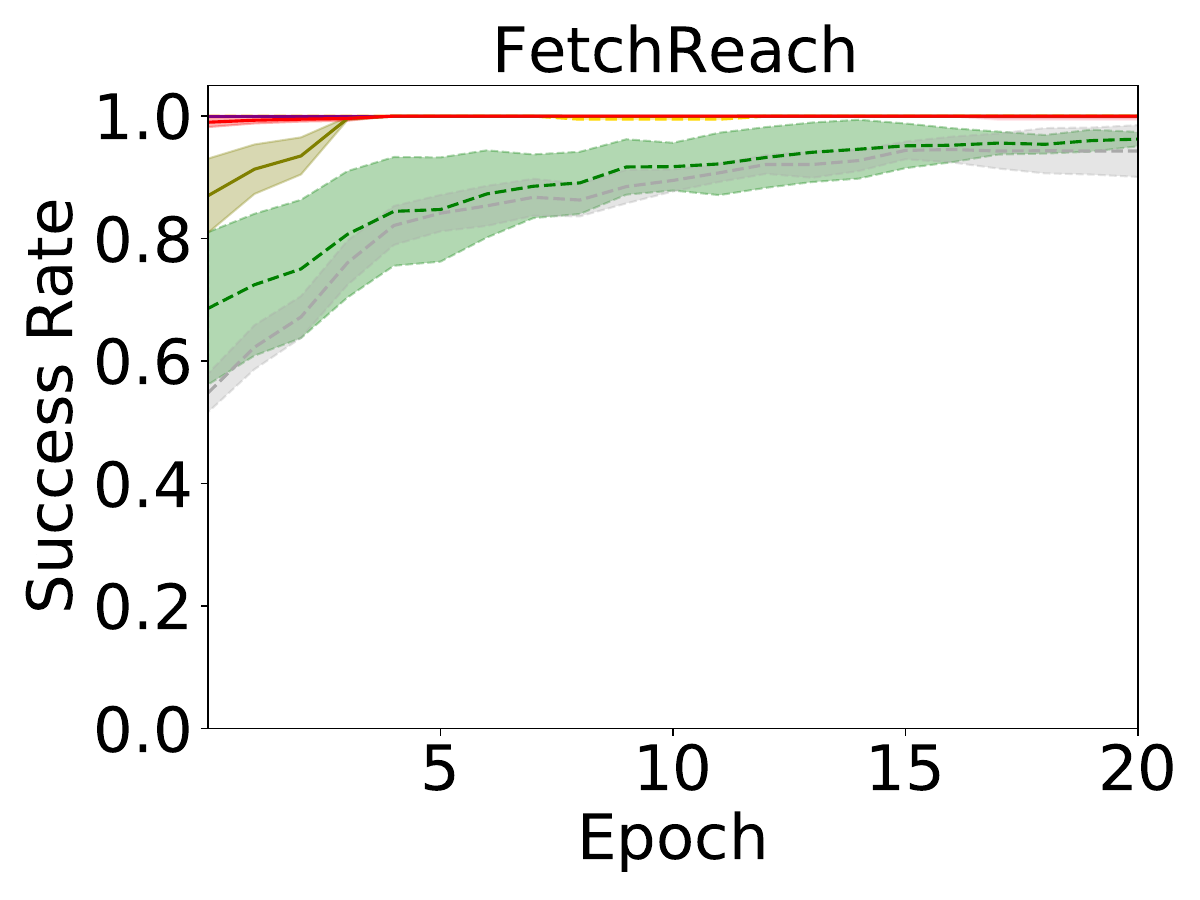}
    \end{minipage}\hfill
    \begin{minipage}{0.22\linewidth}
        \centering
        \includegraphics[width=\textwidth]{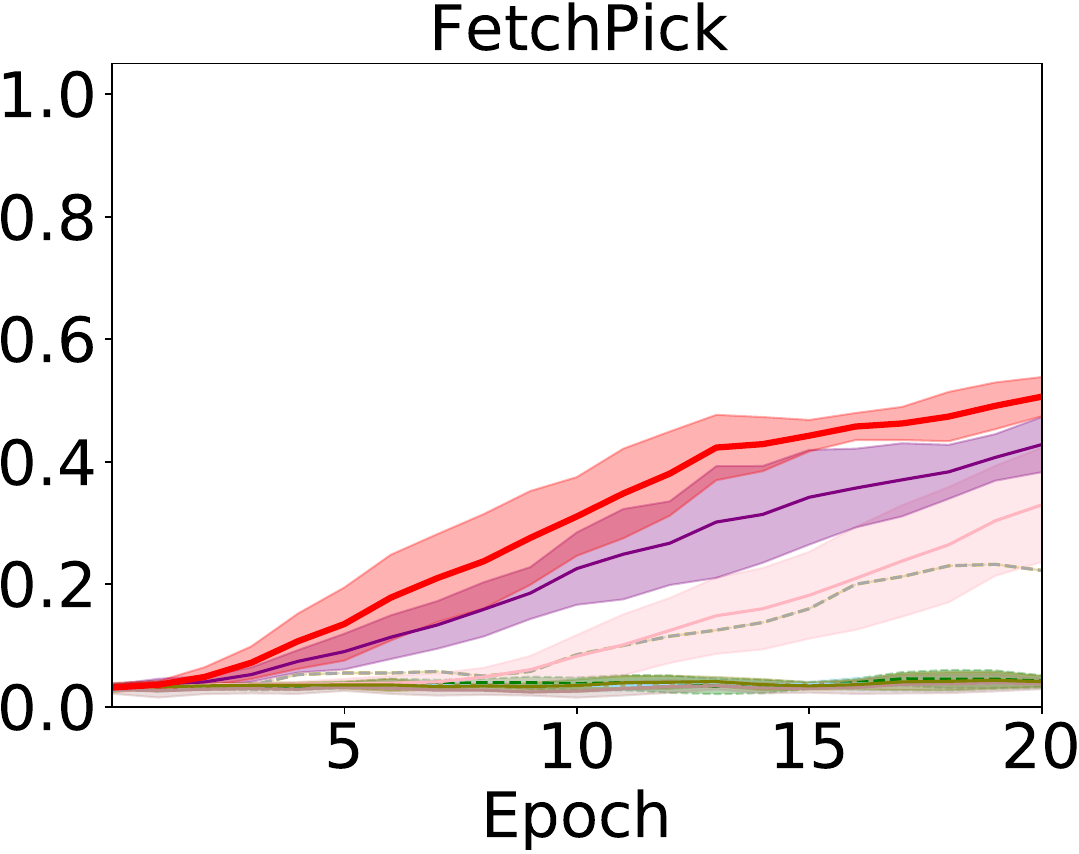}
    \end{minipage}\hfill
    \begin{minipage}{0.22\linewidth}
        \centering
        \includegraphics[width=\textwidth]{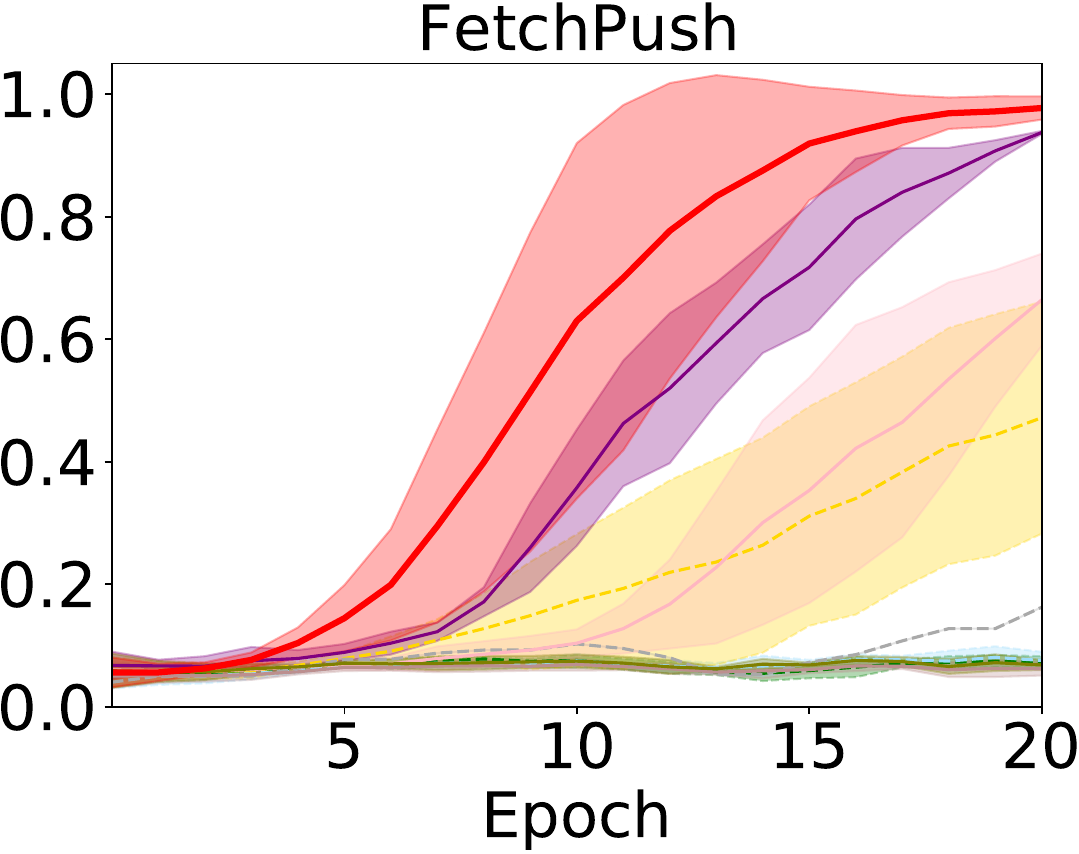}
    \end{minipage}\hfill
    \begin{minipage}{0.22\linewidth}
        \centering
        \includegraphics[width=\textwidth]{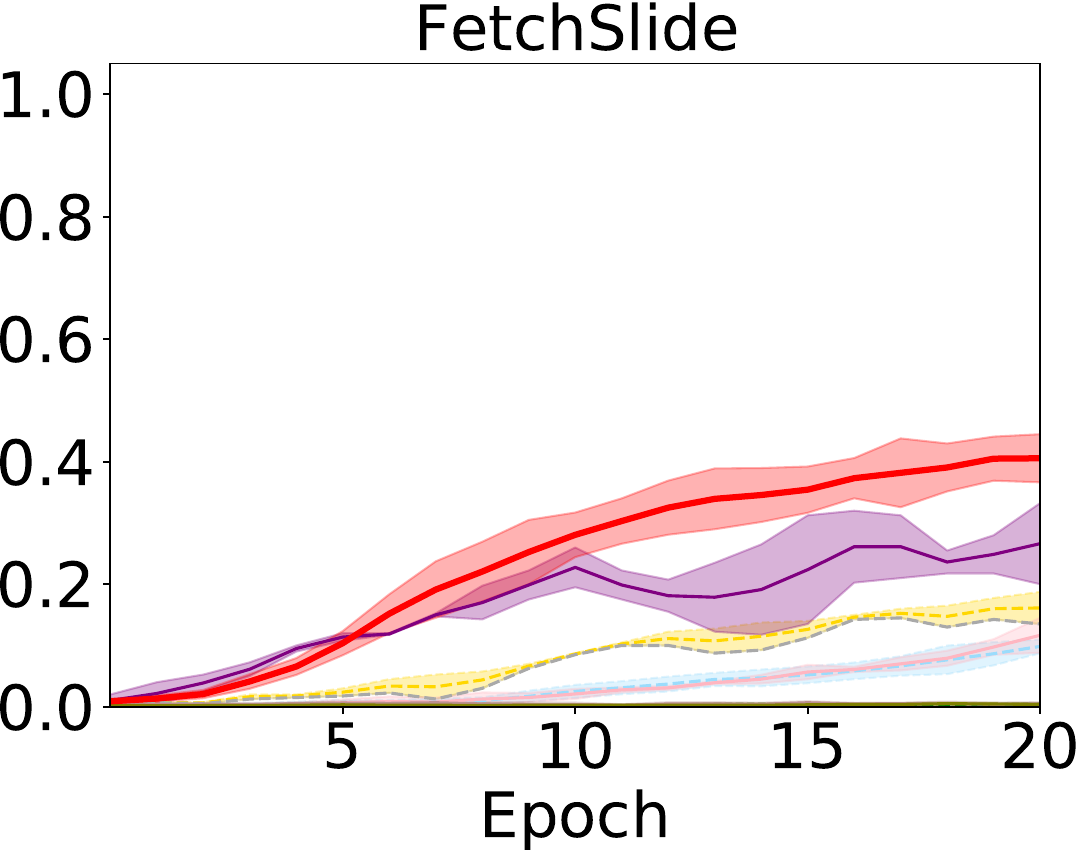}
    \end{minipage}
    
    \begin{minipage}{0.22\linewidth}
        \centering
        \includegraphics[width=\textwidth]{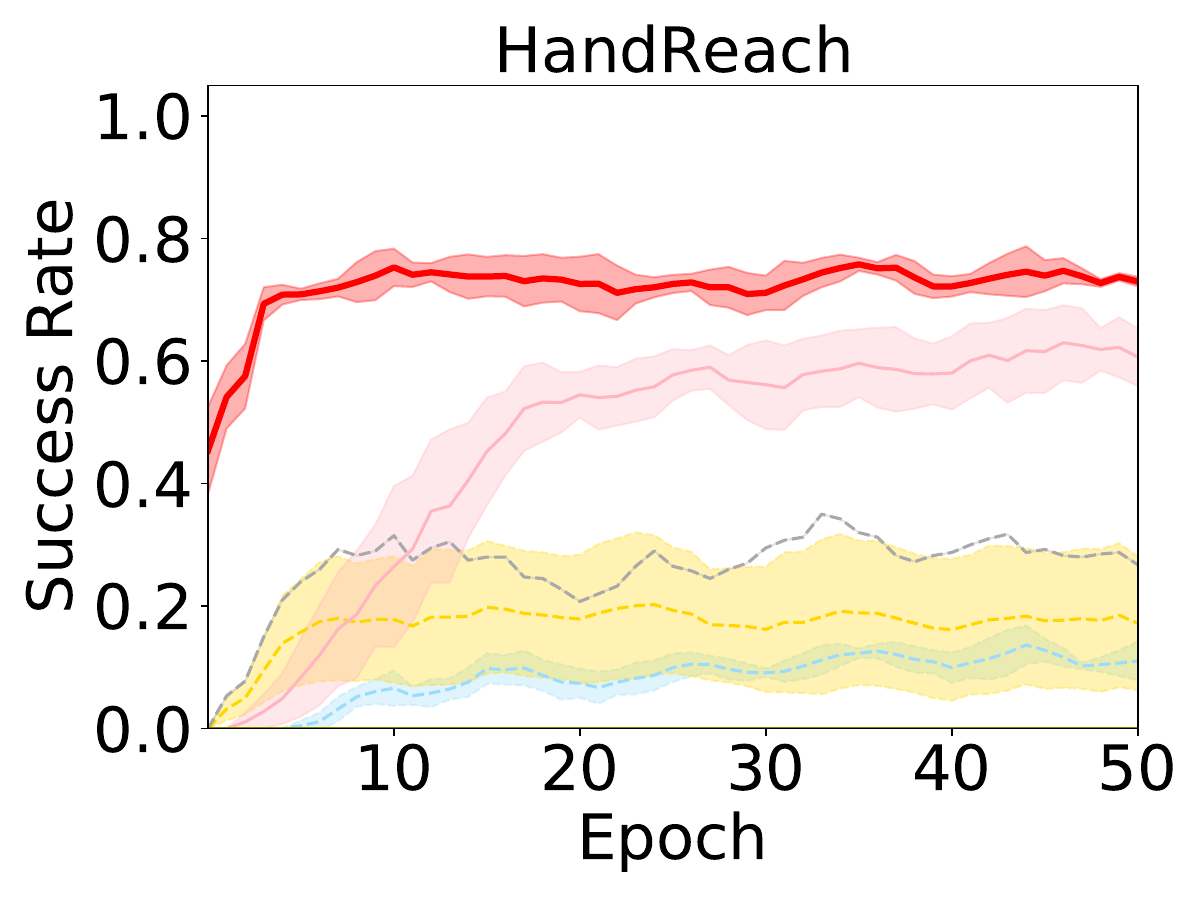}
    \end{minipage}\hfill
    \begin{minipage}{0.22\linewidth}
        \centering
        \includegraphics[width=\textwidth]{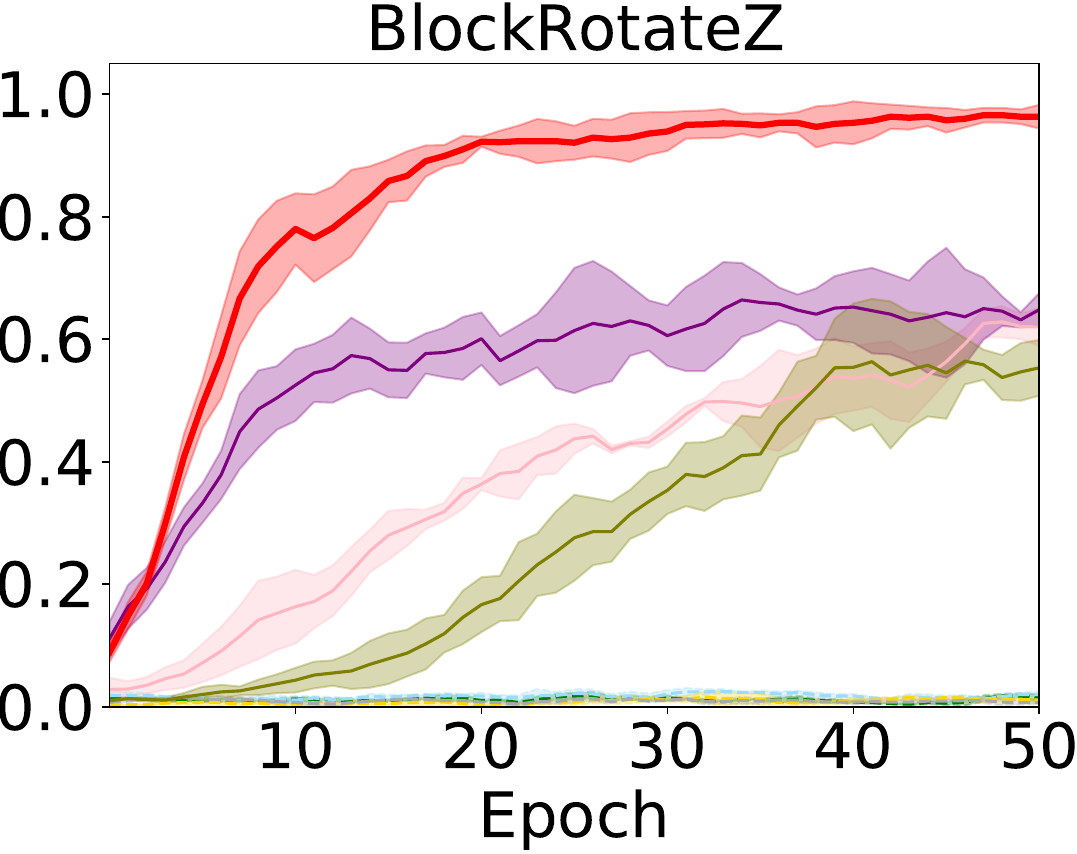}
    \end{minipage}\hfill
    \begin{minipage}{0.22\linewidth}
        \centering
        \includegraphics[width=\textwidth]{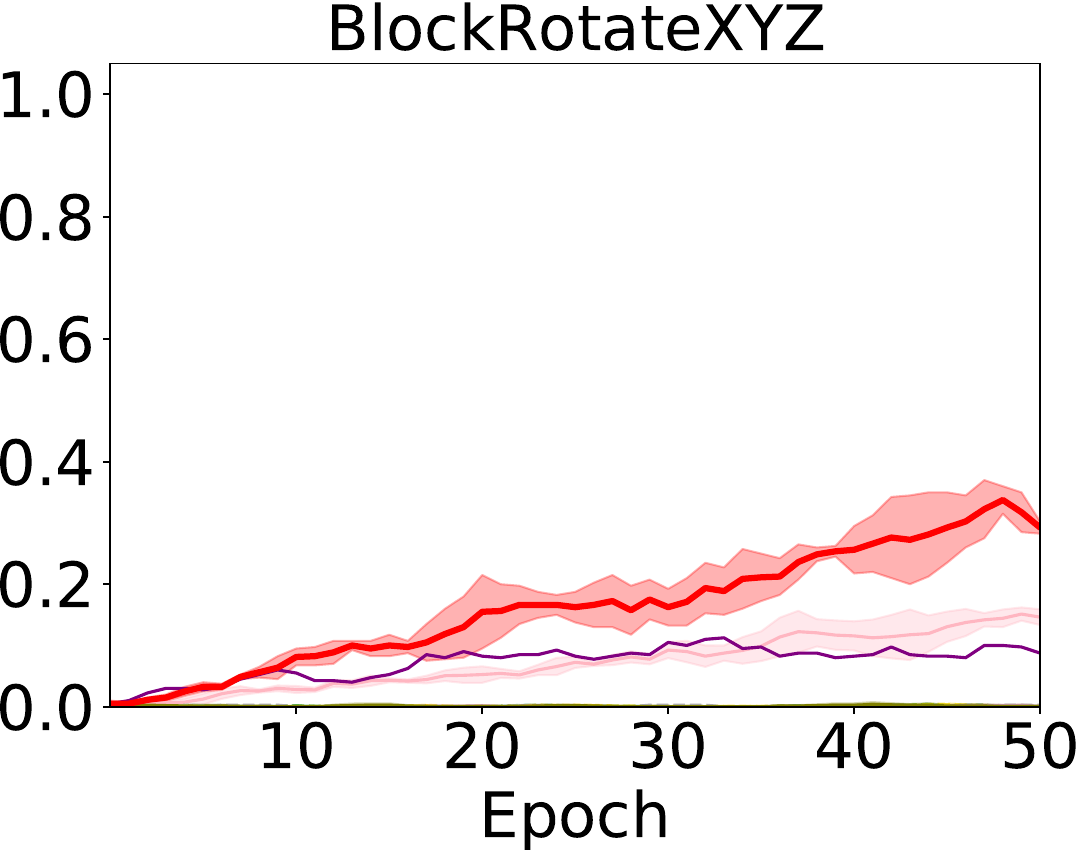}
    \end{minipage}\hfill
    \begin{minipage}{0.22\linewidth}
        \centering
        \includegraphics[width=\textwidth]{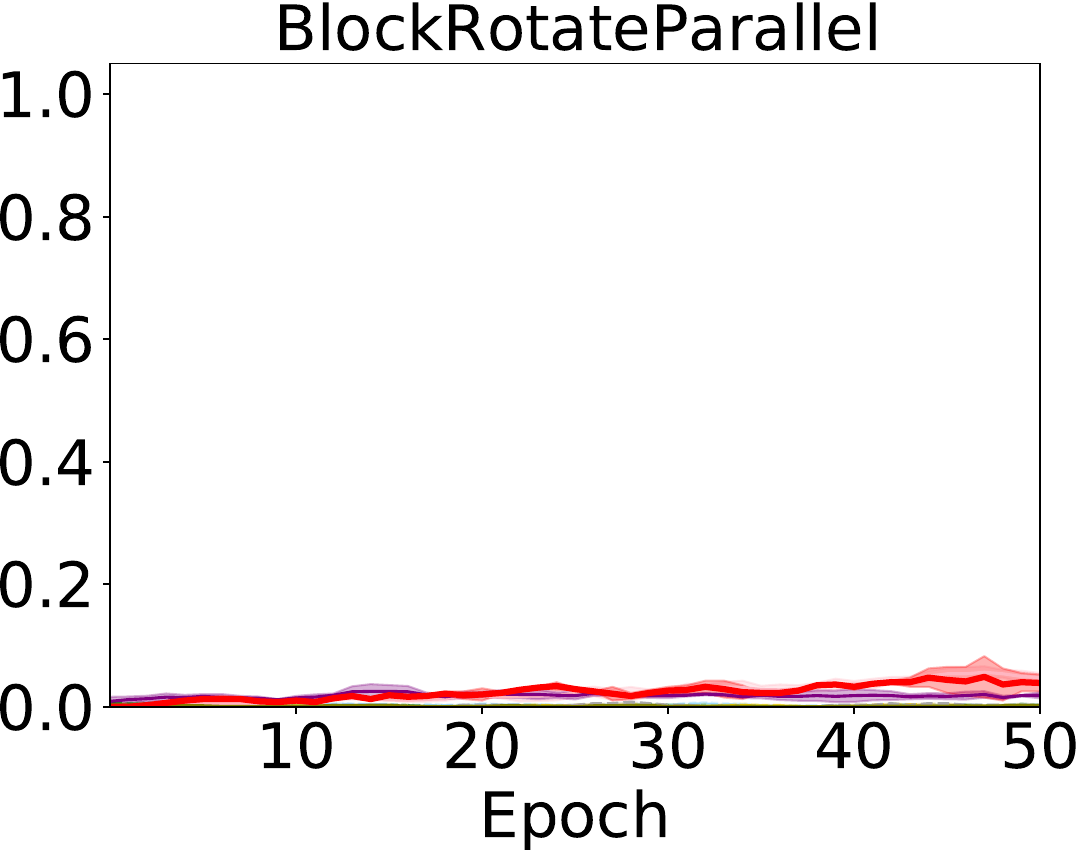}
    \end{minipage}
    
    \caption{Performance on 8 robot goal-reaching tasks in \citet{plappert2018multi}. Each epoch consists of 2000 interaction steps. GCHR (Ours) significantly improves learning efficiency and achieves the best performance.}
    \label{fig:multi_goal results}
\end{figure*}
\begin{theorem}[Prior Quality Improvement]\label{thm:monotonic}
Let $\pi^{(k)}$ be the policy at iteration $k$ with delayed version $\pi'^{(k)} = \pi^{(k-\tau_{\text{delay}})}$. Under Assumption~\ref{asmp:internal} , for any state $s$ and goal $g$ if:
\begin{equation}
\forall s,g: V^{\pi^{(k)}}(s,g) \geq V^{\pi^{(k-1)}}(s,g)
\end{equation}
then for any intermediate visited goal $g'$ between $s$ and $g$, the expected compositional value under the HGR prior improves:
\begin{equation}
\forall s,g, g': \left[ V_{\text{via}}^{\pi'^{(k)}}(s,g;g') \right] \geq  \left[ V_{\text{via}}^{\pi'^{(k-1)}}(s,g;g') \right]
\end{equation}
\end{theorem}
Proof is in Appendix \ref{ap:proof_monotonic}. The improvement follows by inspecting the \emph{two} factors of the via–goal value. First, 
Because the base policy is monotonically better, we have
$V^{\pi^{(k)}}(s,g') \!\ge\! V^{\pi^{(k-1)}}(s,g')$ for \emph{every} $g'$.
Via the identity $p^{\pi}(g'\!\mid s)= (1-\gamma)V^{\pi}(s,g')$  
(cf.\ Eq.~(3) in the preliminaries), this directly yields
$p^{\pi'^{(k)}}(g'\!\mid s) \!\ge\! p^{\pi'^{(k-1)}}(g'\!\mid s)$:
the agent is more likely to hit the intermediate goal.

Second, under Assumption~\ref{asmp:internal}, once the agent is inside
$S_{g'}$ it can manoeuvre within $S_{g'}$ and subsequently reach $g$.
The uniform policy improvement therefore implies
$V^{\pi'^{(k)}}(s',g)\!\ge\!V^{\pi'^{(k-1)}}(s',g)$
for \emph{all} $s'\!\in\!S_{g'}$.
Consequently the expectation
$W^{\pi'}(g',g)=\mathbb{E}_{s'\sim\tilde d^{\pi'}(\cdot\mid s,g')}
                 [V^{\pi'}(s',g)]$
also increases.

Since both factors in the product
$V_{\text{via}}^{\pi'}(s,g;g') = V^{\pi'}(s,g') \, W^{\pi'}(g',g)$
are non-decreasing, their product is non-decreasing as well.
Because the inequality holds \emph{point-wise} in $g'$,  
taking an expectation over any distribution $\mu_g(g')$ preserves it,
establishing that the HGR prior grows steadily stronger as training
progresses.


\section{Experiments} \label{sc:experiment}
We begin by presenting the benchmarks and baseline methodologies utilized in our study, accompanied by a detailed description of the experimental procedures. We then present the results and analyze how they support our initial assumptions and theoretical framework.
\begin{figure}[h]
    \centering
    \begin{minipage}{0.22\linewidth}

        \centerline{\includegraphics[height=1.25cm, width=1.25cm]{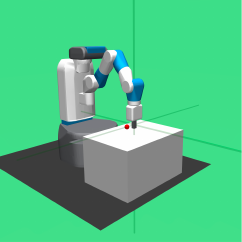}}
        \centerline{(a)}
    \end{minipage}
    \begin{minipage}{0.22\linewidth}

        \centerline{\includegraphics[height=1.25cm, width=1.25cm]{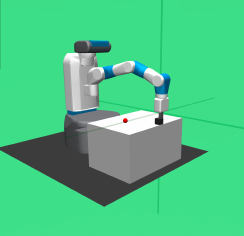}}
        \centerline{(b)}
    \end{minipage}
    \begin{minipage}{0.22\linewidth}

        \centerline{\includegraphics[height=1.25cm, width=1.25cm]{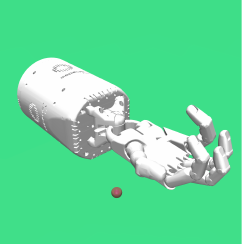}}
        \centerline{(c)}
    \end{minipage}
    \begin{minipage}{0.22\linewidth}

        \centerline{\includegraphics[height=1.25cm, width=1.25cm]{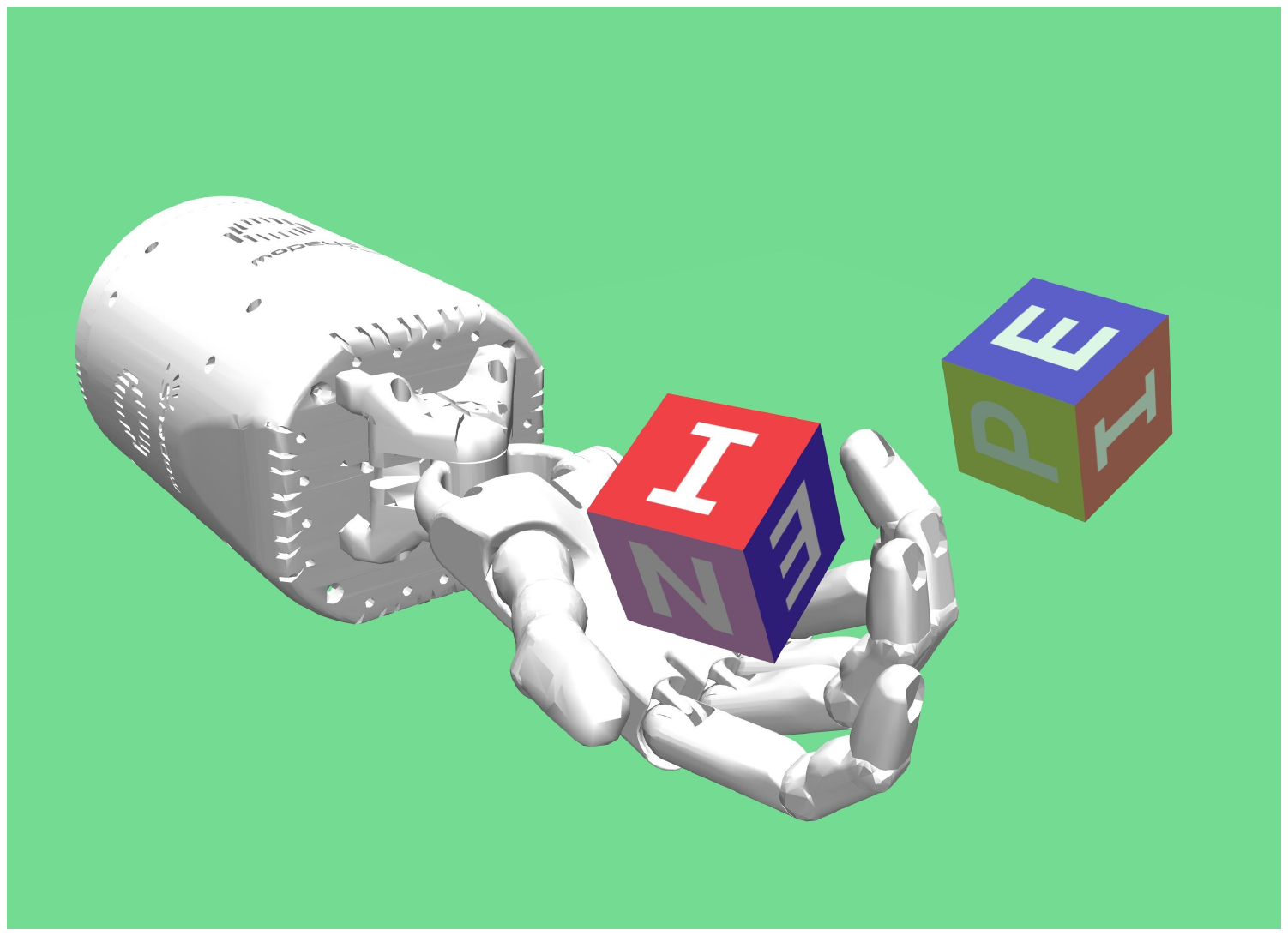}}
        \centerline{(d)}
    \end{minipage}
    \caption{
    Goal-conditioned example tasks:
    (a) FetchReach,
    (b) FetchPush,
    (c) HandReach.
    (d) HandManipulateBlock.
    }
    \label{fig:multi_goal exapmle tasks}
\end{figure}
\begin{figure*}[t]
    \centering
    \begin{minipage}{0.22\linewidth}
        
        \centerline{\includegraphics[width=0.88\textwidth]{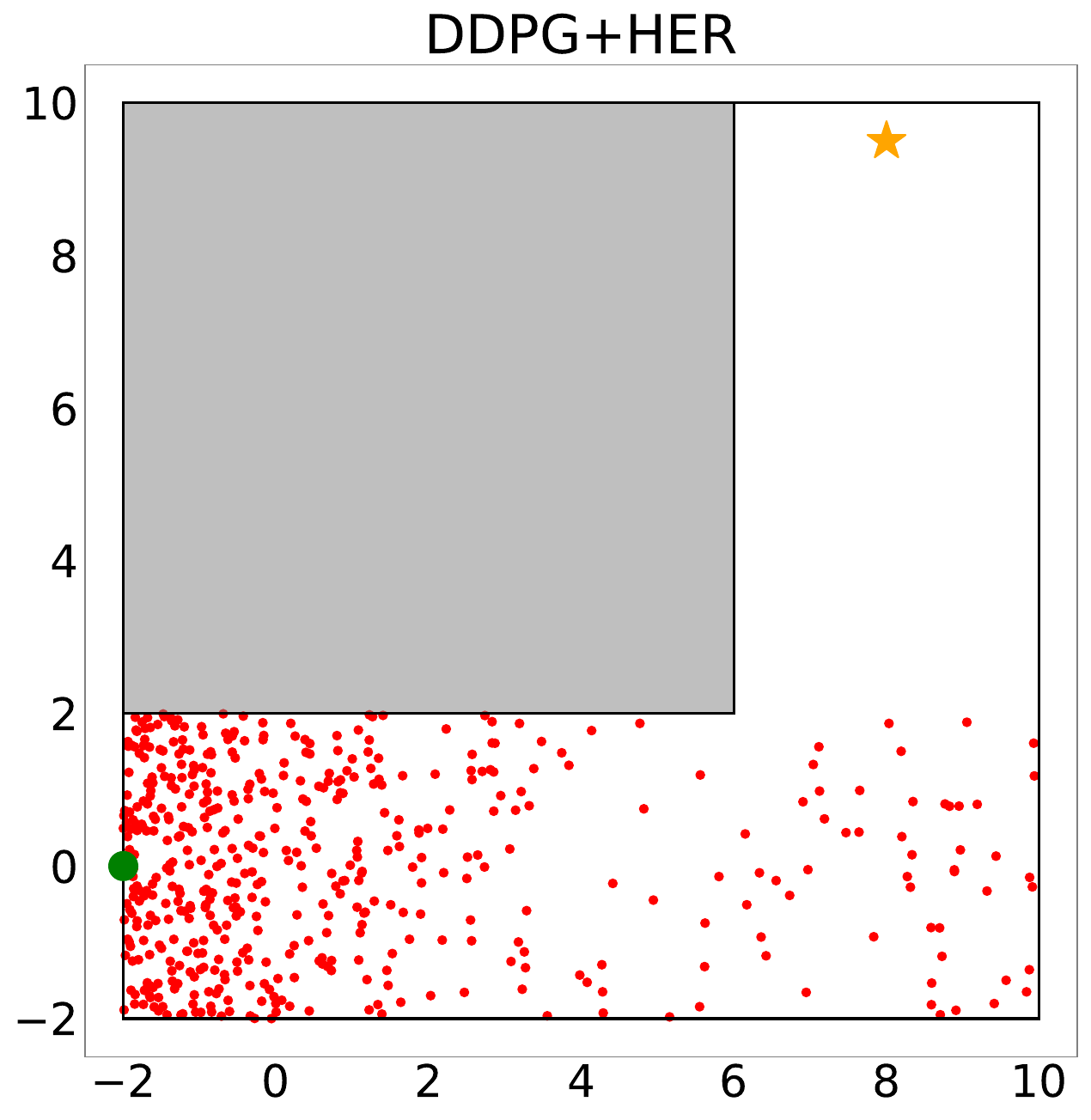}}
    \end{minipage}
    \begin{minipage}{0.22\linewidth}
        
        \centerline{\includegraphics[width=0.88\textwidth]{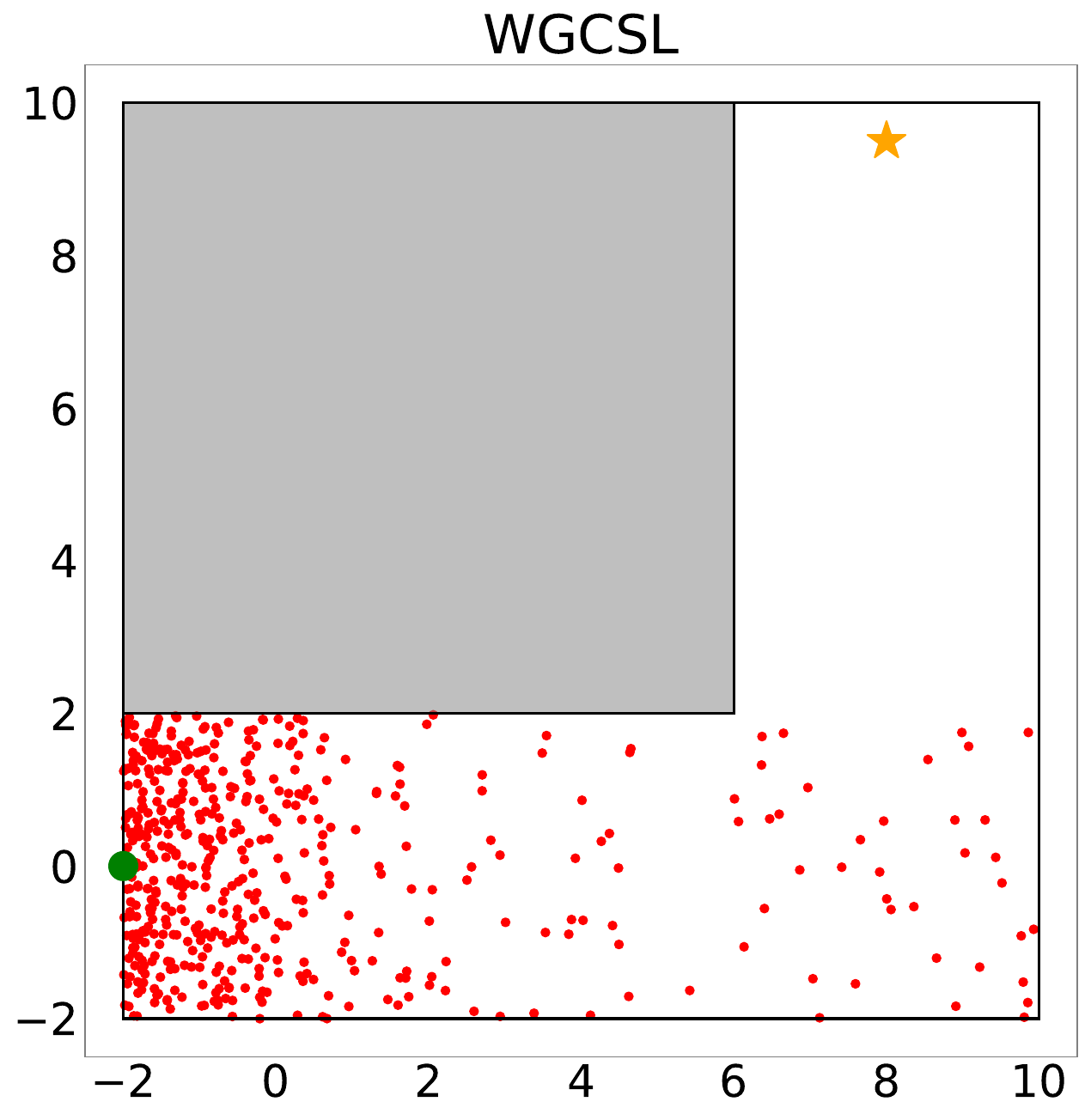}}
    \end{minipage}
    \begin{minipage}{0.22\linewidth}
        
        \centerline{\includegraphics[width=0.88\textwidth]{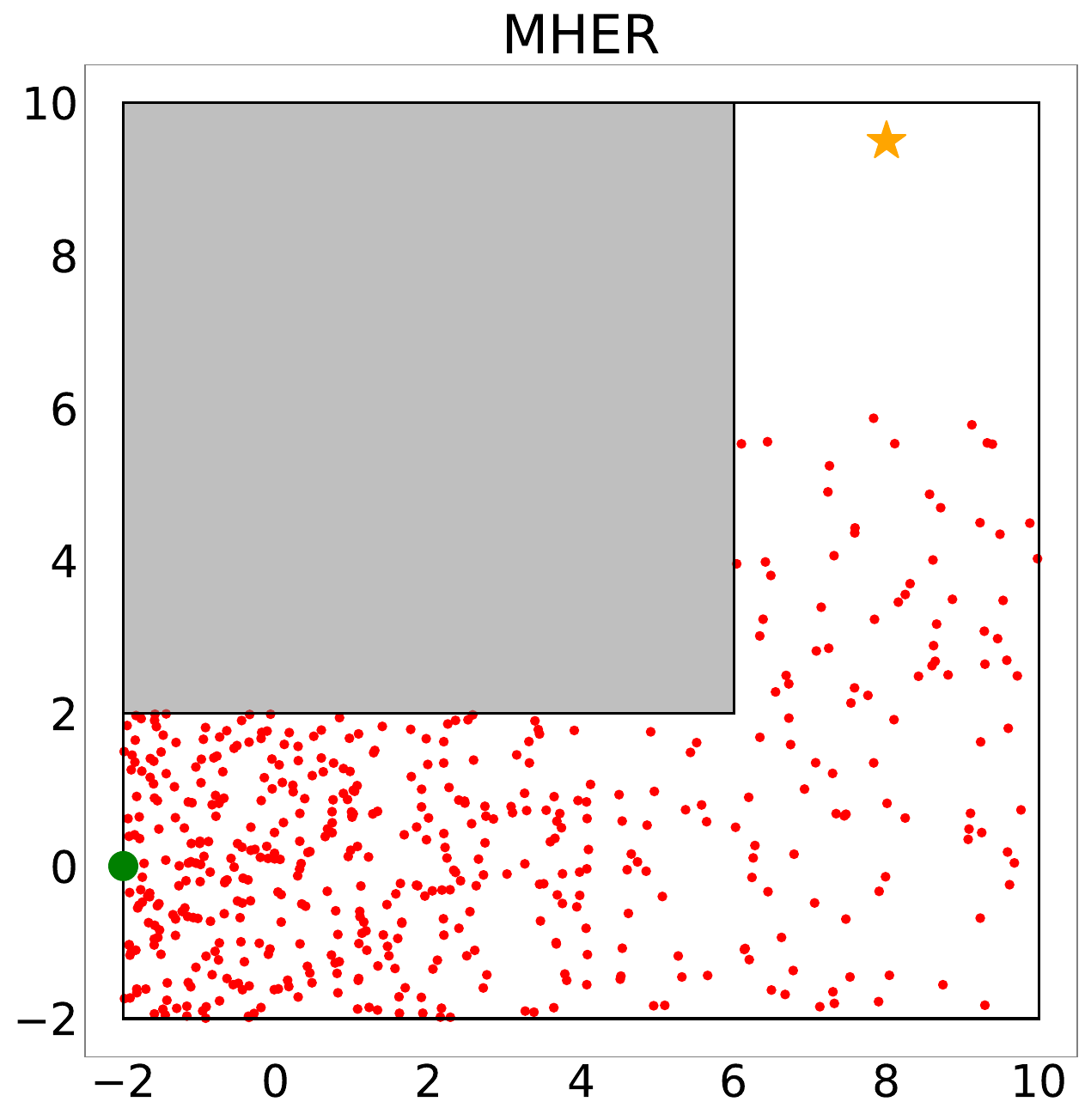}}
    \end{minipage}
    \begin{minipage}{0.22\linewidth}
        
        \centerline{\includegraphics[width=0.88\textwidth]{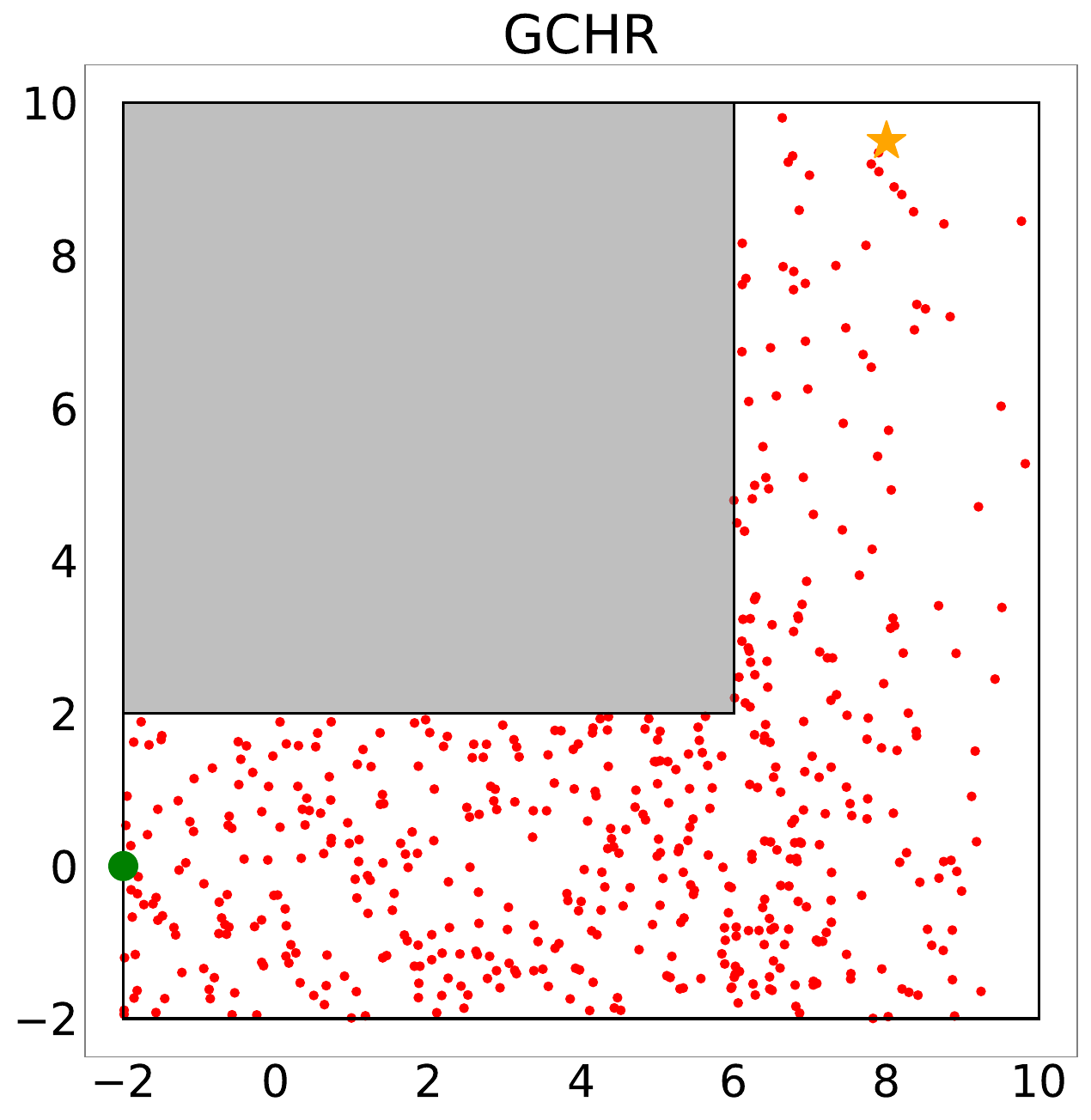}}
    \end{minipage}
    \caption{
     Visualization of the achieved goals accessed after training in the L-Antmaze 2D environment \citep{lee2022dhrl}. Only our methods reach the desired goal area in top right.
    }
    \label{fig:relabel goals}
     
\end{figure*}
\paragraph{Experiments Setup} 
We utilize the established goal-conditioned research benchmarks as detailed by \citet{plappert2018multi}, encompassing four manipulation tasks on the $Shadow-hand$ and all tasks on the $Fetch$ robot. Figure \ref{fig:multi_goal exapmle tasks} presents examples of the tasks.
We conduct a comparative analysis of our proposed method against various established GCRL algorithms. We implement the baseline algorithms in the same off-policy actor-critic framework as our method to ensure fair comparisons. All experiments are repeated using five random seeds.
We compare GCHR with various goal-conditioned RL and self-imitation methods, including \textbf{DDPG} \citep{lillicrap2015continuous}, \textbf{DDPG+HER} \citep{andrychowicz2017hindsight}, \textbf{MHER} \citep{yang2021mher}, \textbf{GCSL} \citep{ghosh2021learning}, \textbf{WGCSL} \citep{yang2022rethinking},
\textbf{GoFar} \citep{ma2022offline}, and \textbf{DWSL} \citep{hejna2023distance}. \textbf{Notably}, although GoFar and DWSL are offline GCRL methods, \citet{yang2023swapped} and \citet{hejna2023distance} indicate that they are both derived from Advantage-Weighted Regression (AWR) \citep{peng2019advantage}. Therefore, we re-implemented them in the online setting. 
We choose SAC as the base algorithm for GCHR (HGR+HSR) due to its off-policy nature and stochastic policy representation. Detailed implementations and hyperparameters are provided in Appendix \ref{ap:baseline details}.

\begin{figure*}[t]
    \centering
    \begin{minipage}{\linewidth}
    
        \centerline{\includegraphics[width=0.7\textwidth]{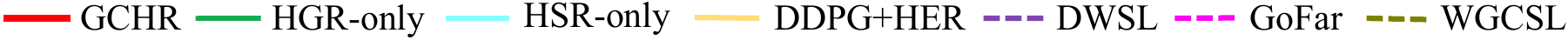}}
    \end{minipage}
    \begin{minipage}{0.24\linewidth}
        
        \centerline{\includegraphics[width=0.95\textwidth]{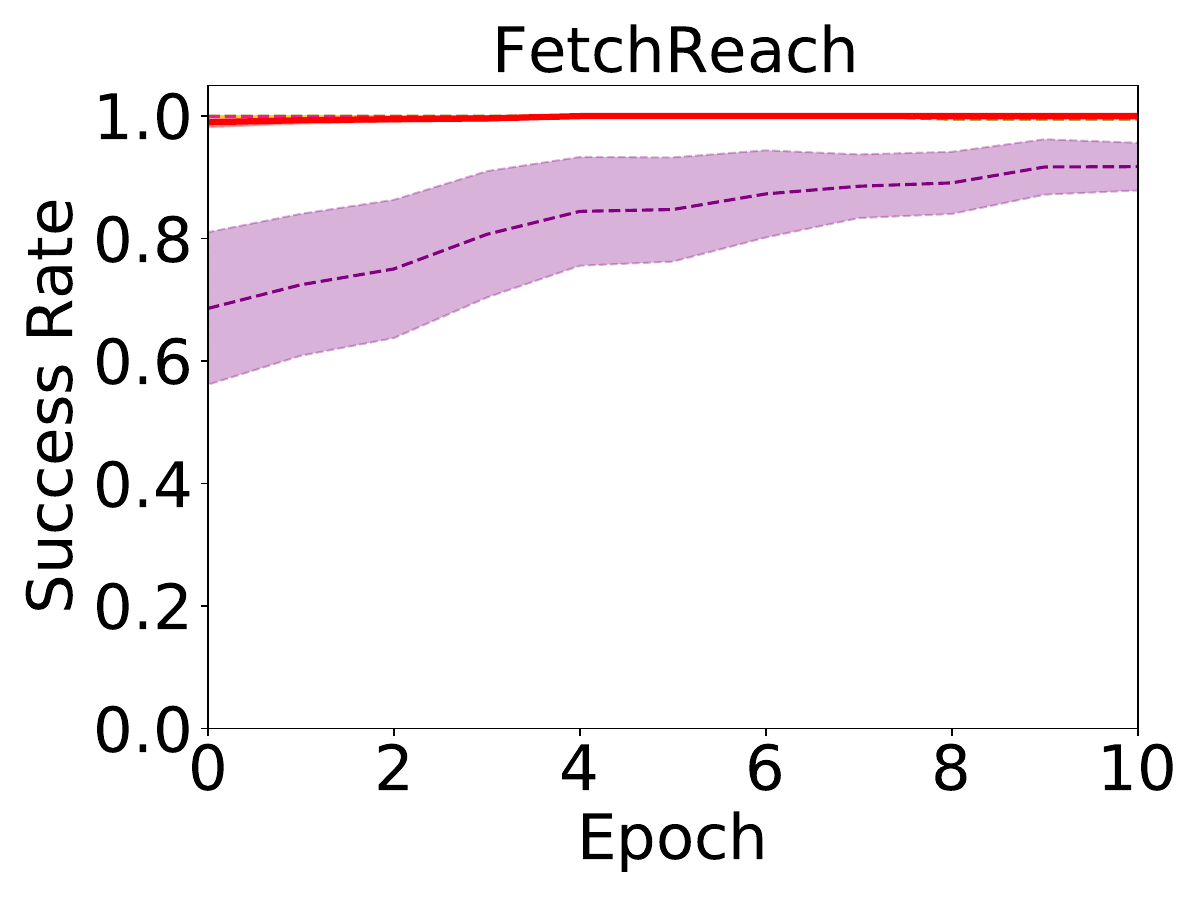}}
    \end{minipage}
    \begin{minipage}{0.24\linewidth}
        
        \centerline{\includegraphics[width=0.9\textwidth]{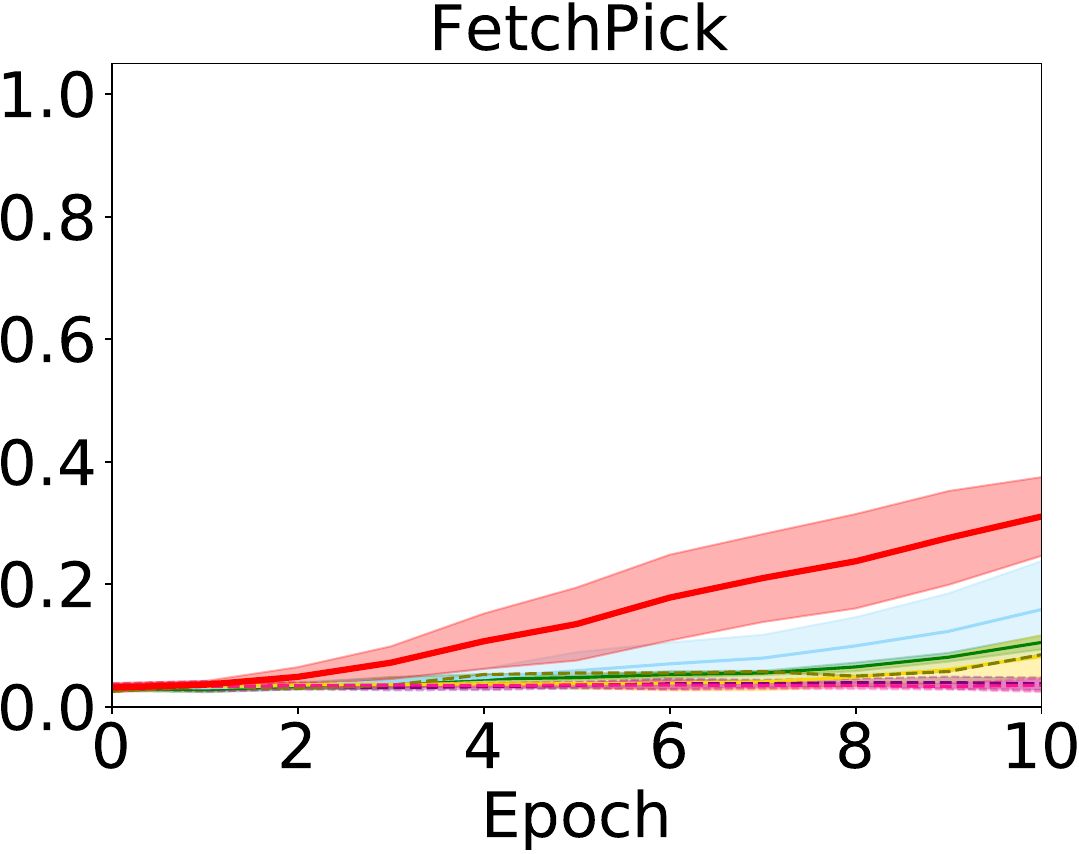}}
    \end{minipage}
    \begin{minipage}{0.24\linewidth}
        
        \centerline{\includegraphics[width=0.9\textwidth]{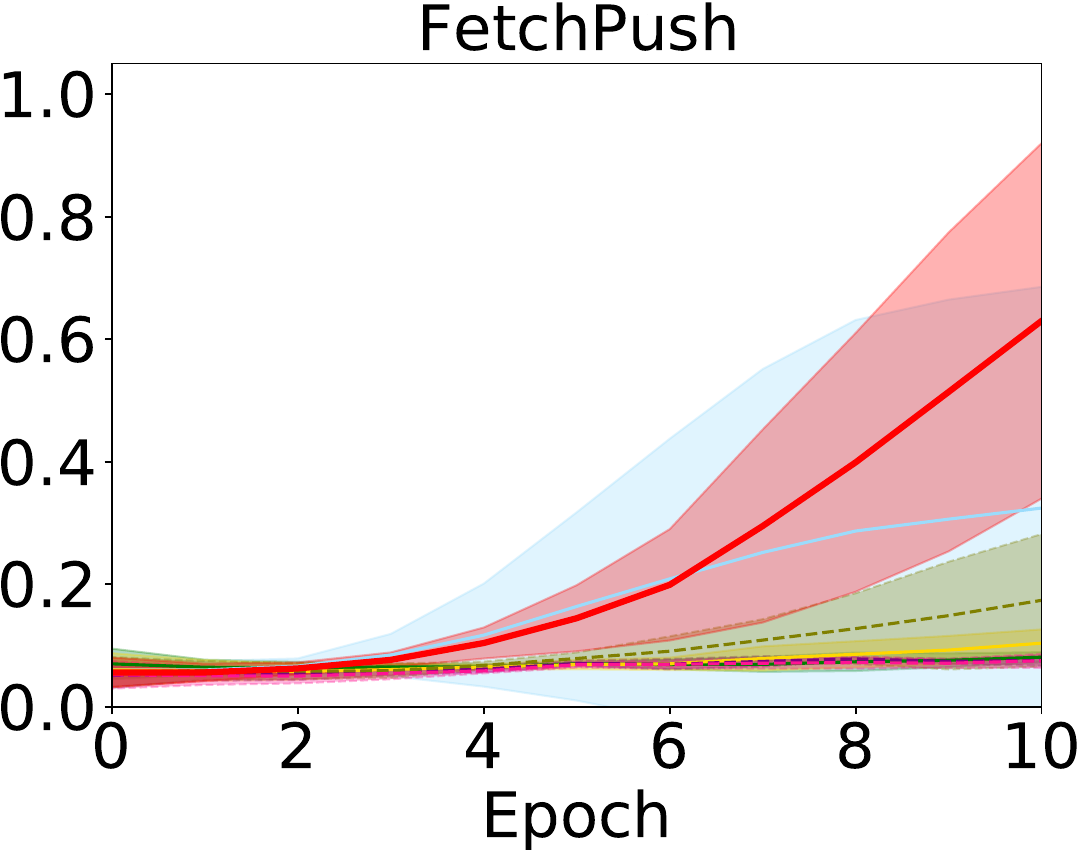}}
    \end{minipage}
    \begin{minipage}{0.24\linewidth}
        
        \centerline{\includegraphics[width=0.9\textwidth]{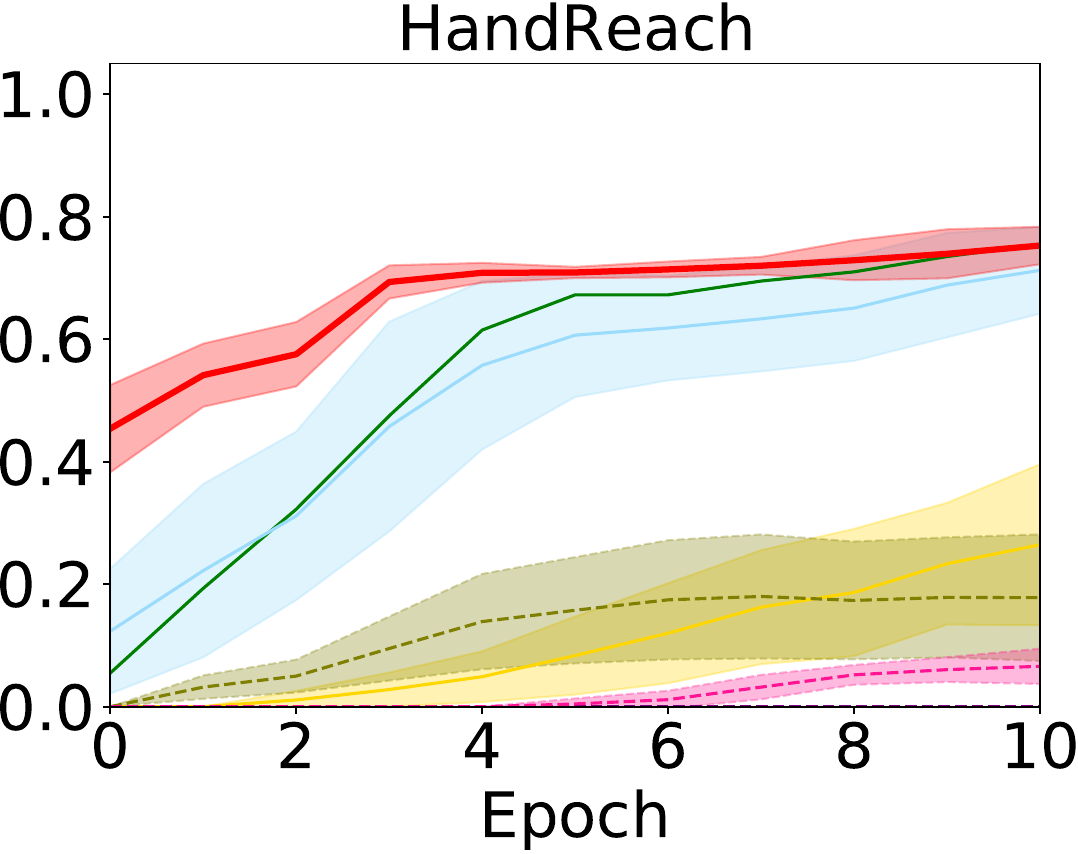}}
    \end{minipage}
    \caption{
      Ablation studies of HSR-only (No $\pi_{\text{HG}_\text{prior}}$) and HGR-only (No BC) on 4 Fetch tasks.
    }
    \label{fig:multi_goal ablation results}
    
\end{figure*}
\subsection{Results on Goal-conditioned Benchmarks} \label{sec:5.1}
As illustrated in Figure \ref{fig:multi_goal results}, GCHR performs better than all baseline methods in all environments, coupled with a faster learning speed. The results indicate that DDPG exhibit slow learning across all tasks, whereas other methods benefit from HER, showcasing its critical role in enhancing learning efficiency and handling sparse rewards in GCRL. 
\subsection{Qualitative Analysis of Exploration} \label{ap:exploration visual}
To better understand how our GCHR can encompass additional actions, we visualize the terminal goal achieved at the end of each episode during training in a L-Antmaze environment (i.e., 2D goal space) from \citep{lee2022dhrl}, as presented in Figure \ref{fig:relabel goals}. 

As shown in Figure \ref{fig:relabel goals}, it can be observed that the goal coverage of HER optimized with GSR and WGCSL is limited, and they fail to transfer to the desired goal distribution in the top-right corner. MHER enhances the transitions to the desired goal by generating new virtual trajectories. However, due to the inaccuracies in the dynamics model, the coverage of the desired goal remains limited. GCHR transitions to goals sampled from the desired goal distribution in the top-right corner. This is because in the early stage, achieved goals of GCHR are near relabeled goals. As the policy improves, they gradually approach their desired goals. The learning process of GCHR is automatically adjusted by the policy, therefore GCHR can substantially achieve more efficient curriculum learning. 
\subsection{Robust to Environmental Stochasticity}
In this experimental setup, we investigated the robustness of various algorithms in modified FetchPush
environments, which are subjected to different levels of Gaussian action noise applied to the policy action outputs.

\begin{figure}[h]
    \centering
    \includegraphics[width=0.8\linewidth]{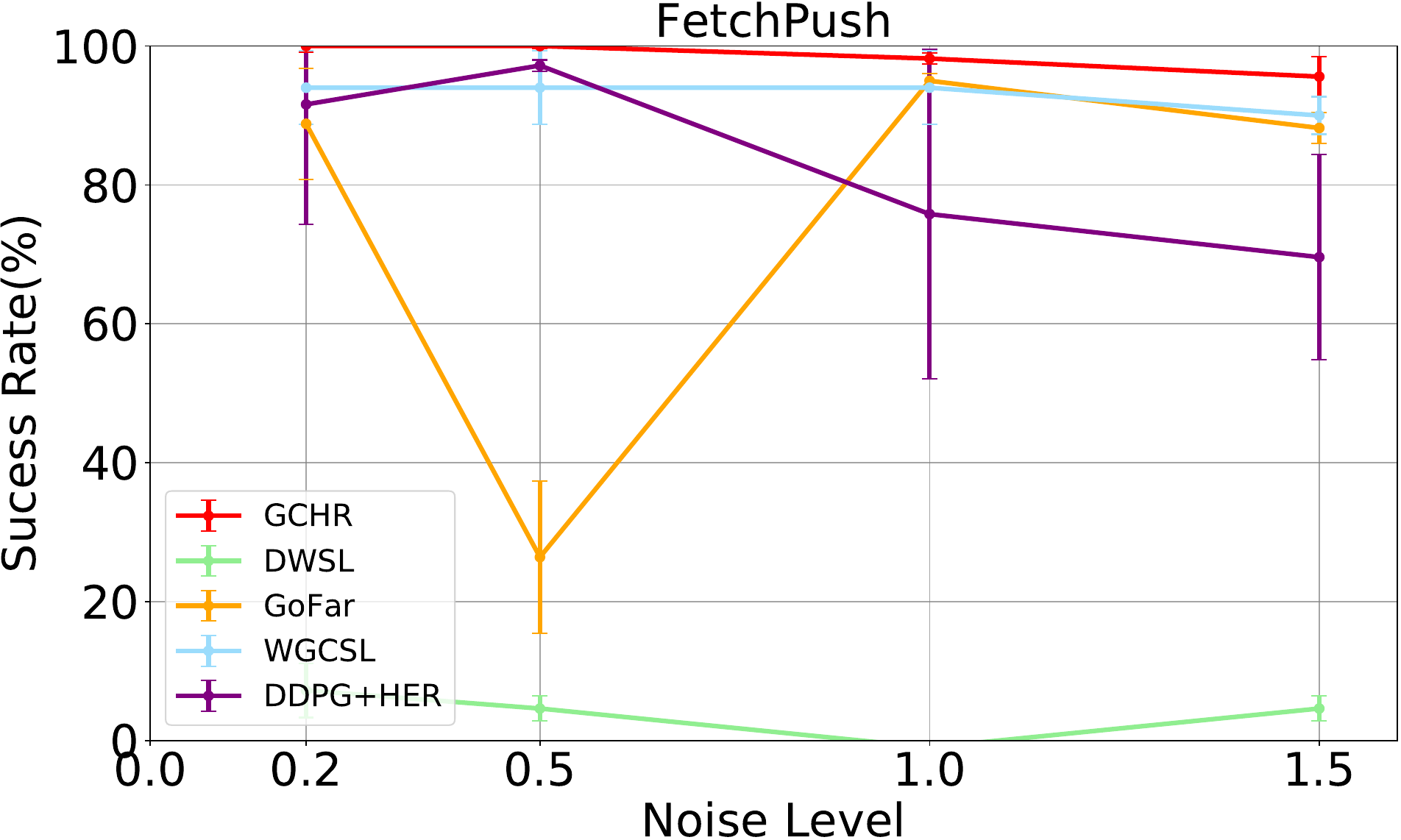}   
    \caption{\footnotesize
   Mean success rate (\%) for FetchPush task under environment stochasticity. The bars at the data points of the line graph represent the standard deviation.
   }
   \label{fig:multi_goal environment stochasticity results}
   
\end{figure}
As we see in Figure \ref{fig:multi_goal environment stochasticity results}, GCHR
is the most robust to stochasticity in the FetchPush environment, 
also outperforming baseline algorithms in terms of mean success rate under various noise levels. In particular, algorithms based solely on self-imitation are highly sensitive to noise. We suggest that the assumption of deterministic dynamics embedded in self-imitation learning methods, such as WGCSL, GoFar, and DWSL may lead to overly optimistic performance assessments in stochastic environments.
\subsection{Ablation Studies} \label{sec:6.3}
To evaluate the effectiveness of HSR and HGR at the stage of training in the GCHR framework, we conducted a series of ablation experiments comparing GCHR variants with HER. In these experiments, the value of $I$ is set to the total number of relabeled goals associated with the trajectory, and the parameter $\beta$ is set to 0.2 by default.
We conduct experiments with the following settings: 1) \textbf{GCHR}: Both HGR and HSR regularizes SAC. 2)  \textbf{HGR-only}: Only using HGR regularization for SAC. 3) \textbf{HSR-only}: Only using HSR regularization for SAC.
\begin{table}[h]
    \centering
    \scalebox{0.8}{\begin{tabular}{c|cccc}
    \toprule
    \textbf{Method}   & Reach(F) & Pick(F) & Push(F) & Reach(H) \\ 
    \midrule
    DDPG &\textbf{100.0} $\pm$ 0.0 &3.8 $\pm$ 2.5 &7.0 $\pm$ 4.2 &0.0 $\pm$ 0.0\\  
    SAC          & 99.8 $\pm$ 0.4           & 1.8 $\pm$ 2.3             & 3.2 $\pm$ 1.9             & 0.0 $\pm$ 0.0 \\
    DDPG+HER     & \textbf{100} $\pm$ 0.0   & 15.3 $\pm$ 1.2            & 10.2 $\pm$ 1.1            & 23.4 $\pm$ 5.6 \\
    SAC+HER      & \textbf{100} $\pm$ 0.0   & 10.8 $\pm$ 10.9           & 9.4 $\pm$ 4.2             & 17.6 $\pm$ 4.6 \\

    GCHR         & \textbf{100} $\pm$ 0.0   & \textbf{35.4} $\pm$ 2.9   & \textbf{60.5} $\pm$ 8.4   & \textbf{75.5} $\pm$ 0.9 \\
    \bottomrule
    \end{tabular}}
    \caption{Mean success rate (\%) for different GCRL variants. "F" indicates the "Fetch-" tasks and "H" indicates the "Hand-" task. }
    \label{tab:gchr_vs_vanilla}
\end{table}

The empirical results shown in Figure \ref{fig:multi_goal ablation results} demonstrate that HGR are more important than HSR in the GCHR framework. The GCHR method attains faster learning compared to competitive baseline DDPG+HER, while the state-of-the-art DWSL struggles to learn effectively in these tasks, with the exception of the FetchReach task. This observation implies that supervised learning approaches are suboptimal for relabeled data.
HGR leads to substantial performance enhancements. This improvement arises from the synergistic interaction between HSR and HGR in the GCHR framework, as discussed in Section \ref{sec:method}. In other words, although the dataset contains suboptimal trajectories, by relabeling them to be optimal in hindsight and treating them as expert data \citep{ghosh2021learning}, we can learn a reasonably good prior policy. This prior policy serves as a beneficial prior that can improve the learning of the final policy for achieving the desired goal.

Table \ref{tab:gchr_vs_vanilla} presents results comparing off-policy actor-critic methods. The SAC-based GCHR with our proposed hindsight regularizations significantly outperforms vanilla GCRL baselines.
In Appendix \ref{sc:beta_alpha hyperparameter}, we provide ablation studies on the parameters $\alpha$ and $\beta$.
\section{Conclusion}
In this paper, we propose a new sample-efficient goal-conditioned method termed GCHR. By
refining the hindsight policy (HSR) used for behavior regularization (HGR), GCHR progressively improves itself within the hindsight policy's
support and provably converges to the optimal policy. Both theoretical analysis and experimental results indicate that GCHR can converge to a more optimal policy and cover a broader goal distribution.

\bibliography{aaai2026}

\begin{thebibliography}{35}
\providecommand{\natexlab}[1]{#1}

\bibitem[{Andrychowicz et~al.(2017)Andrychowicz, Wolski, Ray, Schneider, Fong, Welinder, McGrew, Tobin, Pieter~Abbeel, and Zaremba}]{andrychowicz2017hindsight}
Andrychowicz, M.; Wolski, F.; Ray, A.; Schneider, J.; Fong, R.; Welinder, P.; McGrew, B.; Tobin, J.; Pieter~Abbeel, O.; and Zaremba, W. 2017.
\newblock Hindsight experience replay.
\newblock \emph{Advances in neural information processing systems}, 30.

\bibitem[{Brockman(2016)}]{brockman2016openai}
Brockman, G. 2016.
\newblock OpenAI Gym.
\newblock \emph{arXiv preprint arXiv:1606.01540}.

\bibitem[{Chane-Sane, Schmid, and Laptev(2021)}]{chane2021goal}
Chane-Sane, E.; Schmid, C.; and Laptev, I. 2021.
\newblock Goal-conditioned reinforcement learning with imagined subgoals.
\newblock In \emph{International conference on machine learning}, 1430--1440. PMLR.

\bibitem[{Eysenbach, Salakhutdinov, and Levine(2020)}]{eysenbach2020c}
Eysenbach, B.; Salakhutdinov, R.; and Levine, S. 2020.
\newblock C-learning: Learning to achieve goals via recursive classification.
\newblock \emph{arXiv preprint arXiv:2011.08909}.

\bibitem[{Eysenbach et~al.(2022)Eysenbach, Udatha, Salakhutdinov, and Levine}]{eysenbach2022imitating}
Eysenbach, B.; Udatha, S.; Salakhutdinov, R.~R.; and Levine, S. 2022.
\newblock Imitating past successes can be very suboptimal.
\newblock \emph{Advances in Neural Information Processing Systems}, 35: 6047--6059.

\bibitem[{Fang et~al.(2019)Fang, Zhou, Du, Han, and Zhang}]{fang2019curriculum}
Fang, M.; Zhou, T.; Du, Y.; Han, L.; and Zhang, Z. 2019.
\newblock Curriculum-guided hindsight experience replay.
\newblock \emph{Advances in neural information processing systems}, 32.

\bibitem[{Ghosh et~al.(2021)Ghosh, Gupta, Reddy, Fu, Devin, Eysenbach, and Levine}]{ghosh2021learning}
Ghosh, D.; Gupta, A.; Reddy, A.; Fu, J.; Devin, C.~M.; Eysenbach, B.; and Levine, S. 2021.
\newblock Learning to Reach Goals via Iterated Supervised Learning.
\newblock In \emph{International Conference on Learning Representations}.

\bibitem[{Haarnoja et~al.(2018)Haarnoja, Zhou, Abbeel, and Levine}]{haarnoja2018soft}
Haarnoja, T.; Zhou, A.; Abbeel, P.; and Levine, S. 2018.
\newblock Soft actor-critic: Off-policy maximum entropy deep reinforcement learning with a stochastic actor.
\newblock In \emph{International conference on machine learning}, 1861--1870. PMLR.

\bibitem[{He, Zhuang, and Li(2020)}]{he2020soft}
He, Q.; Zhuang, L.; and Li, H. 2020.
\newblock Soft hindsight experience replay.
\newblock \emph{arXiv preprint arXiv:2002.02089}.

\bibitem[{Hejna, Gao, and Sadigh(2023)}]{hejna2023distance}
Hejna, J.; Gao, J.; and Sadigh, D. 2023.
\newblock Distance Weighted Supervised Learning for Offline Interaction Data.
\newblock \emph{arXiv preprint arXiv:2304.13774}.

\bibitem[{Kingma(2014)}]{kingma2014adam}
Kingma, D.~P. 2014.
\newblock Adam: A method for stochastic optimization.
\newblock \emph{arXiv preprint arXiv:1412.6980}.

\bibitem[{Lee et~al.(2022)Lee, Kim, Jang, and Kim}]{lee2022dhrl}
Lee, S.; Kim, J.; Jang, I.; and Kim, H.~J. 2022.
\newblock DHRL: a graph-based approach for long-horizon and sparse hierarchical reinforcement learning.
\newblock \emph{Advances in Neural Information Processing Systems}, 35: 13668--13678.

\bibitem[{Li, Pinto, and Abbeel(2020)}]{li2020generalized}
Li, A.; Pinto, L.; and Abbeel, P. 2020.
\newblock Generalized hindsight for reinforcement learning.
\newblock \emph{Advances in neural information processing systems}, 33: 7754--7767.

\bibitem[{Li, Wang, and Tan(2023)}]{li2023self}
Li, Y.; Wang, Y.; and Tan, X. 2023.
\newblock Self-imitation guided goal-conditioned reinforcement learning.
\newblock \emph{Pattern Recognition}, 144: 109845.

\bibitem[{Li et~al.(2025)Li, Ji, Ling, and Liu}]{li2025comprehensive}
Li, Z.; Ji, Q.; Ling, X.; and Liu, Q. 2025.
\newblock A comprehensive review of multi-agent reinforcement learning in video games.
\newblock \emph{Authorea Preprints}.

\bibitem[{Lillicrap et~al.(2015)Lillicrap, Hunt, Pritzel, Heess, Erez, Tassa, Silver, and Wierstra}]{lillicrap2015continuous}
Lillicrap, T.~P.; Hunt, J.~J.; Pritzel, A.; Heess, N.; Erez, T.; Tassa, Y.; Silver, D.; and Wierstra, D. 2015.
\newblock Continuous control with deep reinforcement learning.
\newblock \emph{arXiv preprint arXiv:1509.02971}.

\bibitem[{Liu et~al.(2023)Liu, Feng, Liu, and Stone}]{liu2023metric}
Liu, B.; Feng, Y.; Liu, Q.; and Stone, P. 2023.
\newblock Metric residual network for sample efficient goal-conditioned reinforcement learning.
\newblock In \emph{Proceedings of the AAAI Conference on Artificial Intelligence}, volume~37, 8799--8806.

\bibitem[{Liu, Zhu, and Zhang(2022)}]{liu2022goal}
Liu, M.; Zhu, M.; and Zhang, W. 2022.
\newblock Goal-conditioned reinforcement learning: Problems and solutions.
\newblock \emph{arXiv preprint arXiv:2201.08299}.

\bibitem[{Ma et~al.(2022)Ma, Yan, Jayaraman, and Bastani}]{ma2022offline}
Ma, J.~Y.; Yan, J.; Jayaraman, D.; and Bastani, O. 2022.
\newblock Offline goal-conditioned reinforcement learning via $ f $-advantage regression.
\newblock \emph{Advances in Neural Information Processing Systems}, 35: 310--323.

\bibitem[{Oh et~al.(2018)Oh, Guo, Singh, and Lee}]{oh2018self}
Oh, J.; Guo, Y.; Singh, S.; and Lee, H. 2018.
\newblock Self-imitation learning.
\newblock In \emph{International conference on machine learning}, 3878--3887. PMLR.

\bibitem[{Park et~al.(2024)Park, Frans, Eysenbach, and Levine}]{park2024ogbench}
Park, S.; Frans, K.; Eysenbach, B.; and Levine, S. 2024.
\newblock Ogbench: Benchmarking offline goal-conditioned rl.
\newblock \emph{arXiv preprint arXiv:2410.20092}.

\bibitem[{Peng et~al.(2019)Peng, Kumar, Zhang, and Levine}]{peng2019advantage}
Peng, X.~B.; Kumar, A.; Zhang, G.; and Levine, S. 2019.
\newblock Advantage-weighted regression: Simple and scalable off-policy reinforcement learning.
\newblock \emph{arXiv preprint arXiv:1910.00177}.

\bibitem[{Plappert et~al.(2018{\natexlab{a}})Plappert, Andrychowicz, Ray, Mcgrew, Baker, Powell, Schneider, Tobin, Chociej, and Welinder}]{2018Multi}
Plappert, M.; Andrychowicz, M.; Ray, A.; Mcgrew, B.; Baker, B.; Powell, G.; Schneider, J.; Tobin, J.; Chociej, M.; and Welinder, P. 2018{\natexlab{a}}.
\newblock Multi-Goal Reinforcement Learning: Challenging Robotics Environments and Request for Research.

\bibitem[{Plappert et~al.(2018{\natexlab{b}})Plappert, Andrychowicz, Ray, McGrew, Baker, Powell, Schneider, Tobin, Chociej, Welinder et~al.}]{plappert2018multi}
Plappert, M.; Andrychowicz, M.; Ray, A.; McGrew, B.; Baker, B.; Powell, G.; Schneider, J.; Tobin, J.; Chociej, M.; Welinder, P.; et~al. 2018{\natexlab{b}}.
\newblock Multi-goal reinforcement learning: Challenging robotics environments and request for research.
\newblock \emph{arXiv preprint arXiv:1802.09464}.

\bibitem[{Rao et~al.(2025)Rao, Wang, Liu, Lei, and Giernacki}]{rao2025isfors}
Rao, J.; Wang, C.; Liu, M.; Lei, J.; and Giernacki, W. 2025.
\newblock ISFORS-MIX: Multi-agent reinforcement learning with Importance-Sampling-Free Off-policy learning and Regularized-Softmax Mixing network.
\newblock \emph{Knowledge-Based Systems}, 309: 112881.

\bibitem[{Roayaei~Ardakany and Afroughrh(2024)}]{roayaei2024maximize}
Roayaei~Ardakany, M.; and Afroughrh, A. 2024.
\newblock Maximize Score in stochastic match-3 games using reinforcement learning.
\newblock \emph{Signal and Data Processing}, 20(4): 129--140.

\bibitem[{Schramm et~al.(2023)Schramm, Deng, Granados, and Boularias}]{schramm2023usher}
Schramm, L.; Deng, Y.; Granados, E.; and Boularias, A. 2023.
\newblock Usher: Unbiased sampling for hindsight experience replay.
\newblock In \emph{Conference on Robot Learning}, 2073--2082. PMLR.

\bibitem[{Shinn et~al.(2024)Shinn, Cassano, Gopinath, Narasimhan, and Yao}]{shinn2024reflexion}
Shinn, N.; Cassano, F.; Gopinath, A.; Narasimhan, K.; and Yao, S. 2024.
\newblock Reflexion: Language agents with verbal reinforcement learning.
\newblock \emph{Advances in Neural Information Processing Systems}, 36.

\bibitem[{Tang(2020)}]{tang2020self}
Tang, Y. 2020.
\newblock Self-imitation learning via generalized lower bound q-learning.
\newblock \emph{Advances in neural information processing systems}, 33: 13964--13975.

\bibitem[{Uc-Cetina et~al.(2023)Uc-Cetina, Navarro-Guerrero, Martin-Gonzalez, Weber, and Wermter}]{uc2023survey}
Uc-Cetina, V.; Navarro-Guerrero, N.; Martin-Gonzalez, A.; Weber, C.; and Wermter, S. 2023.
\newblock Survey on reinforcement learning for language processing.
\newblock \emph{Artificial Intelligence Review}, 56(2): 1543--1575.

\bibitem[{Yang et~al.(2021)Yang, Fang, Han, Du, Luo, and Li}]{yang2021mher}
Yang, R.; Fang, M.; Han, L.; Du, Y.; Luo, F.; and Li, X. 2021.
\newblock Mher: Model-based hindsight experience replay.
\newblock \emph{arXiv preprint arXiv:2107.00306}.

\bibitem[{Yang et~al.(2022)Yang, Lu, Li, Sun, Fang, Du, Li, Han, and Zhang}]{yang2022rethinking}
Yang, R.; Lu, Y.; Li, W.; Sun, H.; Fang, M.; Du, Y.; Li, X.; Han, L.; and Zhang, C. 2022.
\newblock Rethinking Goal-conditioned Supervised Learning and Its Connection to Offline RL.
\newblock \emph{arXiv preprint arXiv:2202.04478}.

\bibitem[{Yang et~al.(2023)Yang, Wang, Cai, Pajarinen, and K{\"a}m{\"a}r{\"a}inen}]{yang2023swapped}
Yang, W.; Wang, H.; Cai, D.; Pajarinen, J.; and K{\"a}m{\"a}r{\"a}inen, J.-K. 2023.
\newblock Swapped goal-conditioned offline reinforcement learning.
\newblock \emph{arXiv preprint arXiv:2302.08865}.

\bibitem[{Zheng et~al.(2024{\natexlab{a}})Zheng, Eysenbach, Walke, Yin, Fang, Salakhutdinov, and Levine}]{zheng2024stabilizing}
Zheng, C.; Eysenbach, B.; Walke, H.~R.; Yin, P.; Fang, K.; Salakhutdinov, R.; and Levine, S. 2024{\natexlab{a}}.
\newblock Stabilizing Contrastive {RL}: Techniques for Robotic Goal Reaching from Offline Data.
\newblock In \emph{The Twelfth International Conference on Learning Representations}.

\bibitem[{Zheng et~al.(2024{\natexlab{b}})Zheng, Bai, Yang, and Wang}]{zheng2024does}
Zheng, S.; Bai, C.; Yang, Z.; and Wang, Z. 2024{\natexlab{b}}.
\newblock How does goal relabeling improve sample efficiency?
\newblock In \emph{Forty-first International Conference on Machine Learning}.

\end{thebibliography}

\setcounter{tocdepth}{-1}

\clearpage
\onecolumn
\appendix
\setlength{\parindent}{0pt}
\renewcommand{\contentsname}{Contents of Appendix}
\renewcommand{\thesection}{\Alph{section}}

\section{GCHR Algorithm}\label{ap:GCHR algorithm}
\begin{algorithm}[h]   
   \caption{Goal-Conditioned Hindsight Regularization (GCHR)} 
   \label{alg:GCHR}
   \begin{algorithmic}[1]  
      \STATE \textbf{Initialize:} replay buffer $\mathcal{B}$, policy $\pi_{\theta}$, Q-function $Q_\psi$
      \STATE \textbf{Initialize:} target networks $\bar{\theta} \leftarrow \theta$, $\bar{\psi} \leftarrow \psi$
      \STATE \textbf{Initialize:} delayed policy $\pi'_{\phi} \leftarrow \pi_{\theta}$ for HGR prior
      \STATE \textbf{Hyperparameters:} $\alpha$ (HSR weight), $\beta$ (HGR weight), $K$ (hindsight goals)
      \WHILE{not converged}
        \STATE Sample goal $g \sim p(g)$ and initial state $s_0$
        \STATE Collect trajectory $\tau = (s_0, a_0, \ldots, s_T)$ with $\pi_\theta(\cdot|s_t, g)$
        \STATE Store trajectory in replay buffer $\mathcal{B}$
        \STATE Create hindsight relabeled buffer $\mathcal{B}_r$ from $\tau$ using HER
        \FOR{$i = 1, \ldots, I$}
            \STATE \textbf{// Standard RL Update}
            \STATE Sample minibatch $(s_t, a_t, s_{t+1}, g) \sim \mathcal{B} \cup \mathcal{B}_r$
            \STATE Compute TD target: $y_t = r(s_t, g) + \gamma Q_{\bar{\psi}}(s_{t+1}, \pi_{\bar{\theta}}(s_{t+1}, g), g)$
            \STATE Update critic: $\psi \leftarrow \psi - \nabla_\psi \mathbb{E}\left[(y_t - Q_\psi(s_t, a_t, g))^2\right]$
            
            \STATE  \textcolor{orange}{\textbf{// Hindsight Self-imitation Regularization (HSR)}}
            \STATE Sample relabeled transition $(s_t, a_t, g'_t) \sim \mathcal{B}_r$
            \STATE Compute HSR loss: $\mathcal{L}_{\text{HSR}} = -\log \pi_\theta(a_t | s_t, g'_t)$
            
            \STATE \textcolor{orange}{ \textbf{// Hindsight Goal Regularization (HGR)}}
            \STATE Sample $(s, g) \sim \mathcal{B}$ and hindsight goals $\{g'_k\}_{k=1}^K \sim \mathcal{G}_H(\tau)$
            \STATE Construct prior: $\pi_{\text{HG-prior}}(a|s,g) = \frac{1}{K}\sum_{k=1}^K \pi_{\bar{\theta}}(a|s,g'_k)$
            \STATE Compute HGR loss: $\mathcal{L}_{\text{HGR}} = D_{\text{KL}}(\pi_{\text{HG-prior}}(\cdot|s,g) \,\|\, \pi_\theta(\cdot|s,g))$
            
            \STATE \textbf{// Policy Update}
            \STATE Update actor: $\theta \leftarrow \theta + \nabla_\theta \left[\mathbb{E}_{a \sim \pi_\theta}[Q_\psi(s,a,g)] - \alpha \mathcal{L}_{\text{HSR}} - \beta \mathcal{L}_{\text{HGR}}\right]$
        \ENDFOR
        \STATE \textbf{// Update Target Networks}
        \STATE $\bar{\theta} \leftarrow \tau_{\text{soft}} \theta + (1-\tau_{\text{soft}}) \bar{\theta}$
        \STATE $\bar{\psi} \leftarrow \tau_{\text{soft}} \psi + (1-\tau_{\text{soft}}) \bar{\psi}$
        \STATE \textbf{// Update Delayed Policy for HGR}
        \STATE Every $\tau_{\text{delay}}$ steps: $\phi \leftarrow \theta$
      \ENDWHILE
   \end{algorithmic} 
\end{algorithm}

\section{Theoretical Analysis}

\subsection{Proof of Theorem~\ref{thm:coverage}}\label{ap:proof1}
\begin{proof} Let $a \in \mathcal{A}_{\text{HSR}}(s,g)$. By definition, there exists a trajectory $\tau$ where taking action $a$ from state $s$ eventually leads to goal $g$. Since $g$ was achieved, we have $g \in \mathcal{G}_H(\tau)$. As the delayed policy $\pi'$ was trained on this trajectory, it learned that action $a$ from state $s$ can lead to $g$, hence $\pi'(a|s,g) > 0$. This gives us:
\begin{equation}
 a \in \{a : \pi'(a|s,g) > 0\} \subseteq \mathcal{A}_{\text{HGR}}(s,g)   
\end{equation}
For the particular case, when $\mathcal{A}_{\text{HSR}}(s,g) = \emptyset$ (state $s$ has never reached $g$), HGR can still have non-empty action support. If any trajectory $\tau$ exists where state $s$ reached some goal $g' \neq g$, then any action $a$ with $\pi'(a|s,g') > 0$ belongs to $\mathcal{A}_{\text{HGR}}(s,g)$.
\end{proof}
\subsection{Proof of Theorem~\ref{thm:monotonic}}\label{ap:proof_monotonic}
\begin{proof}
Recall that the via-goal value function is:
\begin{equation}
V_{\text{via}}^{\pi}(s,g;g') = p^{\pi}(g'|s) \sum_{s' \in S_{g'}} \tilde{d}^{\pi}(s'|s,g') \cdot V^{\pi}(s',g)  
\end{equation}
Since $\pi'^{(k)} = \pi^{(k-\tau_{\text{delay}})}$ and we have policy improvement at each iteration, it follows that:
\begin{equation}
V^{\pi'^{(k)}}(s,g) \geq V^{\pi'^{(k-1)}}(s,g) \quad \forall s,g  
\end{equation}
By Assumption~\ref{asmp:internal} part (ii), for any $s_1, s_2 \in S_{g'}$:
\begin{equation}
\bigl|V^{\pi}(s_1, g) - V^{\pi}(s_2, g)\bigr| < \delta
\end{equation}

This implies that $V^{\pi}(s', g)$ is approximately constant for all $s' \in S_{g'}$. Let us denote this constant as $C^{\pi'^{(k)}}_{g,g'}$ for policy $\pi'^{(k)}$. Then:

\begin{equation}
\sum_{s' \in S_{g'}} \tilde{d}^{\pi'^{(k)}}(s'|s,g') V^{\pi'^{(k)}}(s',g) \approx C^{\pi'^{(k)}}_{g,g'} \sum_{s' \in S_{g'}} \tilde{d}^{\pi'^{(k)}}(s'|s,g') = C^{\pi'^{(k)}}_{g,g'}
\end{equation}

where the last equality holds because $\tilde{d}^{\pi'^{(k)}}(s'|s,g')$ is a normalized distribution over $S_{g'}$.

Therefore, the via-goal value function becomes:
\begin{equation}
V_{\text{via}}^{\pi'^{(k)}}(s,g;g') \approx p^{\pi'^{(k)}}(g'|s) \cdot C^{\pi'^{(k)}}_{g,g'}
\end{equation}

Similarly for the previous iteration:
\begin{equation}
V_{\text{via}}^{\pi'^{(k-1)}}(s,g;g') \approx p^{\pi'^{(k-1)}}(g'|s) \cdot C^{\pi'^{(k-1)}}_{g,g'}
\end{equation}

From policy improvement, we have:
\begin{enumerate}
\item $C^{\pi'^{(k)}}_{g,g'} \geq C^{\pi'^{(k-1)}}_{g,g'}$ (since $V^{\pi'^{(k)}}(s',g) \geq V^{\pi'^{(k-1)}}(s',g)$ for any $s' \in S_{g'}$)
\item $p^{\pi'^{(k)}}(g'|s) \geq p^{\pi'^{(k-1)}}(g'|s)$ (from the relationship $V^{\pi}(s,g') = \frac{1}{1-\gamma} p^{\pi}(g'|s,g')$)
\end{enumerate}

Since both terms are monotonically improving and non-negative:
\begin{equation}
p^{\pi'^{(k)}}(g'|s) \cdot C^{\pi'^{(k)}}_{g,g'} \geq p^{\pi'^{(k-1)}}(g'|s) \cdot C^{\pi'^{(k-1)}}_{g,g'}
\end{equation}

As $\delta$ can be made arbitrarily small by the assumption, we conclude:
\begin{equation}
V_{\text{via}}^{\pi'^{(k)}}(s,g;g') \geq V_{\text{via}}^{\pi'^{(k-1)}}(s,g;g')
\end{equation}
\end{proof}

\section{Additional Results}
\label{appendix:additional-results}
This section evaluates the resilience of GCHR across several factors, including the hyperparameter $\beta$ and $\alpha$, the number of hindsight goals, sample efficiency, error bars of mean performance, and the relabeling ratio. These details are provided below. 
\begin{figure*}[h]
    \begin{minipage}{\linewidth}
        
        \centerline{\includegraphics[width=0.8\textwidth]{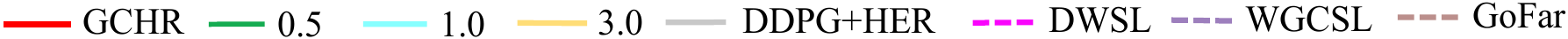}}
    \end{minipage}
    \begin{minipage}{0.245\linewidth}
        
        \centerline{\includegraphics[width=\textwidth]{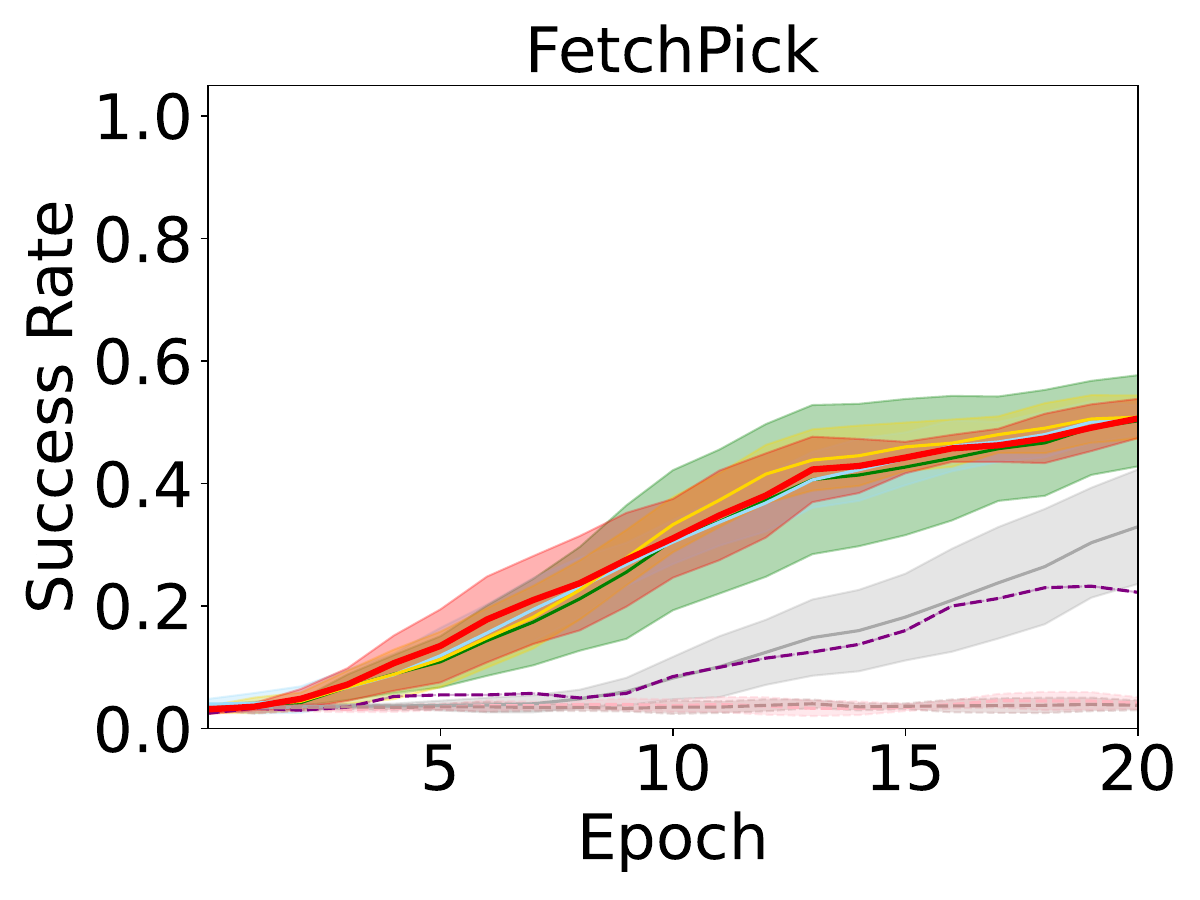}}
    \end{minipage}
    \begin{minipage}{0.245\linewidth}
        
        \centerline{\includegraphics[width=0.96\textwidth]{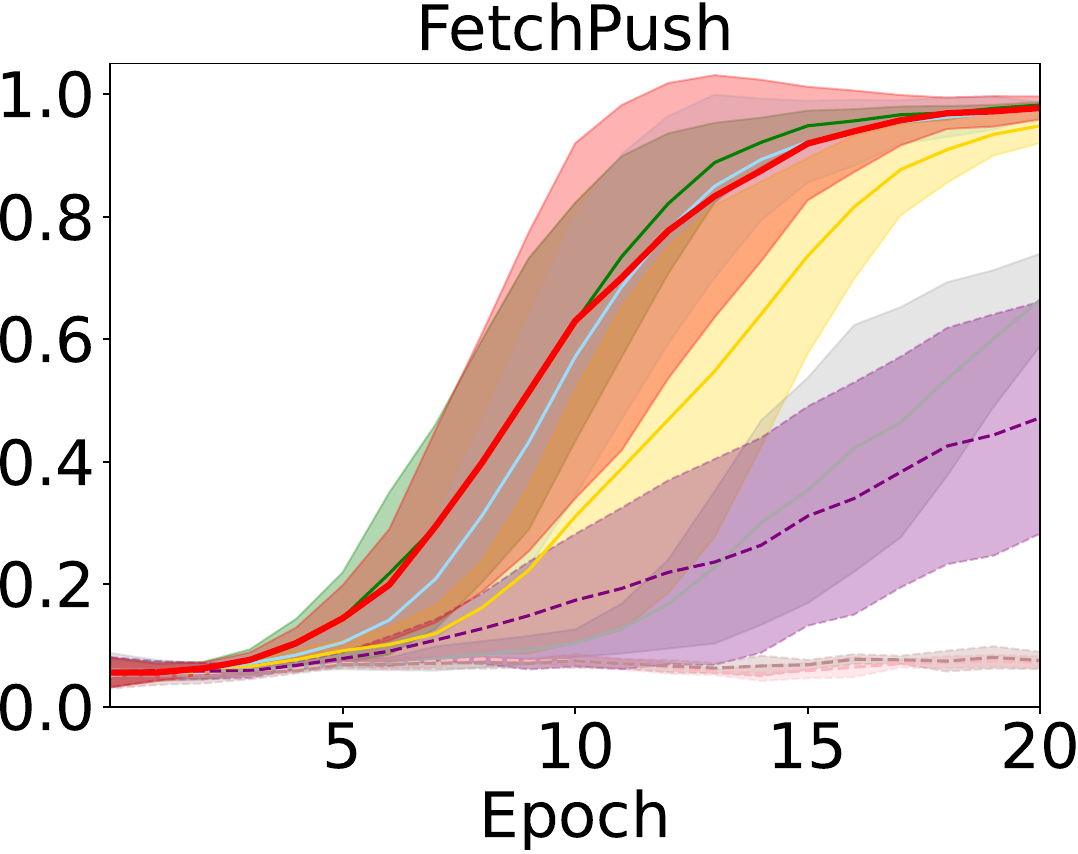}}
    \end{minipage}
    \begin{minipage}{0.245\linewidth}
        
        \centerline{\includegraphics[width=0.96\textwidth]{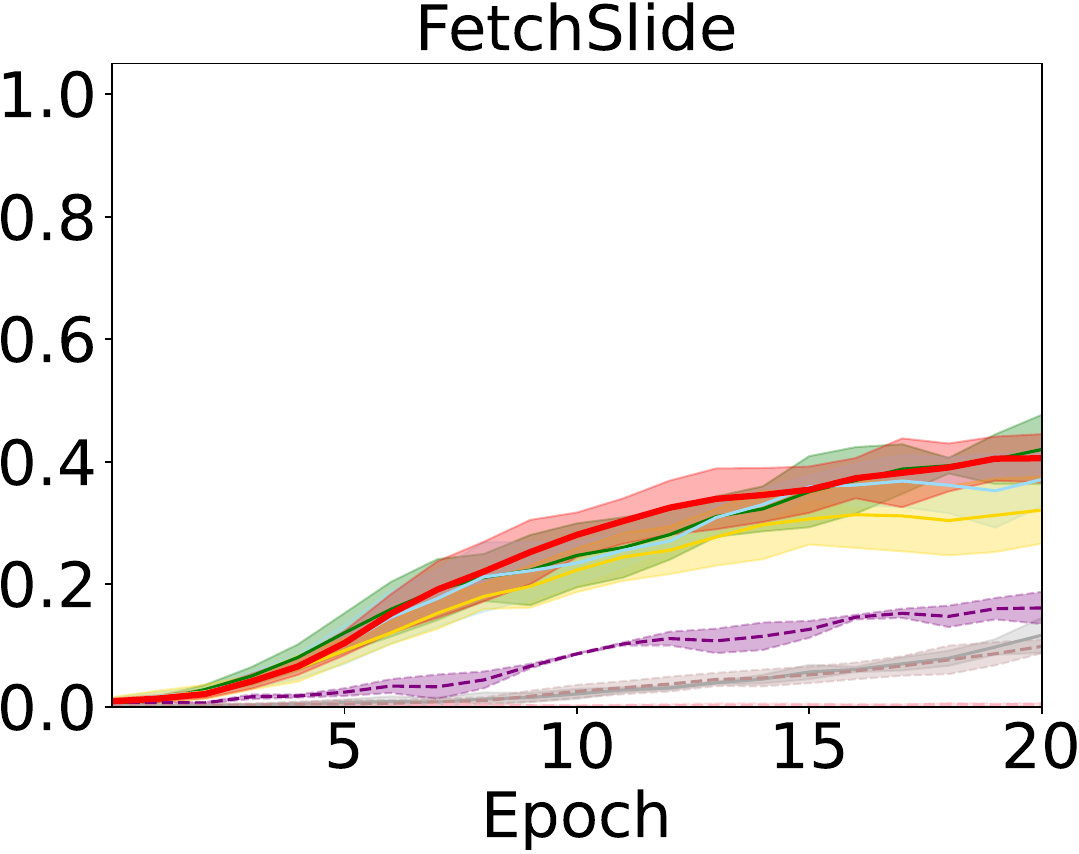}}
    \end{minipage}
    \begin{minipage}{0.245\linewidth}
        
        \centerline{\includegraphics[width=0.96\textwidth]{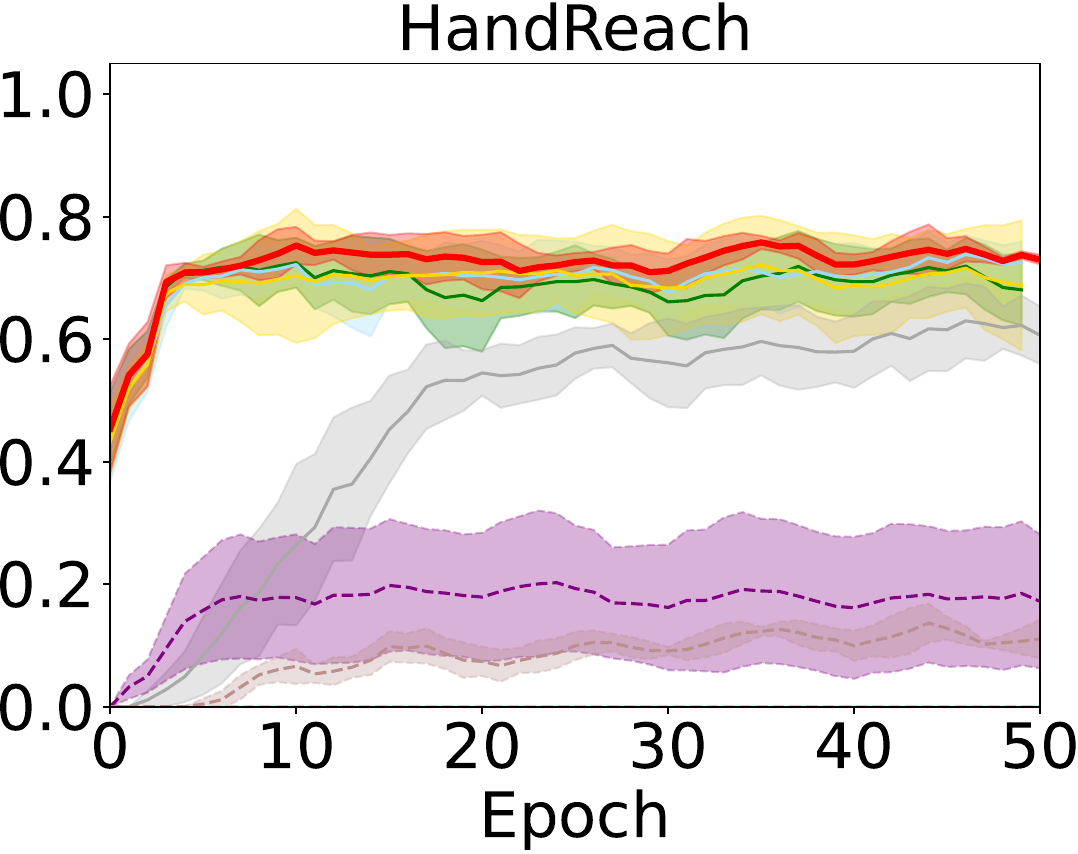}}
    \end{minipage}
    \caption{
     Hyperparameter $\beta$ ablation studies in such goal-conditioned tasks.
    }
    \label{fig:multi_goal parameter results}
    
\end{figure*}
\subsection{The Impact of Hyperparameter $\beta$ and $\alpha$} \label{sc:beta_alpha hyperparameter}
Since the addition of KL regularization term in the policy improvement stage of our method (as shown in Section \ref{sec:method}), this section explores the influence of the 
balancing parameter $\beta$. We evaluate $\beta$ values from the set $\{0.2,0.5,1.0,3.0\}$ and compare the results against competitive HER-based algorithms such as WGCSL and DDPG+HER, as shown in Figure \ref{fig:multi_goal parameter results}. The findings in Figure \ref{fig:multi_goal parameter results} reveal that GCHR consistently delivers superior performance over the other algorithms, regardless of the 
$\beta$ parameter variation. This demonstrates that our method maintains robustness and is not significantly affected by changes in the $\beta$ parameter.
\begin{table}[h]
    \centering
    \begin{tabular}{l|llllll}
    \toprule
    \textbf{$\alpha$}   & FetchPush & HandReach \\ 
    \midrule
    0.2         & 96.4 $\pm$ 1.6              &70.6 $\pm$ 1.9             \\
    0.5         &98.4 $\pm$ 1.5               &72.3 $\pm$ 1.8              \\
    1.0         &\textbf{99.5} $\pm$ 3.1              &\textbf{75.4} $\pm$ 4.3              \\
    3          &97.6 $\pm$ 2.5               &73.2 $\pm$ 2.6        \\
    \bottomrule
    \end{tabular}
    \caption{Mean success rate (\%) for different $\alpha$ after training.}
    \label{tab:wrap}
\end{table}
Similar to the parameter $\beta$, we conducted ablation experiments to evaluate the selection of the parameter $\alpha$. The results are presented in Table \ref{tab:wrap}, and performance increases when $\alpha \leq 1$ and decreases
when $\alpha > 1$. Therefore we choose $1$ as our default parameter. 
\subsection{The Impact Number of Hindsight Goals $K$} \label{ap:subgoal number}
In our approach, hindsight goals play a pivotal role, and thus, apart from their selection, investigating the optimal quantity of hindsight goals is imperative. Of course, we could also consider not selecting all hindsight goals . We systematically vary the proportion (i.e., $K/H$, where $H$ is the maximum length of $\tau^{g^{\prime}}$) of hindsight goals selected from $\tau^{g^{\prime}}$ relative to the total trajectory goals, and benchmark these against competitive algorithms such as WGCSL, DDPG+HER, GoFar, and DWSL. We evaluate algorithmic performance across four different hindsight goal proportions $\{20\%, 50\%, 90\%, 100\%\}$.

\begin{figure*}[h] 
    \begin{minipage}{\linewidth}
        
        \centerline{\includegraphics[width=0.8\textwidth]{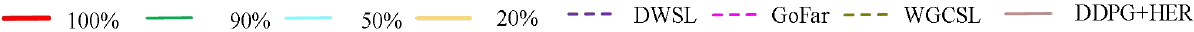}}
    \end{minipage}
 
    \begin{minipage}{0.245\linewidth}
        
        \centerline{\includegraphics[width=\textwidth]{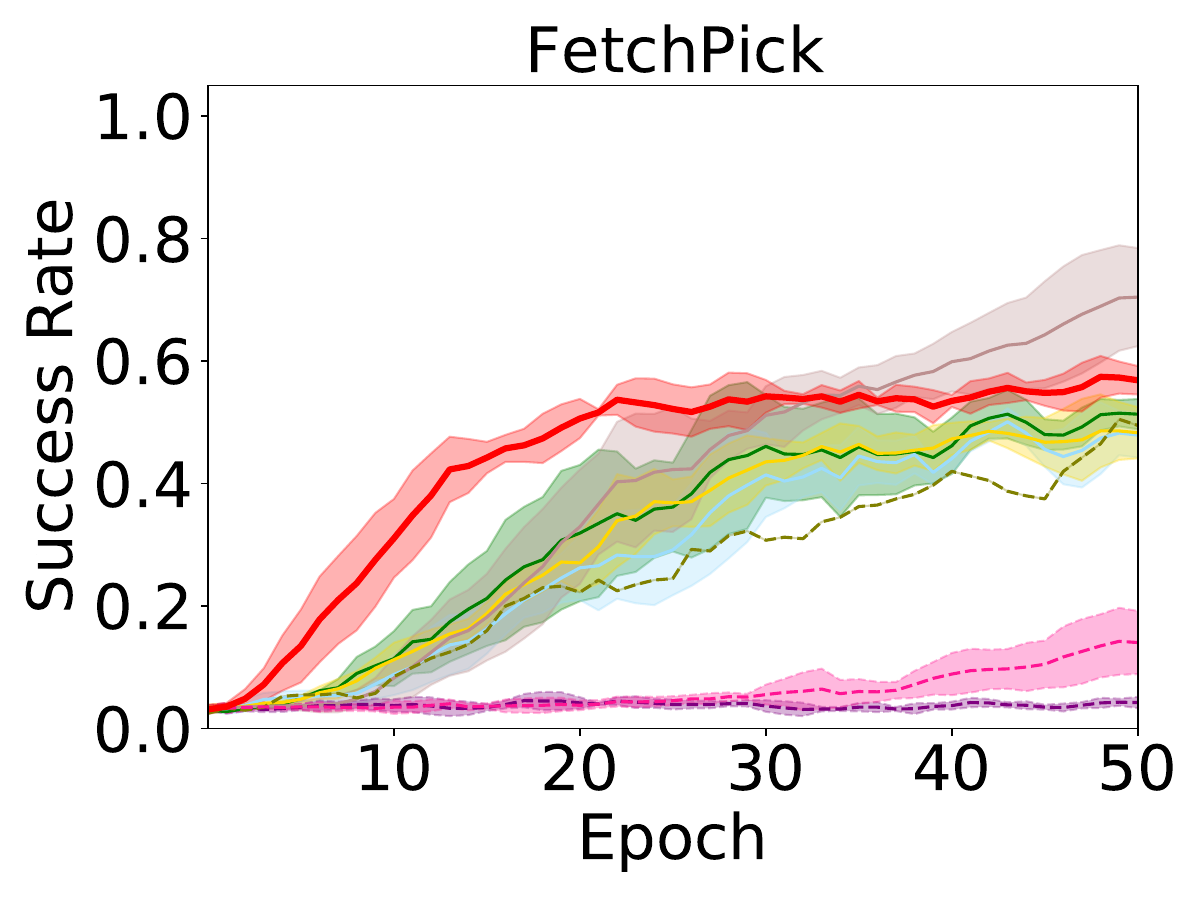}}
        
    \end{minipage}
    \begin{minipage}{0.245\linewidth}
        
        \centerline{\includegraphics[width=0.96\textwidth]{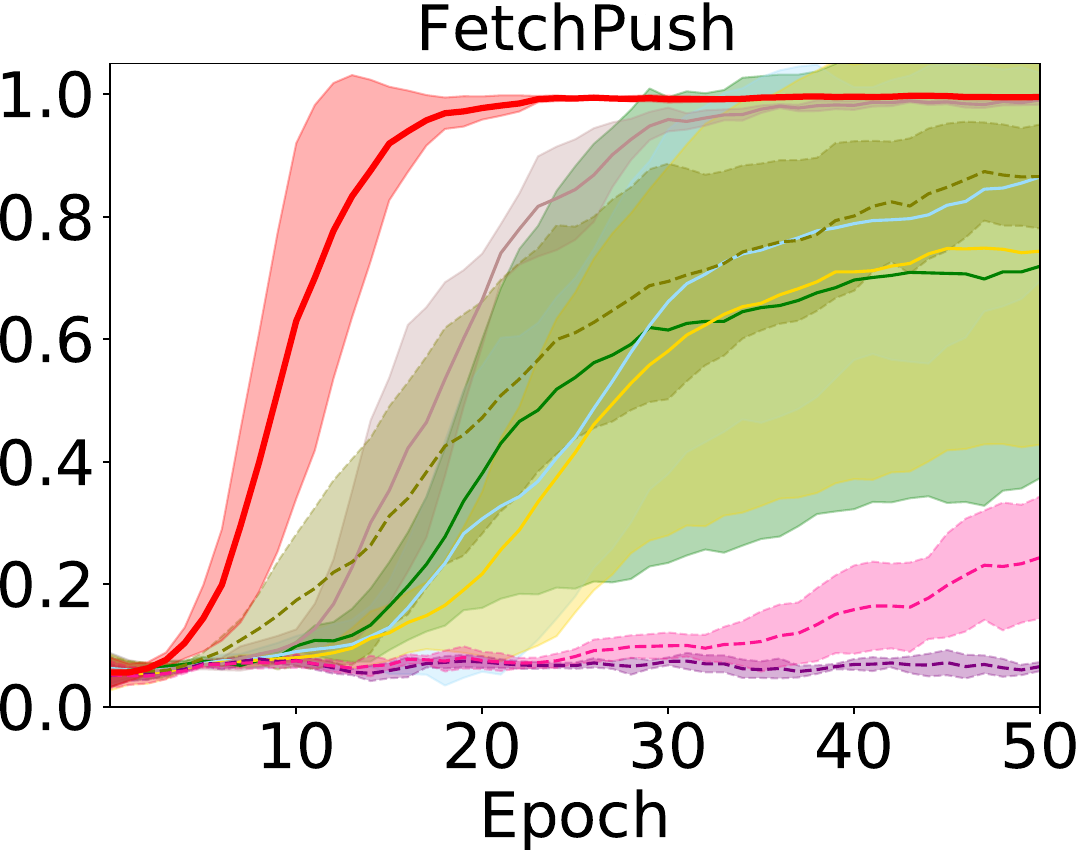}}
        
    \end{minipage}
    \begin{minipage}{0.245\linewidth}
        
        \centerline{\includegraphics[width=0.96\textwidth]{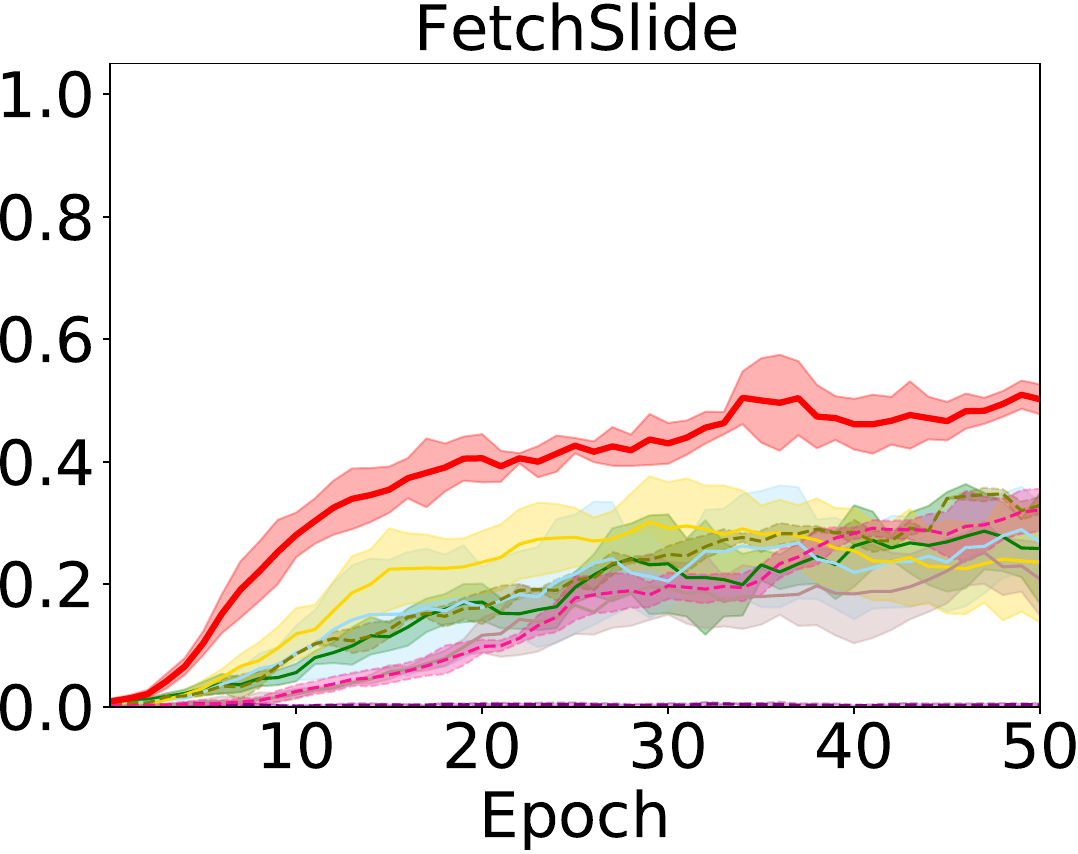}}
      
    \end{minipage}
    \begin{minipage}{0.245\linewidth}
        
        \centerline{\includegraphics[width=0.96\textwidth]{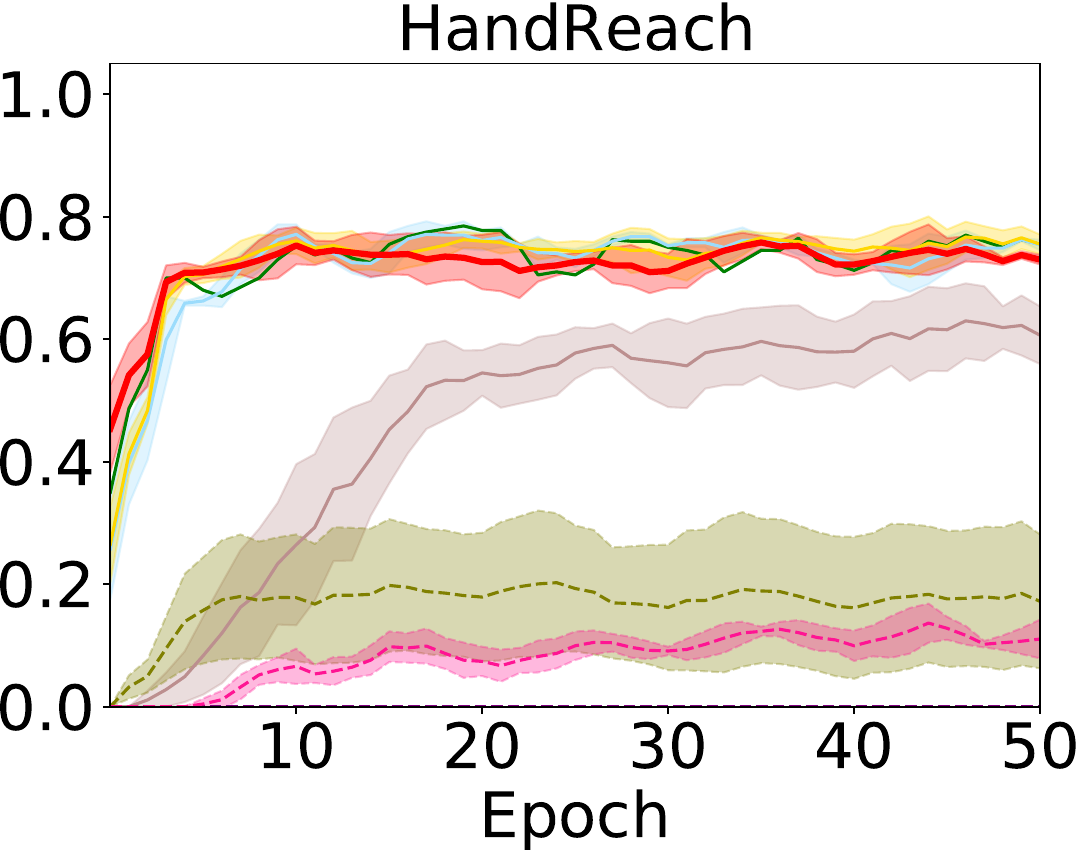}}
      
    \end{minipage}
    \caption{Relabeled goal number ablation studies in some goal-conditioned tasks. Results are averaged over 5 random seeds and the shaded region represents the standard deviation. Different percentages represent the ratio of sampled relabeled goals to the total relabeled goals within that trajectory.}
    \label{fig:multi_goal subgoalsparameter results}
\end{figure*}
Analysis presented in Figure \ref{fig:multi_goal subgoalsparameter results} demonstrates that GCHR consistently surpasses the performance of the aforementioned algorithms, regardless of the percentage of subgoals employed. This implies that selecting relabeled goals as a distribution of sub-goals is advantageous. Additionally, incorporating a greater number of relabeled goals may enhance the accuracy of the KL divergence estimation. This finding highlights the robustness of our method in response to variations in the quantity of subgoals utilized.

\subsection{Sample Efficiency}
\begin{figure}[h]
    \begin{minipage}{\linewidth}
        
        \centerline{\includegraphics[width=\textwidth]{results_pdf/results1.pdf}}
    \end{minipage}
    \begin{minipage}{0.459\linewidth}
        
        \centerline{\includegraphics[width=\textwidth]{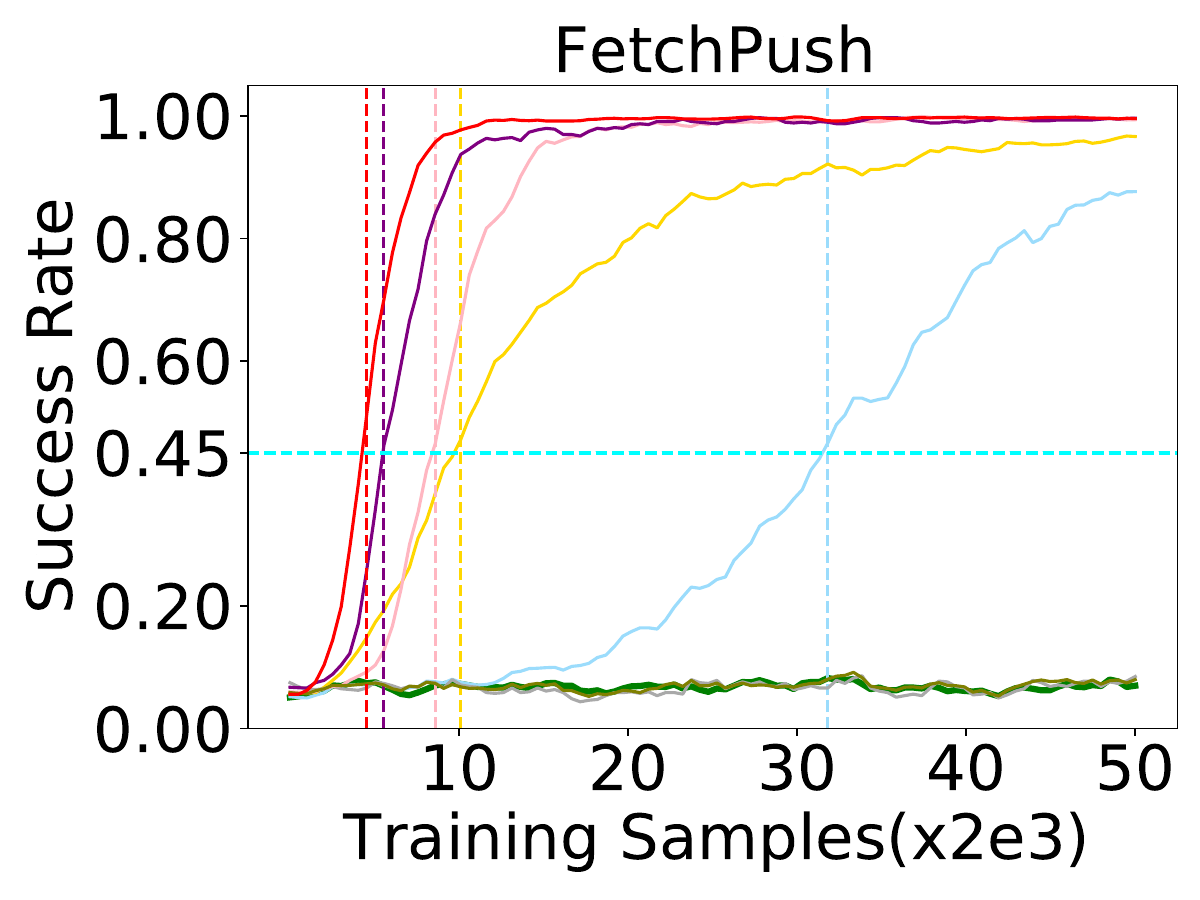}}
    \end{minipage}
    \begin{minipage}{0.459\linewidth}
        
        \centerline{\includegraphics[width=0.96\textwidth]{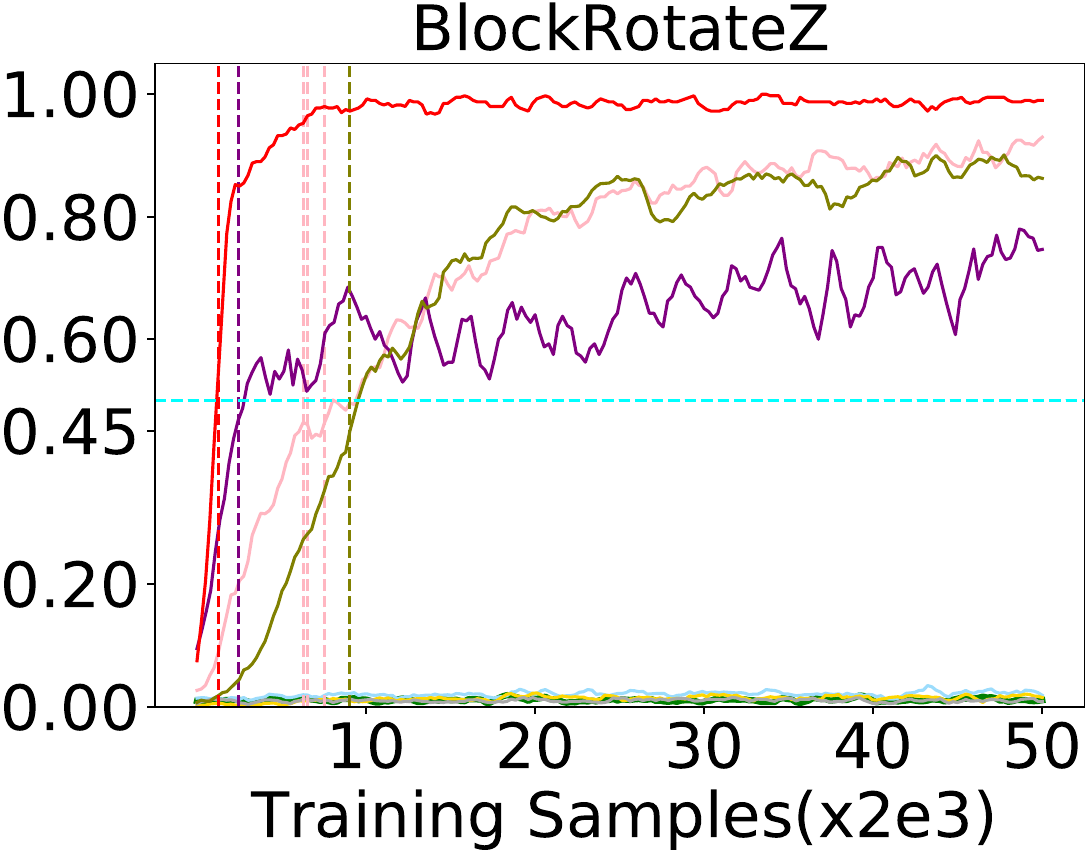}}
    \end{minipage}
    \caption{
    Number of training samples needed with respect to mean success rate for Fetchpush and HandManipulateBlockRotateZ tasks (the lower the better).
    }
   \label{fig:multi_goal sample_efficiency}
\end{figure}

To assess the sample efficiency of baseline methods in comparison to GCHR, we examined the number of training samples (i.e., $\left\langle {s,a,g^{\prime},g} \right\rangle$ tuples) necessary to obtain a particular mean success rate. This comparative analysis is depicted in Figure \ref{fig:multi_goal sample_efficiency}.

From the FetchPush task, depicted on the left side of Figure \ref{fig:multi_goal sample_efficiency}, we observe that to attain the 0.45 mean success rate, the competitive baseline DDPG+HER requires over 6000 training samples, whereas GCHR only needs approximately 4000 samples. This indicates that GCHR is 1.5 times more sample efficient than DDPG+HER.

In another task, BlockRotateZ, GCHR uses the fewest number of samples to attain the same 0.5 mean success rate. These findings demonstrate that GCHR significantly enhances sample efficiency compared to other baseline methods, underscoring its effectiveness in improving learning performance with fewer training samples.

\subsection{Error Bars of Mean Performance}
\begin{figure*}[h]
    \begin{minipage}{0.245\linewidth}
        
        \centerline{\includegraphics[width=\textwidth]{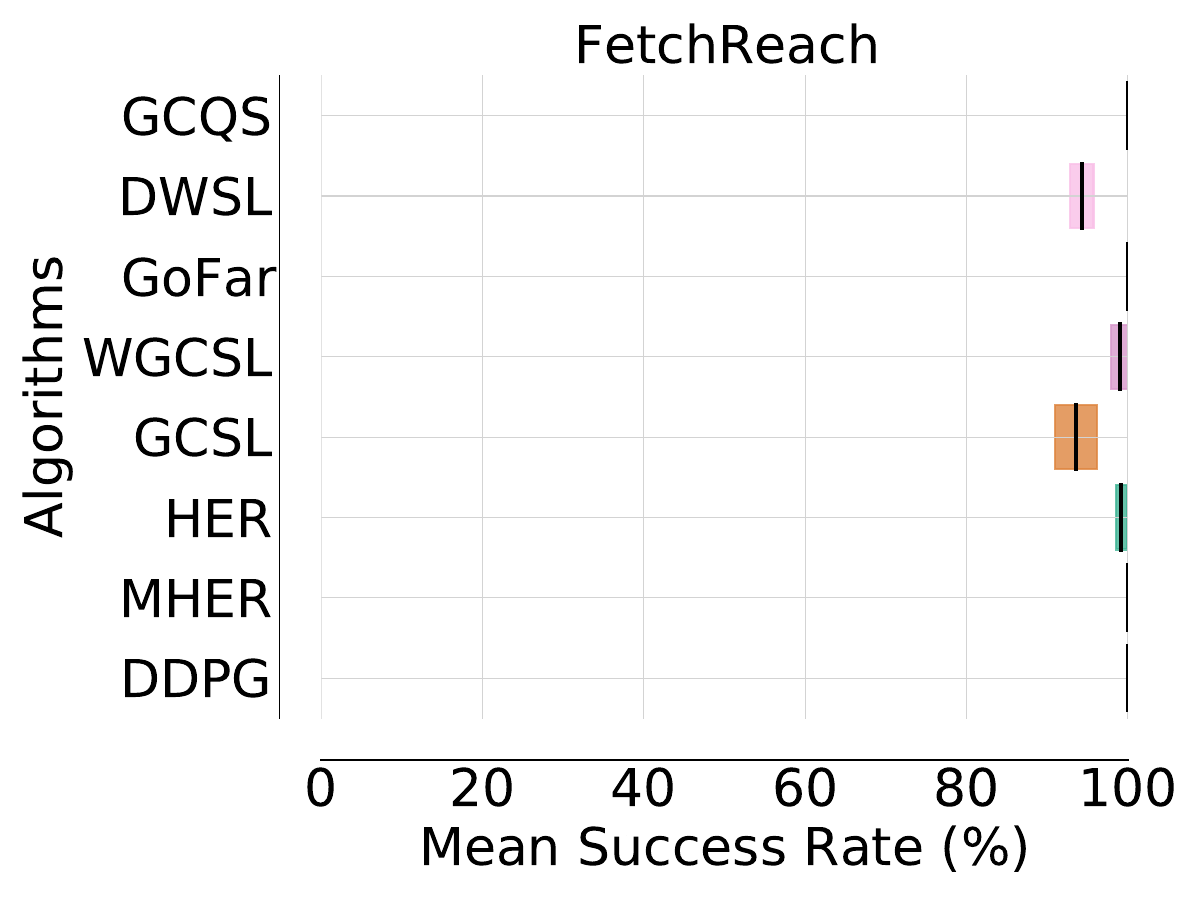}}
    \end{minipage}
    \begin{minipage}{0.245\linewidth}
        
        \centerline{\includegraphics[width=0.96\textwidth]{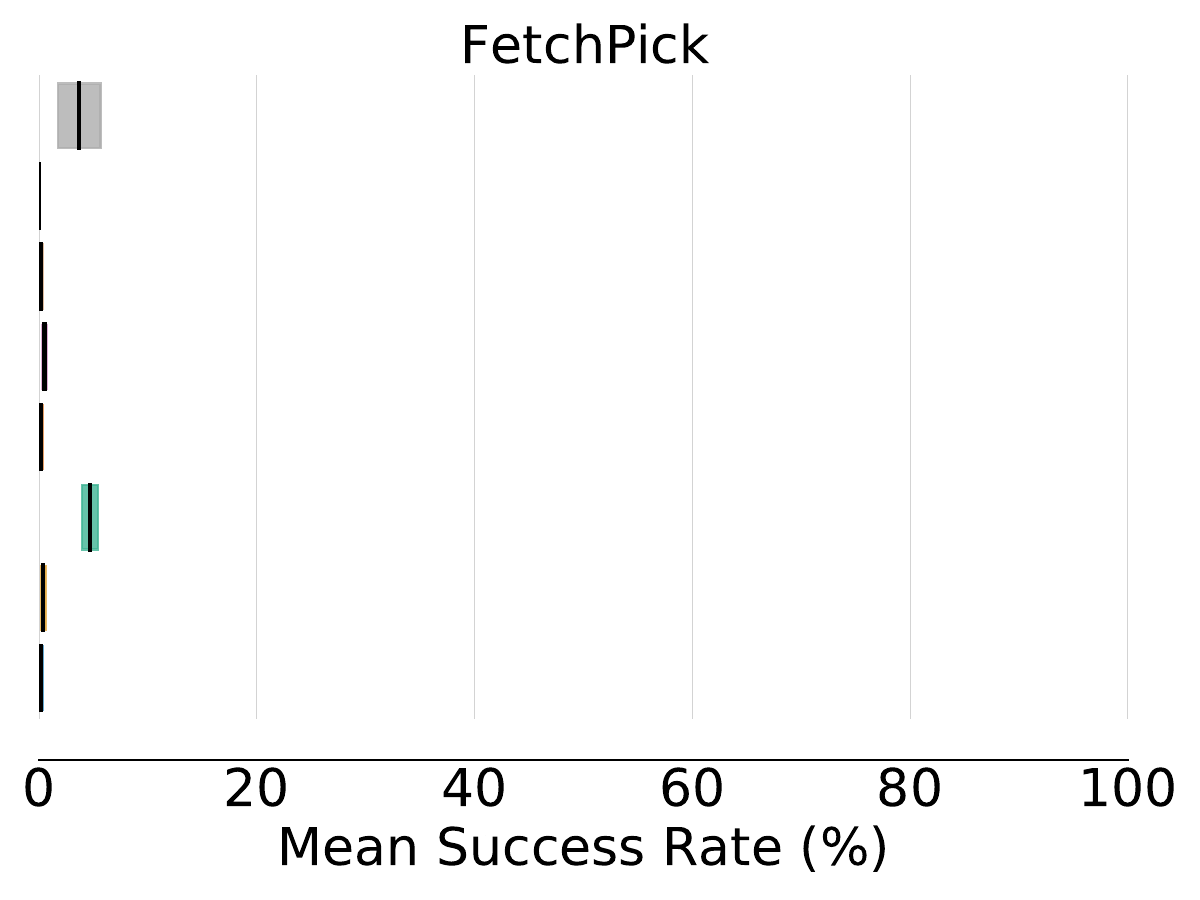}}
    \end{minipage}
    \begin{minipage}{0.245\linewidth}
        
        \centerline{\includegraphics[width=0.96\textwidth]{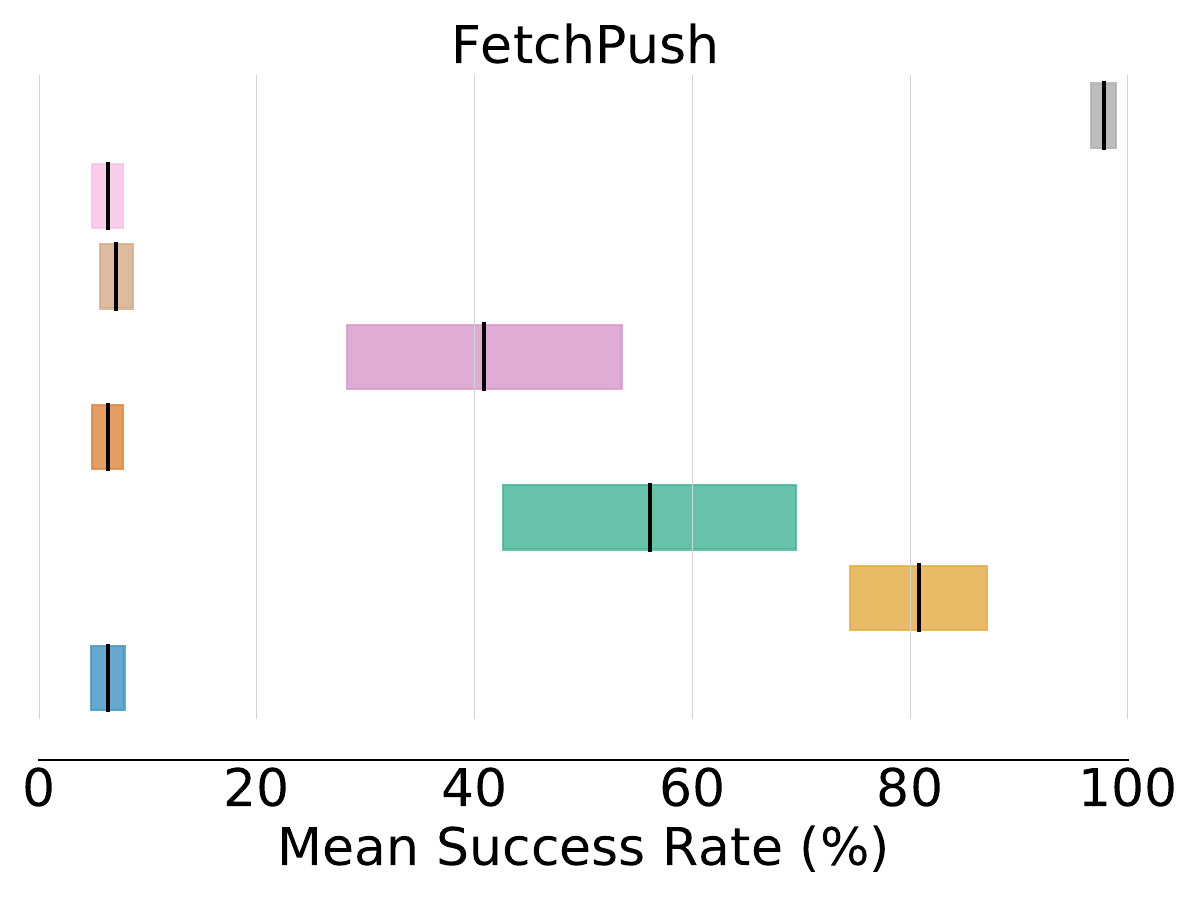}}
    \end{minipage}
    \begin{minipage}{0.245\linewidth}
        
        \centerline{\includegraphics[width=0.96\textwidth]{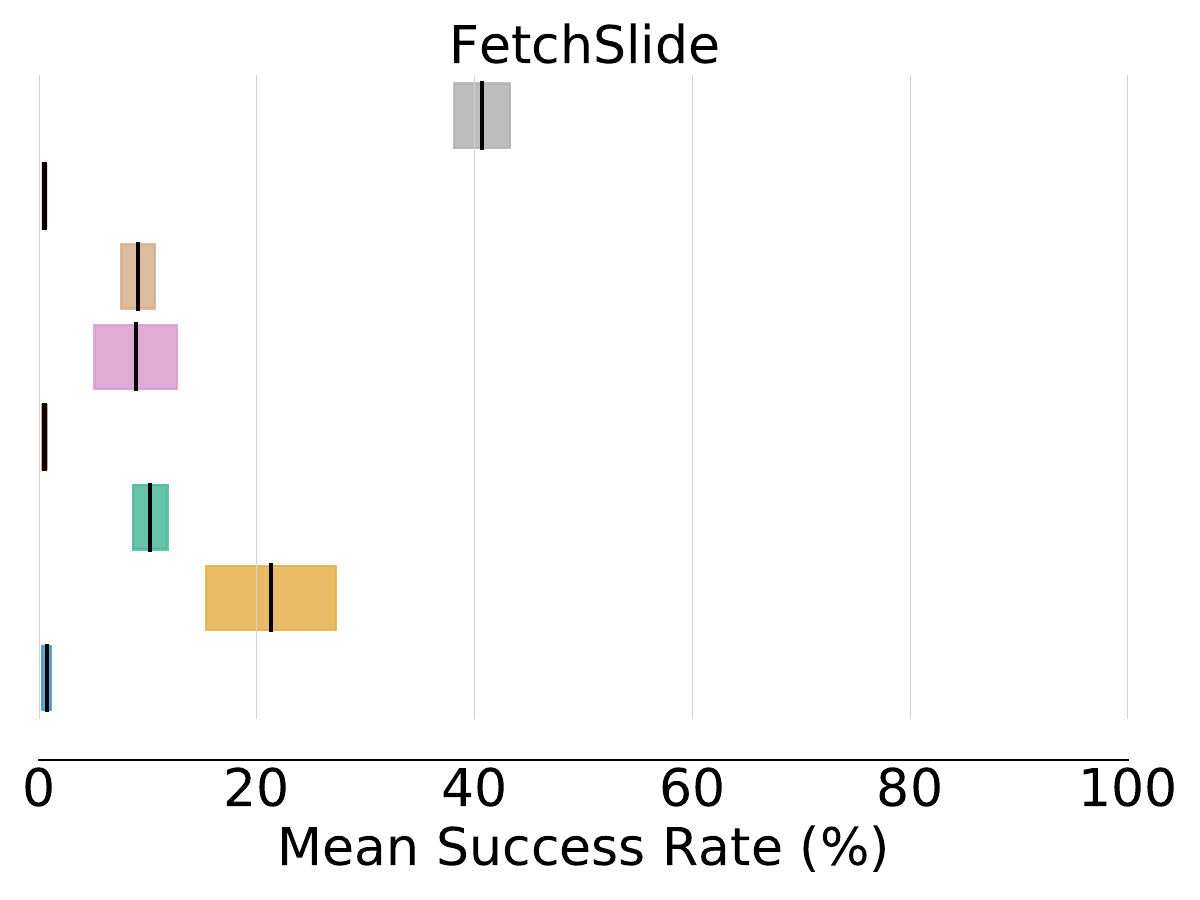}}
    \end{minipage}
    \begin{minipage}{0.245\linewidth}
        
        \centerline{\includegraphics[width=\textwidth]{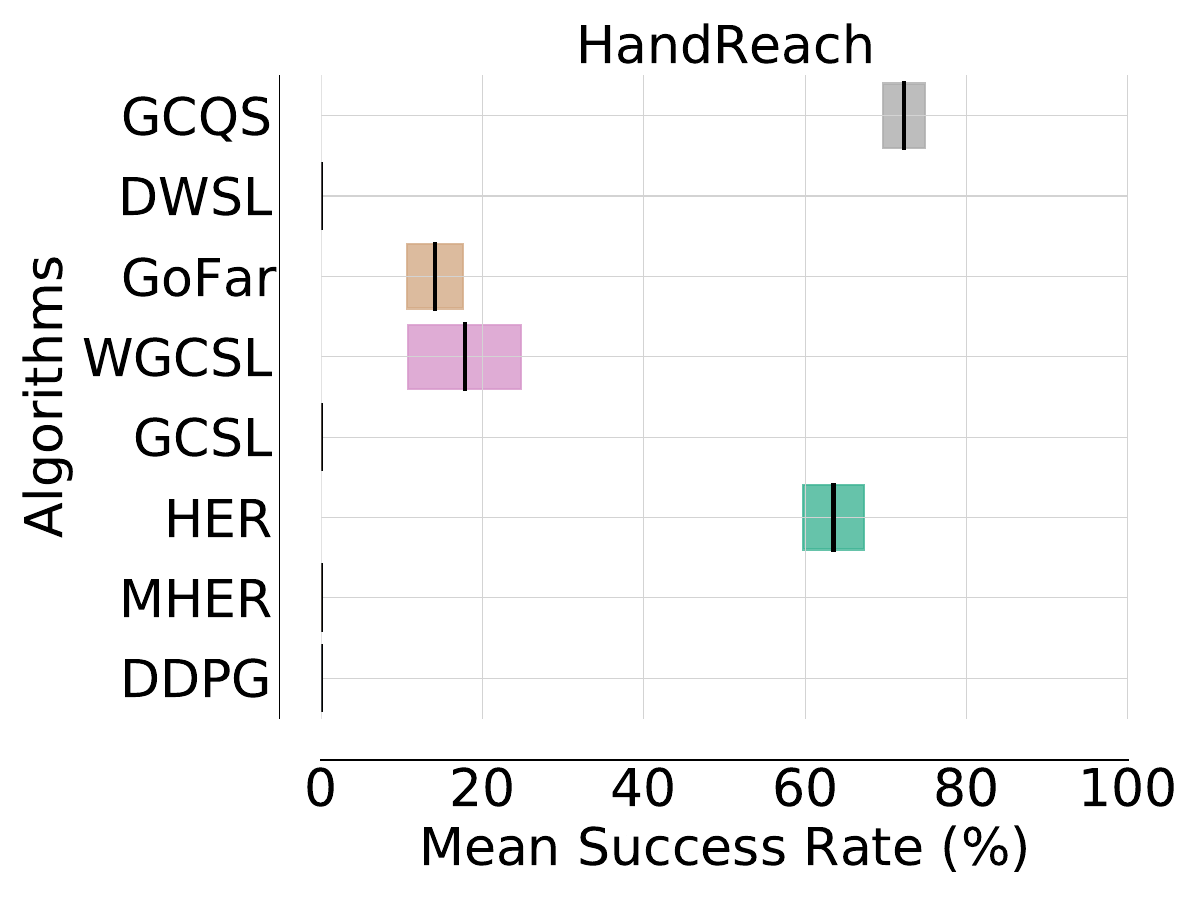}}
    \end{minipage}
    \begin{minipage}{0.245\linewidth}
        
        \centerline{\includegraphics[width=0.96\textwidth]{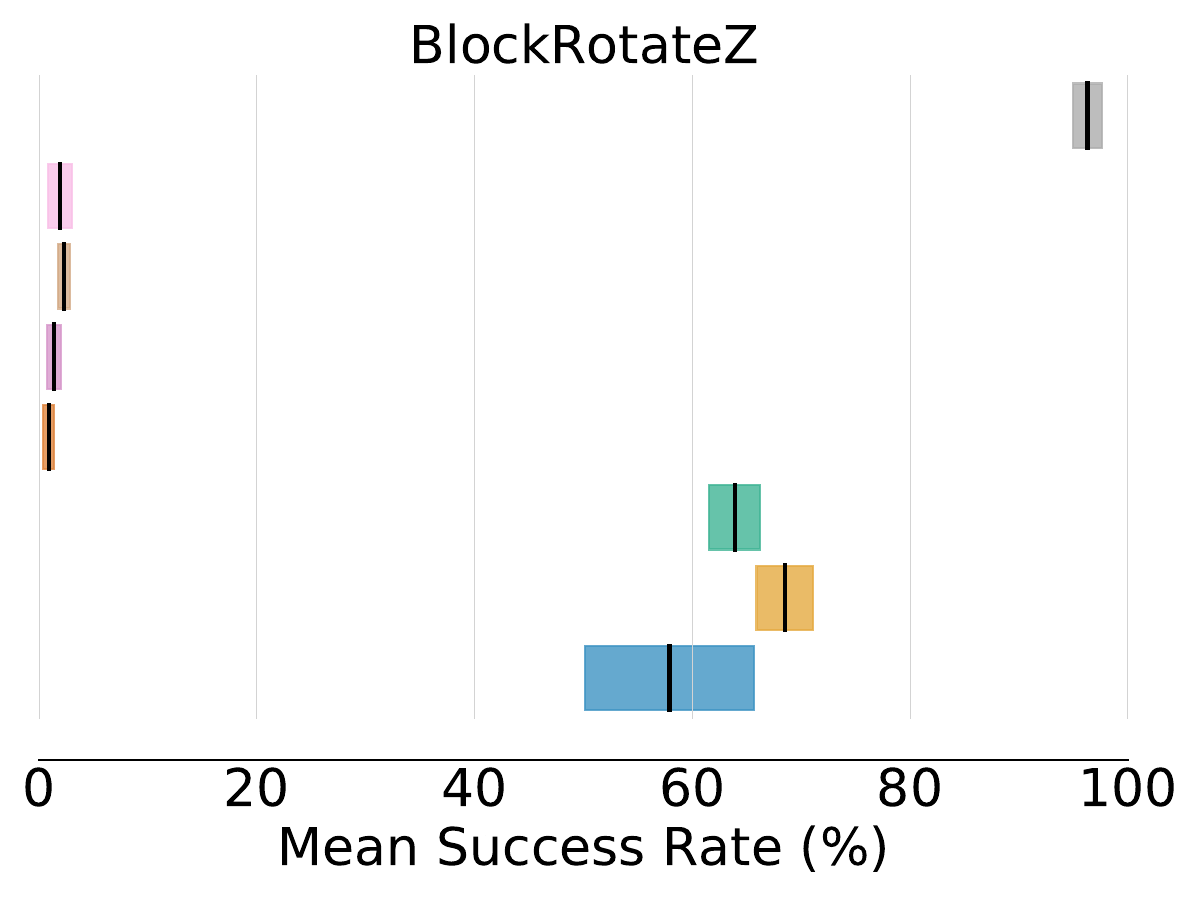}}
    \end{minipage}
    \begin{minipage}{0.245\linewidth}
        
        \centerline{\includegraphics[width=0.96\textwidth]{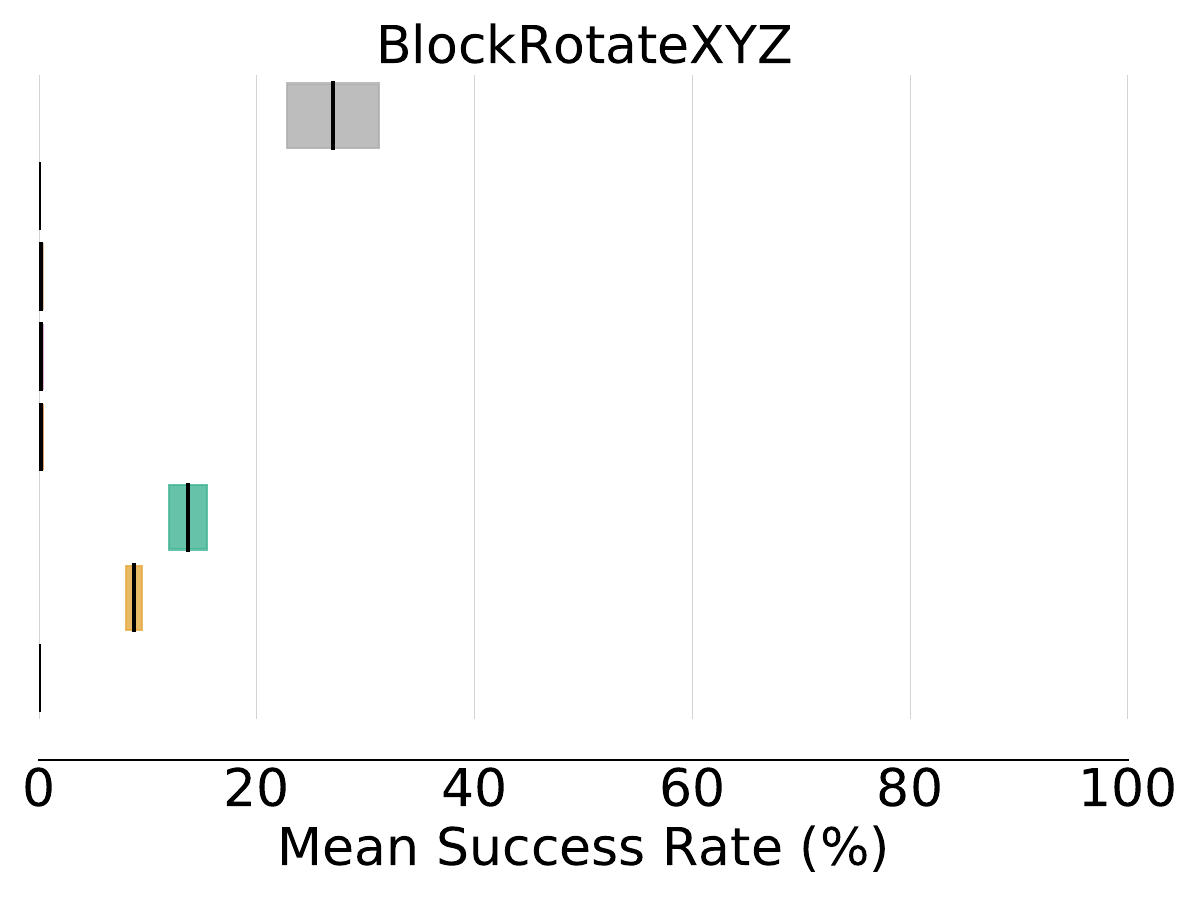}}
    \end{minipage}
    \begin{minipage}{0.245\linewidth}
        
        \centerline{\includegraphics[width=0.96\textwidth]{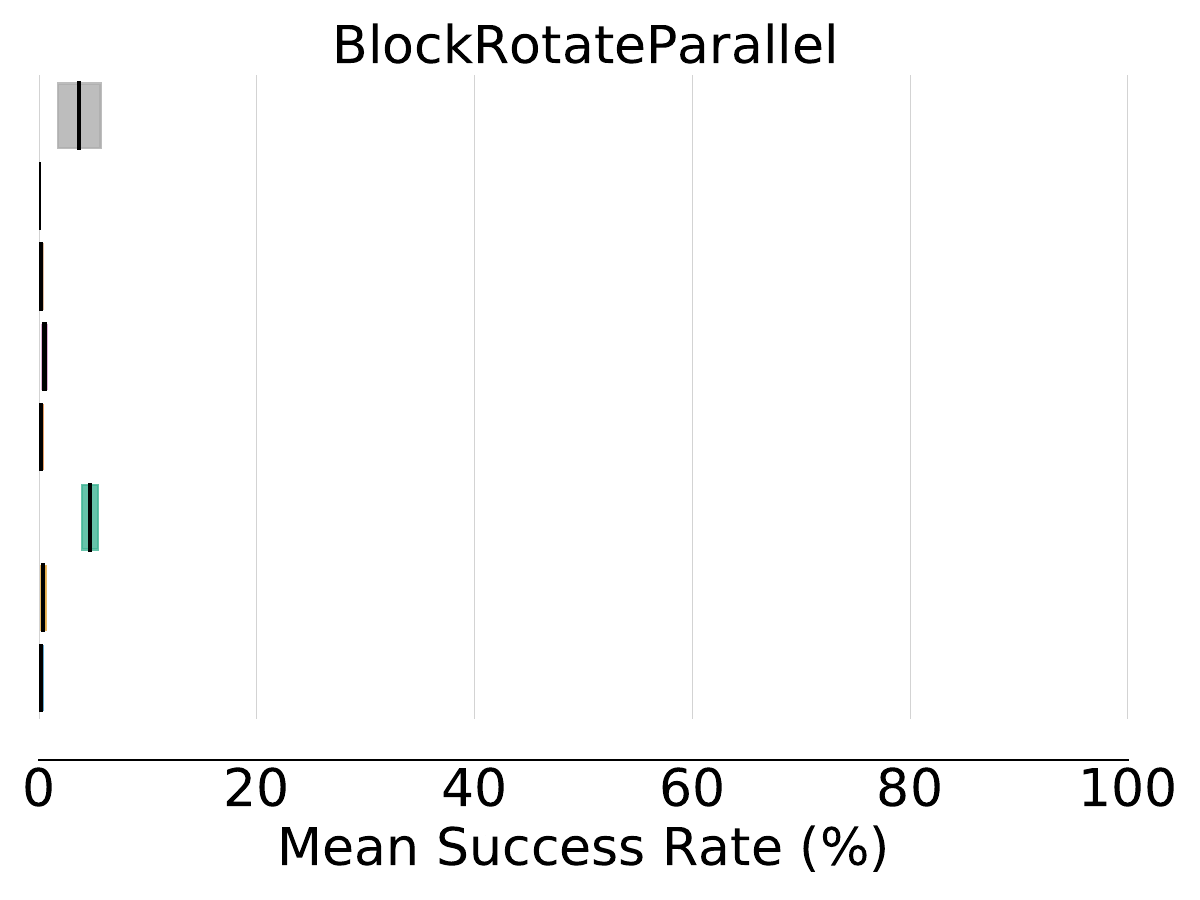}}
    \end{minipage}
    \caption{
     The error bars for each goal-conditioned task presented in \ref{fig:multi_goal results}.
     Error bars represent the standard error of the mean (SEM) for each algorithm's average performance across multiple seeds in each task.}
    \label{fig:error_bars}
\end{figure*}

To further assess the effectiveness and robustness of the algorithm, we present error bar plots for each task based on the mean ± standard deviation (SD) of results across five seeds for each algorithm. As shown in \ref{fig:error_bars}, the GCHR algorithm demonstrates a significant advantage across all goal-conditioned tasks. Its mean success rate approaches 100$\%$ on simpler tasks (e.g., FetchReach and BlockRotateZ), and it substantially outperforms other algorithms on moderately challenging tasks (e.g., FetchPush and HandReach), with shorter error bars indicating greater result stability and robustness. However, in more difficult tasks (e.g., BlockRotateXYZ and BlockRotateParallel), the performance of GCHR declines, as evidenced by lower success rates and longer error bars, suggesting performance fluctuations. Overall, GCHR exhibits strong learning capabilities in complex goal spaces but still has room for improvement, particularly in handling extreme tasks such as high-dimensional rotations and parallel rotations.
\begin{figure*}[h]
    \begin{minipage}{\linewidth}
        
        \centerline{\includegraphics[width=0.6\textwidth]{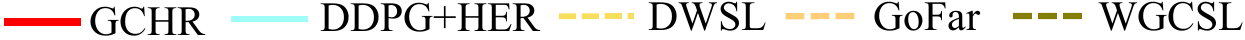}}
    \end{minipage}
    \begin{minipage}{0.245\linewidth}
        
        \centerline{\includegraphics[width=\textwidth]{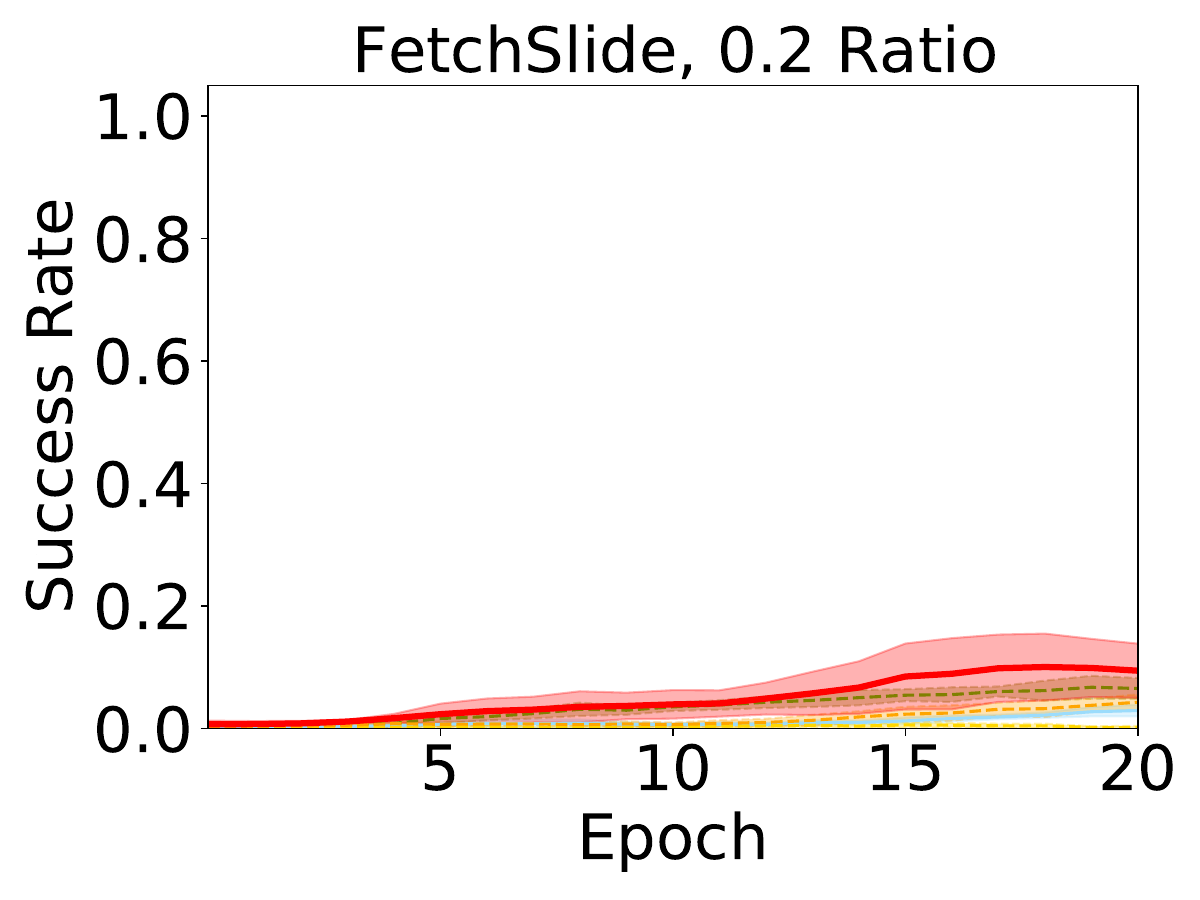}}
    \end{minipage}
    \begin{minipage}{0.245\linewidth}
        
        \centerline{\includegraphics[width=0.96\textwidth]{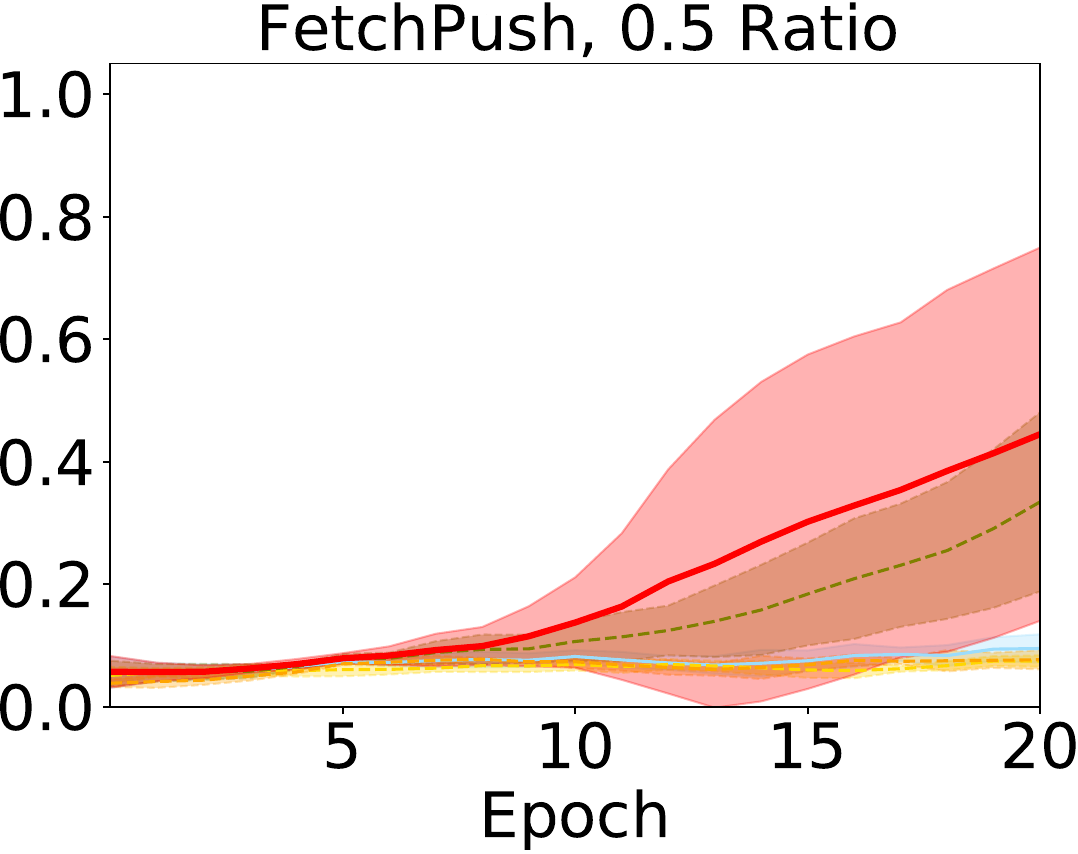}}
    \end{minipage}
    \begin{minipage}{0.245\linewidth}
        
        \centerline{\includegraphics[width=0.96\textwidth]{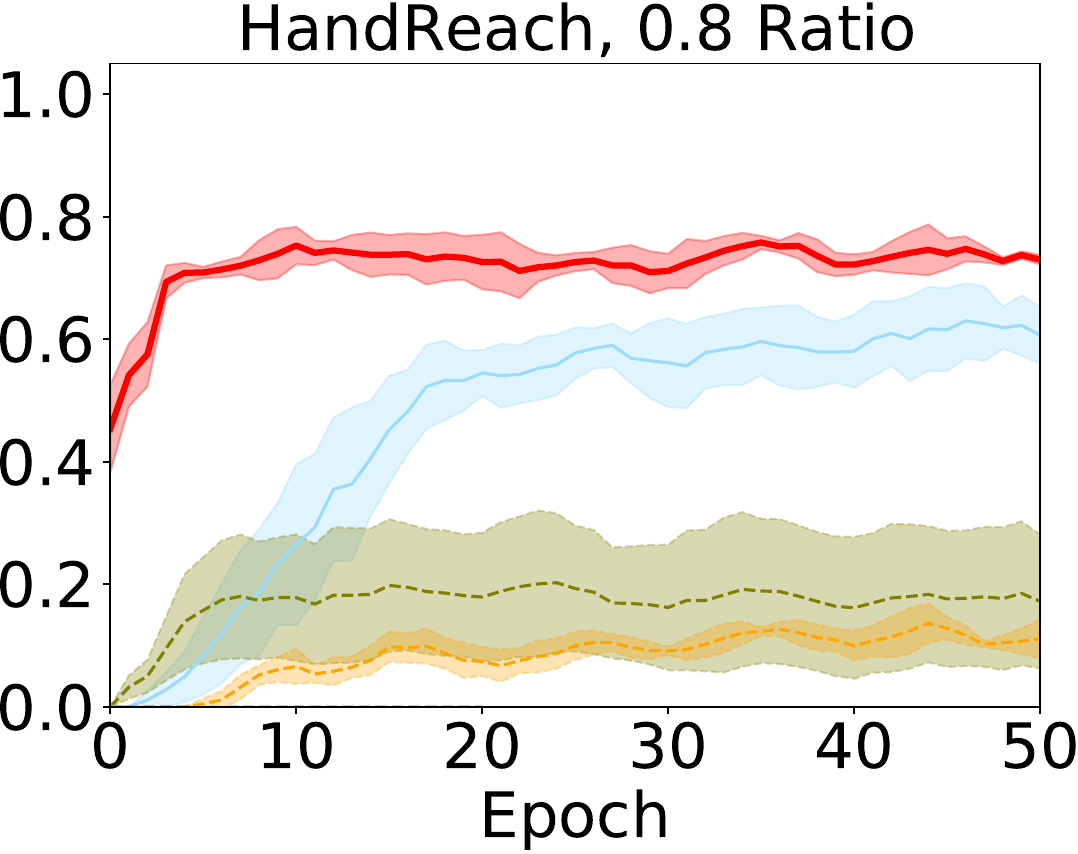}}
    \end{minipage}
    \begin{minipage}{0.245\linewidth}
        
        \centerline{\includegraphics[width=0.96\textwidth]{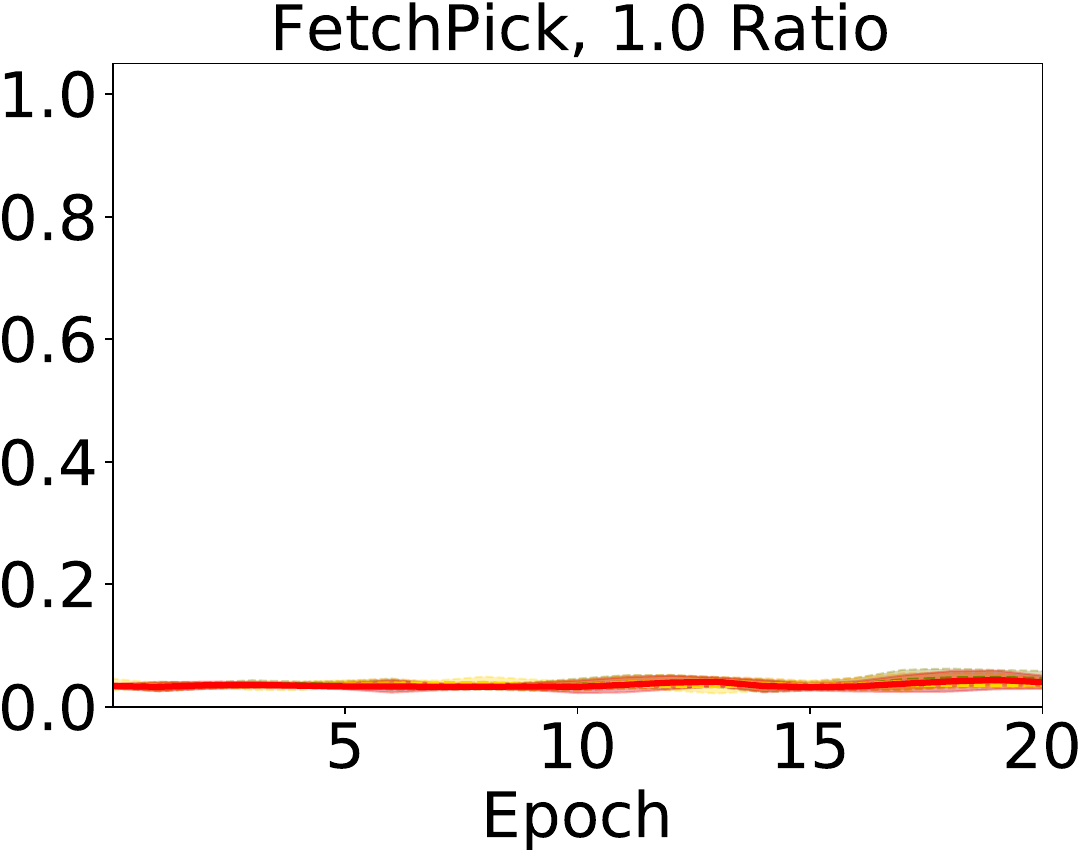}}
    \end{minipage}
    \caption{
     Relabel ratio ablation studies in such goal-conditioned tasks. Results are averaged over 5 random seeds and the shaded region represents the standard deviation.
    }
    \label{fig:multi_goal relabel_rate results}
\end{figure*}
\subsection{Relabeling Ratio} 
Given our approach to learning in GCRL settings, which assumes data annotated with relabeled goals, this study examines the influence of explicit goal labels on performance. We conducted experiments across four distinct relabeling ratios 
(i.e.,
$\{0.2, 0.5, 0.8, 1.0\}$) in various environments to evaluate algorithmic efficacy. As illustrated in \ref{fig:multi_goal relabel_rate results}, GCHR exhibits substantial resilience to variations in the relabeling ratio. Furthermore, GCHR consistently surpasses competing algorithms such as WGCSL and DDPG+HER across different labeling ratios.
\section{Implementations  Details} \label{ap:baseline details}
In this section, we provide experimental details omitted in Section \ref{sc:experiment} of the main paper. These include (1) technical and architecture details for all methods, (2) experimental evaluating setup, (3) hyperparameter settings and (4) environment descriptions.
\subsection{Algorithm and Architecture} We employ the off-policy actor-critic algorithm with HER \citep{andrychowicz2017hindsight} as our foundational GCRL framework. This sequential comparison experiment allows us to directly assess the relative performance and effectiveness of each approach under identical conditions. Additionally, temporal difference (TD)-learning is utilized for value function estimation, and soft updates are applied to network parameters. 
Our implementation adheres to the optimal parameter settings as outlined in \citet{liu2023metric}. The hyperparameters for all baseline methods remain consistent. For GCHR, the policy objective parameter $\beta$ is set to 0.2. For further details of parameter $\beta$, refer to Appendix \ref{sc:beta_alpha hyperparameter}. Our implementation of baselines and GCHR draw knowledge from and references the following six code repositories:
\begin{itemize}
    \item DDPG, DDPG+HER, MHER, GCSL, WGCSL: \url{https://github.com/Cranial-XIX/metric-residual-network};
    \item GoFar: \url{https://github.com/JasonMa2016/GoFAR};
    \item DWSL: \url{https://github.com/jhejna/dwsl/};
\end{itemize}
\textbf{Baseline Descriptions.} We compare our method, GCHR, with state-of-the-art goal-conditioned algorithms, including GCAC and supervised learning methods:
\begin{itemize}
    \item \textbf{DDPG} \citep{lillicrap2015continuous}
    We adopt an open-source implementation of Deep Deterministic Policy Gradient (DDPG) that has been pre-optimized for the set of multi-goal tasks.
    In order to maintain consistency,
    we ensure that all other methods are implemented based on this framework,
    employing identical architectures and hyper-parameters where appropriate.
    The critic objective is:
    \begin{equation}
    \min_Q\mathbb{E}_{(s_t,a_t,s_{t+1},g)\sim \mathcal{B}_r}[(r(s_t,g)+\gamma Q(s_{t+1},\pi(s_{t+1},g),g)
    -Q(s_t,a_t,g))^2]
    \end{equation}
    The policy objective is:
    \begin{equation}
    \min_{\pi}-\mathbb{E}_{(s_t,a_t,s_{t+1},g)\sim \mathcal{B}_r}[Q(s_t,\pi(s_t,g),g)]
    \end{equation}
    where $\mathcal{B}_r$ is the relabeled data from replay buffer.
    \item \textbf{DDPG+HER} \citep{andrychowicz2017hindsight}
    Here we use the goal-conditioned version as our baseline based on open-source implementation of Hindsight Experience Replay
    \citep{andrychowicz2017hindsight},
    maintaining the same architecture and hyper-parameters when appropriate.
    Its critic objective is:
    \begin{equation} 
    \min_Q\mathbb{E}_{(s_t,a_t,g)\sim \mathcal{B}_r}[Q(s_t,\pi(s_t,g),g)].
    \end{equation}
    Its policy objective is:
    \begin{equation}
    \min_{\pi}-\mathbb{E}_{(s_t,a_t,s_{t+1},g)\sim \mathcal{B}_r}(r_t+\gamma Q(s_{t+1},\pi(s_{t+1},g),g)-
    Q(s_t,a_t,g))^2.
    \end{equation}
    In the process of iteratively updating the actor and critic,
    we relabel the goal $g$ for the transition in the replay buffer.
    In our multi-goal settings,
    a batch of trajectories comprising achieved goals is randomly sampled from the replay buffer.
    Subsequently,
    the future strategy is employed to label each state within these trajectories.
    These labeled states are then combined with desired goals in a controlled manner,
    ensuring that the labeling ratio remains below or equal to 0.8.
    This amalgamation of labeled states and desired goals constitutes the relabeled goals,
    which serves as the goals for the policy to learn.
    \item \textbf{MHER}
    Model-based Hindsight Experience Replay 
    \citep{yang2021mher},
    using off-policy samples after model-based relabeling to maximize:
    \begin{equation}
    \min_{\pi}-\mathbb{E}_{joint}=-\mathbb{E}_{(s_t,g)\sim B_m}\big[Q(s_t,\pi(s_t,g),g)\big]+
    \alpha\mathbb{E}_{(s_t,a_t,g)\sim \mathcal{B}_m}\big[\|a_t-\pi(s_t,g)\|_2^2\big],
    \end{equation}
    where $\mathcal{B}_m$ is the model-based relabeled data from replay buffer.
    Its critic objective is:
    \begin{equation}
    \min_Q\mathbb{E}_{(s_{t},a_{t},g,r_{t},s_{t+1})\sim \mathcal{B}_m}\big[(y_{t}-Q(s_{t},a_{t},g))^{2}\big],
    \end{equation}
    where $y_{t}$ indicates target value.
    \item \textbf{GCSL}
    GCSL
    \citep{ghosh2021learning}
    is implemented by removing the critic component of the Deep Deterministic Policy Gradient (DDPG) algorithm and modifying the policy loss to be based on maximum likelihood estimation:
    \begin{equation}
    \min_{\pi}-\mathbb{E}_{(s,a,g)\sim \mathcal{B}_r}\left[\log\pi(a\mid s,g)\right].
    \end{equation}
    The following baselines are all on the offline goal-conditioned.
    We re-implement them on the online.
    \item \textbf{WGCSL}
    WGCSL
    \citep{yang2022rethinking}
    is implemented as an extension of GCSL by incorporating a $Q$-function.
    The $Q$-function is trained using TD error,
    similar to the DDPG algorithm,
    and is augmented with an advantage weighting in the regression loss.
    The advantage term that we compute is denoted as $A(s_t,a_t,g)=r(s_t,g)+\gamma Q(s_{t+1},\pi(s_{t+1},g),g)-Q(s_t,a_t,g)$.
    By employing this denotes,
    the policy objective of WGCSL is:
    \begin{equation}
    \min_{\pi}-\mathbb{E}_{(s_t,a_t,\phi(s_i))\thicksim \mathcal{B}_r}\left[\gamma^{i-t}\exp_{clip}(A(s_t,a_t,\phi(s_i)))\log\pi(a_t\mid s_t,\phi(s_i))\right],
    \end{equation}
    where the clip $\exp_{clip}$ for numerical stability.
    \item \textbf{GoFar}
    GoFar
    \citep{ma2022offline}
    proposed the adoption of a state-matching objective for GCRL,
    wherein a reward discriminator and a bootstrapped value function were employed to assign weights to the imitation learning loss.
    The policy objective of GoFar is:
    \begin{equation}
    \min_{\pi}-\mathbb{E}_{(s_t,a_t,g)\sim \mathcal{B}}[\log\pi(a_t|s_t,g)\max(A(s,a,g)+1,0)],
    \end{equation}
    where the advantage function is estimated by discriminator-based rewards.
    The discriminator $c$ is learned to minimize $\mathbb{E}_{g\sim p(g)}[\mathbb{E}_{p(s;g)}\big[\log c(s,g)\big]+\mathbb{E}_{(s,g)\sim D}[\log(1-c(s,g))]\big]$.
    The value function $V$ is learned to minimize $(1-\gamma)\mathbb{E}_{(s,g)\thicksim\mu_0,p(g)}[V(s,g)]+\frac12\mathbb{E}_{(s,a,g,s^{\prime})\thicksim D}[(r(s;g)+\gamma V(s^{\prime};g)-V(s;g)+1)^2]$ for $V\geq0$.
    We also use GoFar with an efficient version:
    GoFar(HER).
    In GoFar(HER),
    hindsight goal relabeling is the goal-relabeling distribution:
    $p_{\mathrm{HER}}(g\mid s_{t},a_{t},\tau)=q[\phi(s_{t}),...,\phi(s_{T})]$,$s_{t},...,s_{T}$ are come from a trajectory $\tau=\{s_{0},a_{0},r_{0},...,s_{T};g\}$.\\
    \item \textbf{DWSL}
    DWSL \citep{hejna2023distance} is a new supervised learning method which models the entire empirical
    distribution
    $p_{\theta}^r$
    of discrete distances 
    $\hat{d}$
    between states as values in offline data.
    we re-implement it in off-policy setting and this distribution training with following principle:
    \begin{equation}
    \max_\theta\mathbb{E}_{\mathcal{B}_r}\left[\log p_\theta^r\left((j-i-1)//N|s_i,\phi(s_j)\right)\right]
    \end{equation}
    where $s_i,\phi(s_j),j>i$ is the relabeled data from replay buffer $\mathcal{B}$,
    $N$ is N-step.
    DWSL views goal-conditioned value prediction—which enables policy improvement beyond imitation—
    as a supervised classification problem.
    DWSL compute meaningful statistics of distances distribution to extract shortest path estimates between any two states without the optimization challenges of bootstrapping.
    Specifically,
    DWSL use the LogSumExp as a smooth estimate of the minimum distance.
    DWSL compute distance between any two states as followings:
    \begin{equation}
    \hat{d}(s_i,\phi(s_j))=-\alpha\log\mathbb{E}_k\sim p_\theta^r(\cdot|(s_i,\phi(s_j))\begin{bmatrix}e^{-k/(B\alpha)}\end{bmatrix}
    \end{equation}
    and
    \begin{equation}
    \hat{d}(s_{i+1},\phi(s_j))=-\alpha\log\mathbb{E}_k\sim p_\theta^r(\cdot|(s_{i+1},\phi(s_j))\begin{bmatrix}e^{-k/(B\alpha)}\end{bmatrix}
    \end{equation}
    where $s_i,s_{i+1},\phi(s_j),j>i$ is the relabeled data,
    $B=T//\tilde{N}$ is the bins in distribution
    $p_{\theta}^r$ and 
    $\alpha$ is the temperature.
    Finally,
    DWSL re-weight actions by how closely they reduce the estimated distance to the goal:
    \begin{equation}
    \text{adv}=\hat{d}(s_i,\phi(s_j))-c(s_i,\phi(s_j))-\hat{d}(s_{i+1},\phi(s_j)),
    \end{equation}
    where $c(s_i,\phi(s_j))=1\{\phi(c_{i+1})\neq\phi(s_j)\}/B$.
    Then its policy extraction is:
    \begin{equation}
    \max_{\psi}\mathbb{E}_{\mathcal{D}_r}\left[e^{\text{adv}/\beta}\log\pi_{\psi}(a_i|s_i,\phi(s_j))\right],
    \end{equation}
    where $\beta$ is also the temperature and the $\pi$ is the policy with parameter $\psi$.
\end{itemize}
\subsection{Evaluation Setup}
For each baseline and task, evaluations were conducted using five random seeds (e.g., $\{100, 200, 300, 400, 500\{$). The policy was trained for 1,000 episodes per epoch. At the end of each training epoch, policy performance was assessed by computing the mean success rate over 100 independent rollouts, each initialized with randomly sampled goals. The resulting success rates were then averaged across the five seeds, and the mean performance was plotted over learning epochs, with the standard deviation visualized as a shaded region in the performance curves.
\subsection{Experimental Hyperparameters} \label{ap:hyperparameters}
We consistently utilize the Adam optimizer 
\citep{kingma2014adam}
across all experimental setups. For each state, relabeled goals are uniformly sampled from all future states within its trajectory. In environments applying discount factors, we set $\gamma = 0.98$ for all goal-conditioned tasks. Each algorithm follows a predetermined set of hyperparameters specifically designed for goal-conditioned environments. DWSL, GoFar, WGCSL, GCSL, MHER, and DDPG have been previously calibrated for our task set, and we have adopted the parameter values as reported in prior research. Our implementation of GCHR methods sHSRe the same network architecture as DDPG, thus utilizing DDPG's hyperparameter values. All experiments in this paper use the following hyperparameters, which have been found by the aforementioned search:
\begin{table}[H]
    \renewcommand{\arraystretch}{1.5}
    \centering
    \footnotesize 
    \label{table:dt-hyperparameters}
    \begin{tabular}{p{3.5cm}|p{3.0cm}} \toprule
        Actor and critic networks & value \\ \midrule 
        Learning rate & 1e-3 \\
        Buffer size & $10^6$ transitions \\
        Polyak-averaging coefficient & 0.95\\
        Action L2 norm coefficient & 1.0 \\
        Observation clipping & [-200,200] \\
        warmup steps & $5000$ \\
        Batch size & 256 \\
        Rollouts per MPI worker & 2 \\
        Number of MPI workers & 16\\
        Epochs & 50\\
        Cycles per epoch & 50\\
        Batches per cycle & 40\\
        Test rollouts per epoch & 10\\
        Probability of random actions & 0.3\\
        Scale of additive Gaussian noise & 0.2\\
        Probability of HER experience replay & 0.8\\
        Normalized clipping & [-5, 5]\\
        $\beta$ & 0.2\\
        \bottomrule
    \end{tabular}
    \caption{Hyperparameters for Baselines.}
\end{table}
All hyperparameters are described in greater detail in \citep{andrychowicz2017hindsight}.
\subsection{Environment Details}
\label{supp:env}
In this section, we describe the tasks in our experiments in Section \ref{sc:experiment}. All goal-conditioned tasks are sourced from OpenAI Gym \citep{brockman2016openai}.

\textbf{Fetch Tasks.}
The Fetch robotic manipulation suite (i.e., FetchReach, Push, Slide, Pick) evaluates 7-DoF arm control across four goal-oriented tasks: positional alignment, object displacement, planar sliding, and aerial retrieval. These benchmarks sHSRe a unified structure with high-dimensional state vectors (joint positions/velocities) and 4D action spaces (actuator control + gripper operation). Employing goal-conditioned rewards, success requires achieving spatial targets within $\mu=0.05$ tolerance. Task differentiation emerges through manipulation requirements: direct end-effector positioning versus indirect object interactions (propelling beyond workspace limits or precision aerial placement). The reward function is defined as:
\begin{equation}
\ r(s,a,g_{XYZ})=\begin{cases}0,&\|\phi(s)-g_{XYZ} \|_2^2 \leq \mu \\-1,&\text{otherwise}\end{cases},
\end{equation}

\textbf{Hand Tasks.}
The Hand manipulation benchmarks (i.e., HandReach, BlockRotateZ, BlockRotateXYZ, BlockRotateParallel) require precise coordination of a 20-DoF Shadow Hand for dexterous object manipulation. These high-dimensional control tasks utilize multimodal state observations (joint kinematics + object dynamics) and employ sparse binary rewards ( $\mu=0.01$ tolerance) identical to Fetch environments. Task complexity escalates through constrained rotational demands - from single-axis alignment to multi-axis synchronization - providing rigorous evaluation platforms for GCRL in high-DoF settings.

\end{document}